\documentclass[journal]{IEEEtran}

\usepackage{amsmath,amssymb}
\usepackage{algorithmic}
\usepackage{graphicx}
\usepackage[hyphens]{url}
\usepackage{hyperref}
\usepackage{float}

\newcommand{\real}{{\mathbb{R}}}
\usepackage{color}
\newcommand{\fangcong}{}
\newcommand{\hanze}{}
\begin{document}
\bstctlcite{IEEEexample:BSTcontrol}

\title{Mathematical Models of \\Overparameterized Neural Networks}

\author{Cong Fang, ~ 
        \and Hanze Dong, ~
        \and Tong Zhang 
\thanks{C. Fang is with  University of Pennsylvania, USA.}
\thanks{H. Dong and T. Zhang are with the Hong Kong University of
  Science and Technology.}
  }

%

\IEEEoverridecommandlockouts
\IEEEpubid{\begin{minipage}{\textwidth}
\centering
0000--0000~\copyright~2020 IEEE. Personal use is permitted, but republication/redistribution requires IEEE permission.
\\ See \url{https://www.ieee.org/publications/rights/index.html} for more information.
\end{minipage}}


\maketitle

\begin{abstract}
Deep learning has received considerable empirical successes in recent years. However, while many ad hoc tricks have been discovered by practitioners, until recently, there has been a lack of theoretical understanding for tricks invented in the deep learning literature.
Known by practitioners that overparameterized neural networks are easy to learn, in the past few years there have been important theoretical developments in the analysis of overparameterized neural networks. In particular, it was shown that such systems behave like convex systems under various restricted settings, such as for two-layer NNs, and when learning is restricted locally in the so-called neural tangent kernel space around specialized initializations.   
This paper discusses some of these recent progresses leading to significant better understanding of neural networks. We will focus on the analysis of two-layer neural networks, and explain the key mathematical models, with their algorithmic implications. We will then discuss challenges in understanding deep neural networks and some current research directions.

\end{abstract}


\IEEEpeerreviewmaketitle

\section{Introduction}
Neural Networks (NNs)   are computational models that are  composed of (possibly multiple) feature representation layer(s), and a final linear learner.
In recent years, deep NNs have largely improved the state-of-the-art
performances in  numerous real applications, such as image
classification \cite{krizhevsky2012imagenet,he2016deep},  speech
recognition \cite{hinton2012deep},  natural language processing
\cite{bahdanau2014neural}, etc. However, theoretical understanding of
these empirical successes for NNs is still limited. One main
conceptual difficulty is the high non-convexity of these models, which
means that first-order algorithms such as gradient descent (GD) or
stochastic gradient descent (SGD)
may converge to bad local stationary points.  

However, it is observed in practice that with the help of a number of tricks such as dropout \cite{SHKSS-dropout} and batch normalization \cite{ioffe15}, deep neural network (DNN) can be reliably trained from random initialization with reproducible results. The solutions obtained by proper training procedures behave well and consistently. In other words, two different random initializations (using the same initialization and training strategy) generally lead to models that give similar predictions on test data. Thus, we may conclude that proper neural network training leads to similar solutions. This behavior resembles that of convex optimization, instead of generic non-convex optimization problems that tend to get stuck in suboptimal local stationary solutions. Because solutions from different random initializations are similar and reproducible, it can also be conjectured that with proper training, deep neural networks can reach solutions that are near global optimal. 
These empirical observations appear to be rather mysterious, and they require the developments of new mathematical models for neural networks that can bridge the gap between non-convex and convex models to understand.

In addition to the above empirical observations, it is also known by practitioners that overparameterized neural networks (NNs) with many hidden units
are easy to learn \cite{zhang2016understanding}. They achieve better and more consistent performance. Related to this empirical observation, it was noticed in the 1990s that
neural networks with infinitely many hidden units are easier to model and analyze theoretically \cite{Neal95-thesis,Williams97-nips}.
In the past few years,
there have been many significant theoretical developments in the analysis
of overparameterized NNs with massive hidden units that approach infinity.
Especially, it was shown that  such systems behave  like convex systems under various restricted settings. This provides theoretical justifications of the empirical observations of the reproducibility of neural network training.


\fangcong{ In this paper, we review some recently developed  mathematical models for   overparameterized  NNs, with the focus on the {\em neural tangent kernel} (NTK) view and the {\em mean field} (MF) view. Section~\ref{sec:2two-levelNN} introduces the basic formulation for two-layer NNs and Section~\ref{sec:random} introduces a closely related learning model, random kitchen sinks \cite{rahimi2009weighted}.  In Section \ref{sec:ntk}, we examine the   NTK view for two-layer NNs, which shows that a two-layer NN can be written equivalently as a linear model  in the tangent space under some specialized conditions.  Section \ref{sec:meanfield} considers the MF view for two-layer NNs. In this view, a continuous two-layer NN is regarded as a learned distribution over the weights, which leads to a more realistic mathematical model for analyzing practical behaviors of NNs. In Section \ref{sec:exp}, we compare the three models from the feature learning perspective. Section \ref{sec:deepnn} considers the possible extensions on deep NNs for NTK and MFs. In Section \ref{sec:complexity}, we introduce some basic complexity results for NTK.  Section \ref{sec:othermodel} reviews some other mathematical models. Finally, we conclude the paper and outline active research directions in Section \ref{sec:conclusion}.}

\IEEEpubidadjcol

\section{Two-layer Neural Networks}\label{sec:2two-levelNN}

Two-layer NNs have a history dating back to the 1940s \cite{kleene1951representation}.
A discrete two-layer neural network can be viewed as a $k$-dimensional vector valued function of a $d$-dimensional
input vector $x$, which has the following form:
\begin{equation}
f([u,\theta],x) = \frac{\alpha}{m} \sum_{j=1}^m u_j h(\theta_j,x) , \label{eq:twolayer-nn}
\end{equation}

where $x \in \real^d$, $\theta_j \in \real^d$, and
$u_j \in \real^k$. 
The model parameters are $\{[u_j,\theta_j]: j=1,\ldots,m\}$, which will be learned via training. 
Here $\alpha>0$ is a real-valued scaling parameter that is not learned. It is included here to differentiate two different regimes of overparameterized neural networks. In this system, there are $m$ hidden units (or neurons), and each hidden unit corresponds to a function $h(\theta_j,x)$ of the input $x$.
It maps the original input feature $x$ to a new feature $h(\theta_j,x)$, with a parameter $\theta_j$ that is learned. 
The function $h(\theta,x)$ is a real valued function. In applications, it often takes the following form 
\[
h(\theta,x)= h_0(\theta^\top x) ,
\]
where $h_0(\cdot)$ is called an activation function, and the standard choices include rectified linear unit (ReLU) $h_0(z)=\max(0,z)$ and sigmoid $h_0(z)=e^z/(1+e^z)$. 

In order to learn the NN parameters, we consider the minimization of the following optimization problem:
\begin{align}
&  \min_{u,\theta} \phi(u,\theta) , \label{eq:opt} \\
&\phi(u,\theta)= \frac1n \sum_{i=1}^n L(f([u,\theta],x_i),y_i)
+ R(u,\theta) . \nonumber
\end{align}
Here $\{(x_i,y_i): i=1,\ldots,n\}$ are the training data, 
$L$ is a loss function, such as the soft-max loss for $k$-class classification problem with $v \in \real^k$ and $y \in \{1,\ldots,k\}$:
\[
  L(v,y) = - \log \frac{\exp(v_{y})}{\sum_{j=1}^k \exp(v_j)} ,
\]
and $R(u,\theta)$ is a regularizer (also called weight-decay in the neural network literature), such as the $L_2$ regularization:
\[
R(u,\theta) = \frac1{2m} \sum_{j=1}^m \left[
\lambda_u \|u_j\|_2^2
+ \lambda_\theta \|\theta_j\|_2^2 \right] .
\]
In this paper, we consider the situation that the regularizer $R(u,\theta)$ is convex in $(u,\theta)$, and the loss function $L(v,y)$ is convex in $v$. 

In general, we consider random initialization, and specifically random Gaussians:
\[
u_j \sim N(0,\sigma^2), 
\quad \theta_j \sim N(0,\sigma^2) ,
\]
with different scalings of $\alpha$. The training is often performed via SGD, where we randomly pick a training datum $i$ (or a mini-batch of data points) and update parameters as:
\begin{align*}
  u \leftarrow& u - \eta \nabla_u [L(f([u,\theta],x_i),y_i) + R(u,\theta)] ,\\
  \theta \leftarrow& \theta - \eta \nabla_\theta [L(f([u,\theta],x_i),y_i) + R(u,\theta)] .
\end{align*}
The parameter $\eta$ is referred to as learning rate. It is known in optimization that if we let $\eta \to 0$ properly, then the procedure converges to a point $(u_\infty,\theta_\infty)$ such that $\nabla \phi(u_\infty,\theta_\infty)=0$. 
If the loss function $L(v,y)$ is convex in $v$, then $\phi(u,\theta)$ is convex in $u$ but not convex in $\theta$. Therefore in general, $\nabla \phi(u_\infty,\theta_\infty)=0$ does not imply that $(u_\infty,\theta_\infty)$ achieves a global optimal solution. In order to understand the empirical observation that $(u_\infty,\theta_\infty)$ behaves like a solution that is close to global optimal, especially when $m$ is large, mathematical models have been developed in recent years to explain the empirical phenomenon.

\section{Random Features}
\label{sec:random}

Before developing mathematical models for two-layer NNs, we will first consider a closely related machine learning model that employs random features. We note that in a two-layer NN, the hidden units are feature functions $h(\theta_j,x)$ that
contain parameters $\theta_j$ to be learned during neural network training. The random feature approach (denoted by RF in this paper) is also referred to as {\em random kitchen sinks} \cite{rahimi2009weighted,rahimi2008uniform,bach2017equivalence}. In RF,
we still consider the function \eqref{eq:twolayer-nn}, but assume that $\theta_j$ are fixed at $\tilde{\theta}_j$ that is generated from a random distribution, typically a Gaussian distribution:
\[
\tilde{\theta}_j \sim N(0,\sigma^2) ,
\]
which is not learned during training. Only parameters $\{u_j:j=1,\ldots,m\}$ are learned. In this case, we may take any fixed scaling
$\alpha$ such as $\alpha=1$, because the scaling is not important for the random feature approach.

In RF, the model $f(\cdot,x)$ becomes linear with respect to the model parameters $\{u_j\}$. Therefore, if $L(v,y)$ is convex in $v$, then the objective function \eqref{eq:opt} is convex, and thus the convergence of SGD is easy to analyze. 

Since the parameters $\{\tilde{\theta}_j\}$ do not change during training, we apply SGD to learn $\{u_j\}$ only in the training process that optimizes \eqref{eq:opt}:
\[
u \leftarrow u - \eta \nabla_u [L(f([u,\tilde{\theta}],x_i),y_i) + R(u,\tilde{\theta})] .
\]

We are interested in the situation that the number of hidden units $m \to \infty$. It can be shown that in this case, the function learned by the random feature method converges to a well-defined limit \cite{rahimi2009weighted}. To understand its behavior, it is useful to consider the kernel view for RF.

Note that if we let 
\begin{align*}
u=&\frac{\alpha}{\sqrt{m}} [u_1,\ldots,u_m]^\top \in \real^{d \times k}, \\
h(x)=& \frac{1}{\sqrt{m}}[h(\tilde{\theta}_1,x),\ldots,h(\tilde{\theta}_m,x)]^\top \in \real^d ,
\end{align*}
then the random feature method corresponds to the linear model
\begin{equation}
f(u,x)= u^\top h(x) . \label{eq:model-rand}
\end{equation}
We consider the $L_2$ regularization
\[
  R(u,\theta) = \frac{\lambda}{2m}\sum_{j=1}^m \|\alpha u_j\|_2^2
  = \frac{\lambda}{2} \|u\|_F^2 ,
\]
where $\|\cdot\|_F$ is the Frobenius norm of a matrix.
Then with fixed $\theta_j$, the objective function $\phi(u,\theta)$ of \eqref{eq:opt} becomes
\begin{equation}
\phi(u)=\frac{1}{n}\sum_{i=1}^n L(u^\top h(x_i),y_i) + \frac{\lambda}{2} \|u\|_F^2 , \label{eq:opt-rand}
\end{equation}
which is convex in $u$.

In order to obtain the kernel representation, we note that the first order optimality condition of \eqref{eq:opt-rand} at the optimal solution can be written as:
\begin{equation}
  u = \sum_{i=1}^n h(x_i) \beta_i^\top,\label{eq:u-beta}
 \end{equation}
 where
\[
\beta_i = -\frac{1}{\lambda n} L_1'(u^\top h(x_i),y_i) ,
\]
and $L_1'(v,y) \in \real^k$ is the gradient of $L(v,y)$ with respect to $v$.

This leads to the kernel representation
\cite{SchSmo18}, where the  kernel is defined as the inner product of the feature vectors $h(x)$ and $h(x')$ for two input variables $x \in \real^d$ and $x' \in \real^d$:
\[
  k_m(x,x') = h(x)^\top h(x')= \frac{1}{m} \sum_{j=1}^m h(\tilde{\theta}_j, x) h(\tilde{\theta}_j,x') .
\]
Consider a function represented using this kernel with parameters $\beta_i \in \real^k$ for $i=1,\ldots,n$:
\[
  f_m(\beta,x) = 
  \sum_{i=1}^n \beta_i k_m(x,x_i) .
\]
Then we can see from equation \eqref{eq:u-beta} that
$f_m(\beta,x)=u^\top h(x)=f(u,x)$. That is, the original linear function has a kernel representation. Moreover, the regularizer also has a kernel representation:
\[
\|u\|_F^2 = \sum_{j=1}^n\sum_{\ell=1}^n (\beta_j^\top \beta_\ell) k_m(x_i,x_j) .
\]
Using the kernel representation, 
the solution of RF, which minimizes the objective function \eqref{eq:opt-rand} over $u$, is equivalent to the solution of the following kernel optimization problem:
\begin{align}
&\min_\beta \frac1n \sum_{i=1}^n L(f_m(\beta,x_i),y_i) + R_m(\beta) , \label{eq:ker-opt}\\
&  R_m(\beta) = \frac{\lambda}{2} \sum_{j=1}^n \sum_{\ell=1}^n (\beta_j^\top \beta_\ell) k_m(x_i,x_j) .\nonumber
\end{align}
Let $\beta$ be the solution of \eqref{eq:ker-opt}, 
we may obtain the solution of \eqref{eq:opt-rand}
using the relationship between $\beta$ and $u$ in \eqref{eq:u-beta}.

The kernel formulation \eqref{eq:ker-opt} is particularly useful for analyzing the behavior of  overparameterized RF in the limiting case of $m \to \infty$. 
This is because when $m \to \infty$, we have
\[
  k_m(x,x') \to k_\infty(x,x') = \int h(\tilde{\theta},x) h(\tilde{\theta},x') d \rho_0(\tilde{\theta}) ,
\]
which is a well-defined kernel, 
where $\rho_0(\tilde{\theta})$ is the random distribution of $\tilde{\theta}$, such as the Gaussian distribution $N(0,\sigma^2)$ in our case.
It follows that as $m \to \infty$, the kernel function 
\[
 f_m(\beta,x) = \sum_{i=1}^n \beta_i k_m(x,x_i) \to f_\infty(\beta,x) ,
\]
where
\[
 f_\infty(\beta,x) = \sum_{i=1}^n \beta_i k_\infty(x,x_i) .
\]
Moreover the corresponding optimization problem of \eqref{eq:ker-opt} becomes
\begin{align}
&\min_\beta \frac1n \sum_{i=1}^n L(f_\infty(\beta,x_i),y_i) + R_\infty(\beta) , \label{eq:opt-rand-inf}\\
&  R_\infty(\beta) = \frac{\lambda}{2} \sum_{i=1}^n \sum_{\ell=1}^n (\beta_i^\top \beta_\ell) k_\infty(x_i,x_\ell) , \nonumber
\end{align}
which is also well-defined.

Note that in the kernel formulation, the number of model parameters $\{\beta_i\}$ is $n$, which remains finite when $m \to \infty$. Therefore the limit of $\{\beta_i\}$ is well-defined. 

If we consider the original random feature function \eqref{eq:model-rand} in the case of $m \to \infty$, the number of parameters $\{u_j\}$ will also approach infinity. In this case, we may consider $u$ as a function of $\tilde{\theta}$, and write \eqref{eq:model-rand} as
\[
  f(u,x) = \int u(\tilde{\theta}) h(\tilde{\theta},x) d \rho_0(\tilde{\theta}) 
\]
in the limit of $m \to \infty$,
and write the two norm regularizer as
\[
  R(u) = \frac{\lambda}{2} \int \|u(\tilde{\theta})\|_2^2 d \rho_0(\tilde{\theta}) . 
\]
With this notation for $m=\infty$, we have the relationship
\[
  u(\tilde{\theta}) = \sum_{i=1}^n \beta_i h(\tilde{\theta},x_i) ,
\]
where $\beta$ is the solution of the kernel formulation \eqref{eq:opt-rand-inf}.

It is known that RF works well for certain problems \cite{rahimi2009weighted,rahimi2008uniform}. However, for many real-world applications such as image classification, RF is inferior to NN that learns better feature representations than random ones.  We will discuss the feature learning perspective in Section~\ref{sec:exp}.

\fangcong{
 One interesting extension of the RF theory which can be used to analyze the behavior of NN  is presented in \cite{daniely2017sgd} and   \cite{daniely2016toward}. A  kernel called {\em conjugate kernel} was introduced,  and it was shown that the Gradient Descent (GD) process for the NNs under some special initializations belongs to this kernel space.  Moreover,  any
function in this kernel space can be approximated by changing the weights of the last layer. Inspired by the above findings,  the authors proved the global convergence of GD for NNs. However, in their analysis,  only the GD updates  on the last layer
contributes to the global convergence, which ignores weight updates in the lower layers. In the next section, we will introduce the neural tangent kernel view, which develops theoretical results showing that weight updates in the bottom layer of two-layer NNs can also contribute to the  convergence of GD.
}

\section{Neural Tangent Kernels}\label{sec:ntk}

In practice, the random feature approach is often inferior to two-layer neural networks because the parameters $\{\theta_j\}$ are not trained.
As we have seen, the overparameterized  case with $m \to \infty$ corresponds to kernel learning with a well-defined kernel. One natural question is whether this point of view can be generalized to handle two-layer neural networks, where the parameters $\{\theta_j\}$ are trained together with $\{u_j\}$.
Such a generalization leads to {\em neural tangent kernels} (NTK) \cite{jacot2018neural}, which we shall describe in this section.
We note that the connection of infinitely-wide NNs and kernel methods (Gaussian processes) has already been known in the 1990s \cite{Neal95-thesis,Williams97-nips}. However, the more rigorous theory of NTK has only appeared very recently, e.g., \cite{jacot2018neural,du2018gradient}. 

In NTK, we consider a specialized scaling and random initialization of parameters. The special scaling makes it possible to consider NN parameters in a small region around the initial value when $m \to \infty$.
The resulting NN with parameters restricted in this region can be well approximated by a  linear model fitted with random features.
Similar to RF of Section~\ref{sec:random}, this linearization induces a  kernel in the tangent space around the initialization, which becomes  Neural Tangent Kernel \cite{jacot2018neural}, since the weights are near their initial values during the training of the neural network. This phenomenon is also referred to as the  ``lazy training''  regime in \cite{chizat2019lazy}. In this regime, the system becomes linear, and the dynamics of GD (or SGD) within this region can be tracked via properties of the associated NTK. 

There have been a series of studies about NTK in  recent years,  in which a number of researchers  proved  polynomial convergence rates to the global optimal, e.g., \cite{li2018learning, du2018gradient, allen2019can,zou2018stochastic, arora2019fine,allen2019convergence,  allen2019learning, yang2019scaling,oymak2018overparameterized,zou2019improved,mei2019mean,arora2019exact, tzen2020mean, yang2019theoretical,cai2019neural,liu2019neural,wang2019neural,zhang2020generative,liao2020provably}  and  sharper generalization errors, e.g.,  \cite{li2018learning,arora2019fine,allen2019can,su2019learning,cao2019generalization,ji2019polylogarithmic}.
The scaling $\alpha$ also plays a significant role in the Neural Tangent Kernel view,  where a relatively large scaling $\alpha$ of order $\sqrt{m}$ is needed.
This means that when $m \to \infty$, $\alpha \to \infty$. 

To derive NTK under the assumption of $m \to \infty$, we consider the case that $h(\theta,x)$ is differentiable with respect to $x$, such as the sigmoid or tanh activation function. Note that the non-differentiable ReLU activation function can also be handled similarly, although many works consider the differentiable assumption for simplicity.

We now consider a random initialization at $[\tilde{u},\tilde{\theta}]$, around which we can linearly approximate the neural network
as:
\begin{align}
  f([u,\theta],x) \approx& \frac{\alpha}{m} \sum_{j=1}^m \left[
    \tilde{u}_j h(\tilde{\theta}_j,x) + (u_j-\tilde{u}_j) h(\tilde{\theta}_j,x)\right.\label{eq:lin-approx}\\
&\left.    + \tilde{u}_j (\theta_j-\tilde{\theta}_j)^\top \nabla_\theta h(\tilde{\theta}_j,x)
    \right] + \text{high order terms} ,\nonumber
\end{align}
where we assume that both $u-\tilde{u}$ and $\theta-\tilde{\theta}$ are small. 
Note that the theory of NTK requires $\alpha=O(\sqrt{m})$, so that the term
$(\alpha/m)\sum_{j=1}^m \tilde{u}_j h(\tilde{\theta}_j,x)$ has a bounded variance.
A large scaling $\alpha$ ($\alpha \to \infty$ as $m \to \infty$) is important for this linearization, because  if we fix the coefficient $\alpha \tilde{u}_j (\theta_j-\tilde{\theta})$ for the random feature $\nabla_\theta h(\tilde{\theta}_j,x)$, then when the scaling $\alpha \to \infty$ (as $m \to \infty$), \fangcong{one can show that a  small change of $\theta$ leads to a big change of the output function values. This means that in the continuous limit, we should let  $(\theta_j-\tilde{\theta}) \to 0$}. A similar claim holds for $u_j-\tilde{u}_j$. In this case, the higher order terms will be $o(\alpha \|u_j-\tilde{u}\|+\alpha \|\theta_j-\tilde{\theta}_j\|)$ that approach zero, and the linear approximation of \eqref{eq:lin-approx} is accurate.

This linear approximation employs random features $h(\tilde{\theta}_j,x)$ and $\tilde{u}_j\nabla_\theta h(\tilde{\theta}_j,x)$ for $j=1,\ldots,m$. Compared to the RF approach, which only uses the random features $h(\tilde{\theta}_j,x)$, two-layer NNs use additional random features $\tilde{u}_j\nabla_\theta h(\tilde{\theta}_j,x)$. 

In order to motivate the NTK kernel, we consider the following representation, similar to \eqref{eq:u-beta} for RF:
\begin{equation}
  \begin{cases}
    u_j =& \tilde{u}_j + \alpha^{-1} \sum_{i=1}^n \beta_i^u h(\tilde{\theta}_j,x_i)  \\
    \theta_j =& \tilde{\theta}_j + \alpha^{-1} \sum_{i=1}^n \tilde{u}_j^\top\beta_i^\theta \nabla h(\tilde{\theta}_j,x_i) 
  \end{cases}. \label{eq:ntk-param}
\end{equation}
Here we have both $\beta_i^u \in \real^k$ and $\beta_i^\theta \in \real^k$. 
Using this representation, the linear approximation of \eqref{eq:lin-approx} becomes
\begin{align*}
&\frac{1}{m} \sum_{j=1}^m \left[
    \alpha\tilde{u}_j h(\tilde{\theta}_j,x) + \sum_{i=1}^n\beta_i^u h(\tilde{\theta}_,x_i) h(\tilde{\theta}_j,x)\right.\\
&\left.    + \sum_{i=1}^n \tilde{u}_j \tilde{u}_j^\top \beta_i^\theta \nabla_\theta h(\tilde{\theta}_j,x_i)^\top \nabla_\theta h(\tilde{\theta}_j,x)
    \right] .
\end{align*}

Using the relationship of kernel as the inner product of features for linear models, this linear approximation of two-layer NN
corresponds to the following kernel function representation
\[
  f_m([\beta^u,\beta^\theta],x) = \sum_{i=1}^n [\beta_i^u k_m^u(x,x_i) + k_m^\theta(x,x_i) \beta_i^\theta] ,
\]
where
\[
  k_m^u(x,x') = \frac{1}{m}\sum_{j=1}^m h(\tilde{\theta}_j,x) h(\tilde{\theta}_j,x') ,
\]
which corresponds to the features of the linear coefficients $u$ (also used by the RF model of Section~\ref{sec:random}), and 
\[
  k_m^\theta(x,x') = \frac{1}{m}\sum_{j=1}^m \tilde{u}_j\tilde{u}_j^\top \nabla_\theta h(\tilde{\theta}_j,x)^\top\nabla_\theta h(\tilde{\theta}_j,x') ,
\]
which is a $k \times k$ matrix corresponding to the features of the linear coefficients $\theta$ (which was not used by the RF model of Section~\ref{sec:random}).
This kernel representation is referred to as NTK.

In the NTK representation, from the relationship of $[u,\theta]$ and $[\beta^u,\beta^\theta]$ in \eqref{eq:ntk-param}, we can see that as $\alpha \to \infty$, $\theta \to \tilde{\theta}$ and $u \to \tilde{u}$, which means that we have a more accurate linear approximation when $\alpha$ is large.

In the infinity-width limit of $m \to \infty$, the kernel becomes the infinite-width NTK kernel, which is well-defined:
\begin{align*}
  k_m^u(x,x') \to& k_\infty^u(x,x') = \int h(\tilde{\theta},x) h(\tilde{\theta},x')  d \rho_0(\tilde{\theta}) \\
  k_m^\theta(x,x') \to& k_\infty^\theta(x,x')\\ &= \int \tilde{u}\tilde{u}^\top \nabla_\theta h(\tilde{\theta},x)^\top \nabla_\theta h(\tilde{\theta},x')  d \rho_0(\tilde{u},\tilde{\theta}) ,
\end{align*}
where $\rho_0(\tilde{u},\tilde{\theta})$ is the random initialization distribution for $[\tilde{u},\tilde{\theta}]$, and in our case, it is chosen as $N(0,\sigma^2) \times N(0,\sigma^2)$.
Moreover, $\rho_0(\tilde{\theta})$ is the marginal random initialization distribution for $\tilde{\theta}$, which in our case is $N(0,\sigma^2)$.
With this choice, we note that $\int \tilde{u}\tilde{u}^\top d \rho_0(\tilde{u}|\tilde{\theta})= \sigma^2 I_{k \times k}$ is proportional to a diagonal matrix. Therefore, we may also replace the $k \times k$ matrix kernel $k_\infty^\theta(x,x')$ by the following scalar kernel:
\[
\sigma^2 \int \nabla_\theta h(\tilde{\theta},x)^\top \nabla_\theta h(\tilde{\theta},x) d \rho_0(\tilde{\theta}) .
\]

We may view the infinite-width NN as a kernel method in a small neighborhood around the initialization as:
\[
  f_\infty([\beta^u,\beta^\theta],x) = \sum_{i=1}^n \left[\beta_i^u k_\infty^u(x,x_i) + k_\infty^\theta(x,x_i)\beta_i^\theta \right] .
\]
This function gives an equivalent representation of the linear approximation of two-layer NN in \eqref{eq:lin-approx} with $m \to \infty$ and $\alpha \to \infty$. Therefore in this case, the optimization in $[u,\theta]$ can be equivalently represented using optimization in the kernel representation. 
The corresponding optimization problem using kernel representation can be written as:
\begin{equation}
\min_\beta \frac1n \sum_{i=1}^n L(f_\infty([\beta^u,\beta^\theta],x_i),y_i)  . \label{eq:opt-ntk}
\end{equation}
It is worth mentioning that in the NTK view of neural networks, we usually do not employ regularization $R(u,\theta)$. This is because the solution of NN with a nontrivial regularization will not lie in a small region around $[\tilde{u},\tilde{\theta}]$. 

In the NTK view, under appropriate conditions, it is possible to show that the optimization problem \eqref{eq:opt-ntk} in the kernel space is equivalent to the solution by SGD in the original representation \eqref{eq:twolayer-nn}, when $\alpha$ is large. In this case, the general SGD for two-layer NN in the original parameter space can be expressed as:
\begin{align*}
u \leftarrow &
u -\eta_u \nabla_u L(f([u,\theta],x_i),y_i) ,
\\
\theta \leftarrow &
\theta - \eta_\theta 
\nabla_\theta L(f([u,\theta],x_i),y_i) .
\end{align*}
If we consider using the same learning rate for the rescaled parameter
$\alpha u/m$ and $\theta$, as often done in practice, then we shall set
\begin{equation}
\frac{\alpha}{m} \eta_u = \eta,
\quad
\eta_\theta = \eta , \label{eq:ntk-lr-large}
\end{equation}
where $\eta$ is a small learning rate.
We note that in this case, the learning rate $\eta_u= O(\eta m/\alpha)$ will be large compared to $\eta_\theta$, if we set $\alpha=O(\sqrt{m})$ required by NTK.
The large learning rate $\eta_u$ will move $u$ to be far away from the initialization $\tilde{u}$, violating the standard NTK requirement of $u \approx \tilde{u}$. Nevertheless, we note that the requirement of $\theta \approx \tilde{\theta}$ is more important in NTK for linearly approximating the nonlinear function $h(\theta,x)$. With additional complexity, it is thus possible to extend the NTK analysis to handle the situation that $\theta \approx \tilde{\theta}$ but $u$ may not be close to $\tilde{u}$. 

Because of the above mentioned complexity, for the theoretical analysis of NTK, one often assumes a smaller learning rate for $u$ \cite{du2019gradient}, such as
\begin{equation}
{\alpha} \eta_u = \eta,
\quad
\eta_\theta = \eta , \label{eq:ntk-lr1}
\end{equation}
or as
\begin{equation}
\eta_u = \eta,
\quad
\eta_\theta = \eta . \label{eq:ntk-lr2}
\end{equation}


For the learning rates in \eqref{eq:ntk-lr1},
since $\eta_u$ is much smaller than
$\eta_\theta$, we know that $u$ moves very little compared to $\theta$. Consequently, we can see from \eqref{eq:ntk-param} that $\beta^u$ moves very little compared to $\beta^\theta$. Therefore the kernel $k_\infty^\theta(\cdot,\cdot)$ is effective, while the kernel $k_\infty^u(\cdot,\cdot)$ is not effective. 
The learning rate \eqref{eq:ntk-lr2} does not suffer from this problem. From \eqref{eq:ntk-param}, 
it can be seen that the corresponding modification of both $\beta^u$ and $\beta^\theta$ are at the same order. 
Therefore in such case, both kernels are effective.

\begin{figure}
    \centering
    \includegraphics[width=0.5\textwidth]{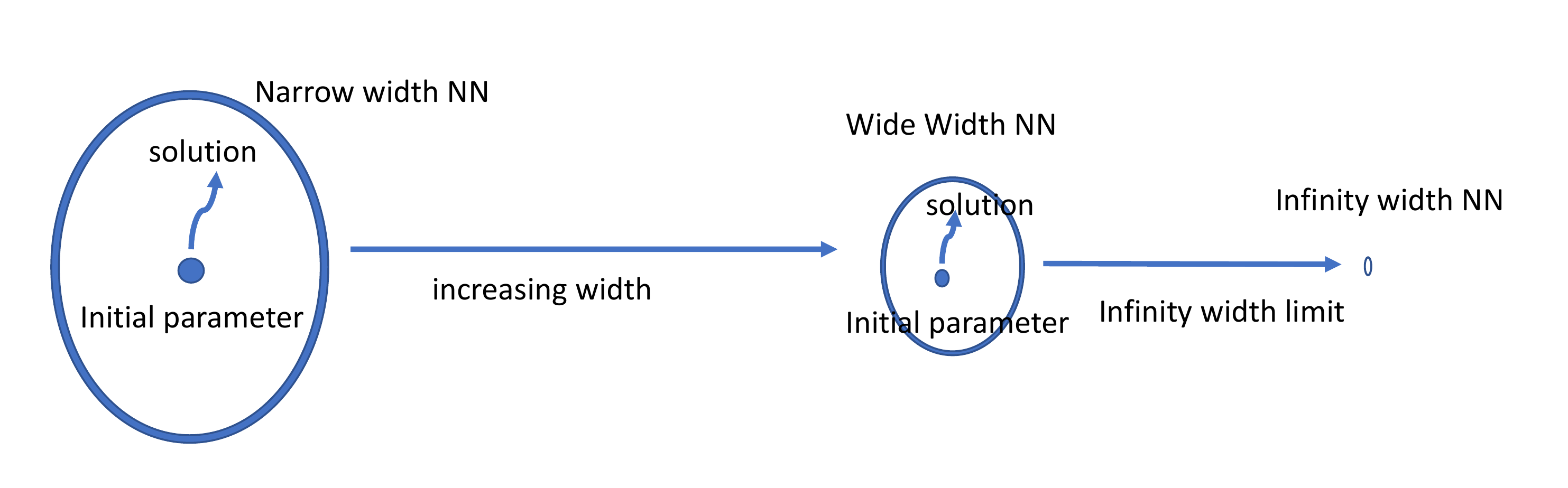}
    \caption{Solution neighborhood size versus NN width $m$}
    \label{fig:ntk}
\end{figure}

It can be shown that with learning rates set as either \eqref{eq:ntk-lr1} or \eqref{eq:ntk-lr2}, when $m \to \infty$ and $\alpha \to \infty$, the final solution of \eqref{eq:opt} without regularization 
can reach zero-error within a very small neighborhood of the initialization,
with radius approaching zero as $\alpha \to \infty$ (e.g. \cite{du2018gradient}). This phenomenon is illustrated in
Figure~\ref{fig:ntk}.  This regime is called the NTK regime, where the two-layer NN can be linearized as a kernel method, and the optimization process lies inside a small neighborhood around the initialization.

\begin{figure*}
	\centering
	\includegraphics[width=0.45\textwidth]{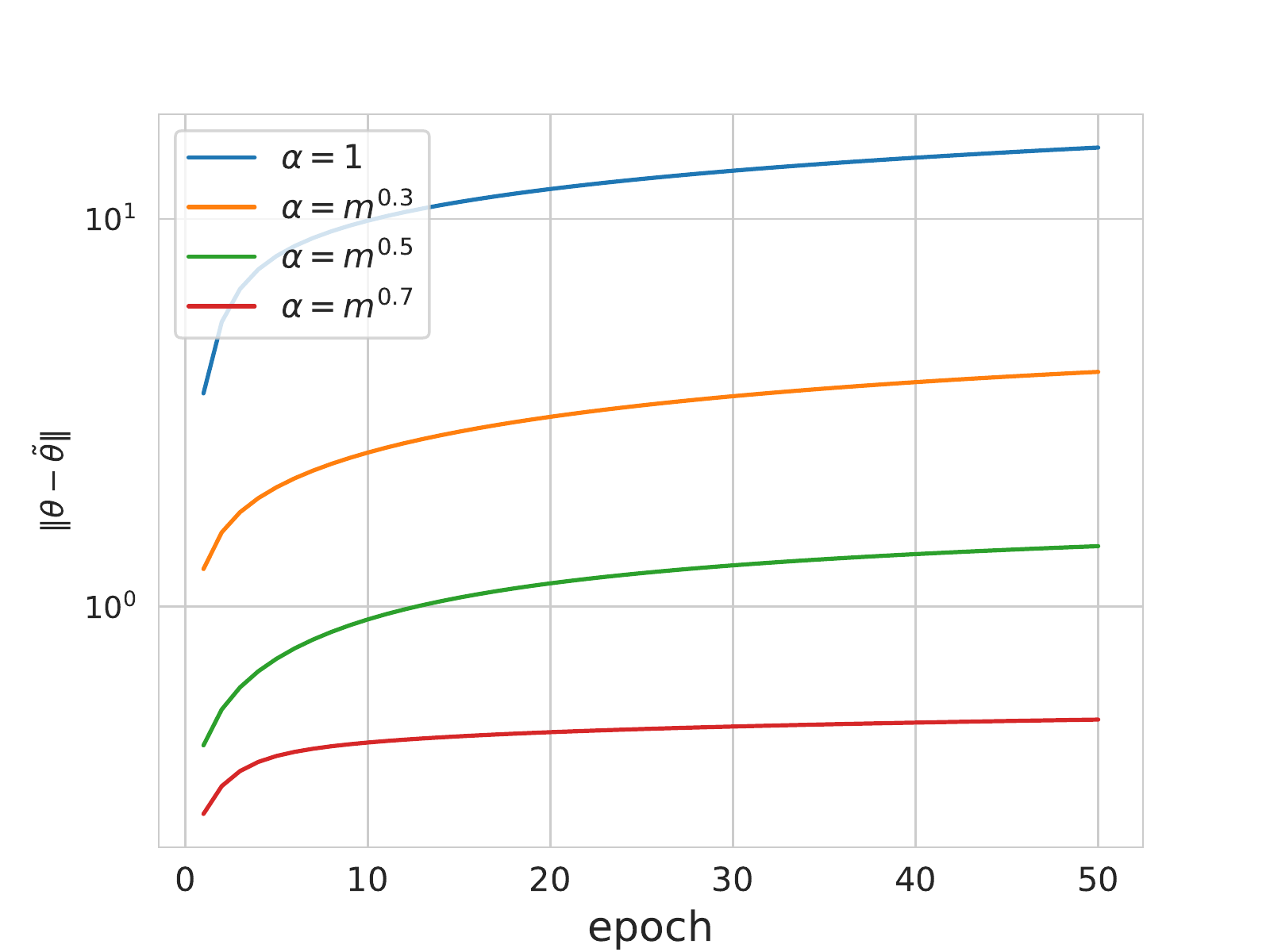}
\includegraphics[width=0.45\textwidth]{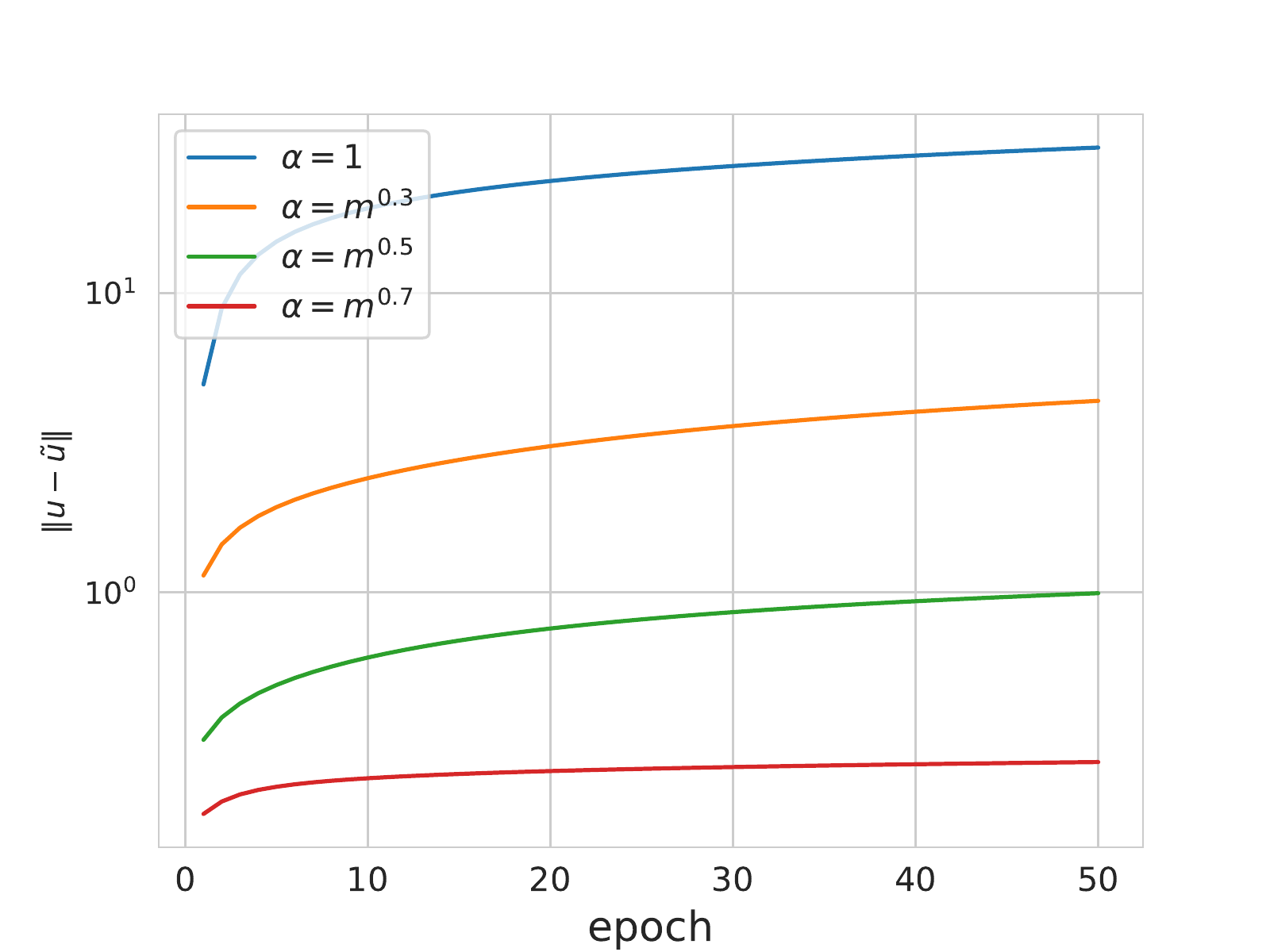}
	\caption{Impact of $\alpha$ for NN optimization ($\eta_u/\eta_\theta = 1, m = 10^4$) }
	\label{fig:alpha}
\end{figure*}

This property can also be validated empirically in the actual NN optimization process. In this paper, we use the MNIST handwritten digits dataset available from \url{http://yann.lecun.com/exdb/mnist/} \cite{lecun1998gradient} to demonstrate various aspects of the different frameworks. This dataset has a training set of 60,000 examples, and a test set of 10,000 examples, and is one of the standard dataset for image classification.  
The theory of NTK implies that when the scaling parameter $\alpha$ increases, the NN training process belongs to a smaller and smaller neighborhood of the initialization, and NTK approximation becomes more and more accurate. This phenomenon is shown in Figure~\ref{fig:alpha}, where the average distances between $\theta$ (and $u$) and the initialization are plotted for different values of $\alpha$. 

\begin{figure*}
    \centering
    
    \begin{tabular}{cc}
        \includegraphics[width=0.45\textwidth]{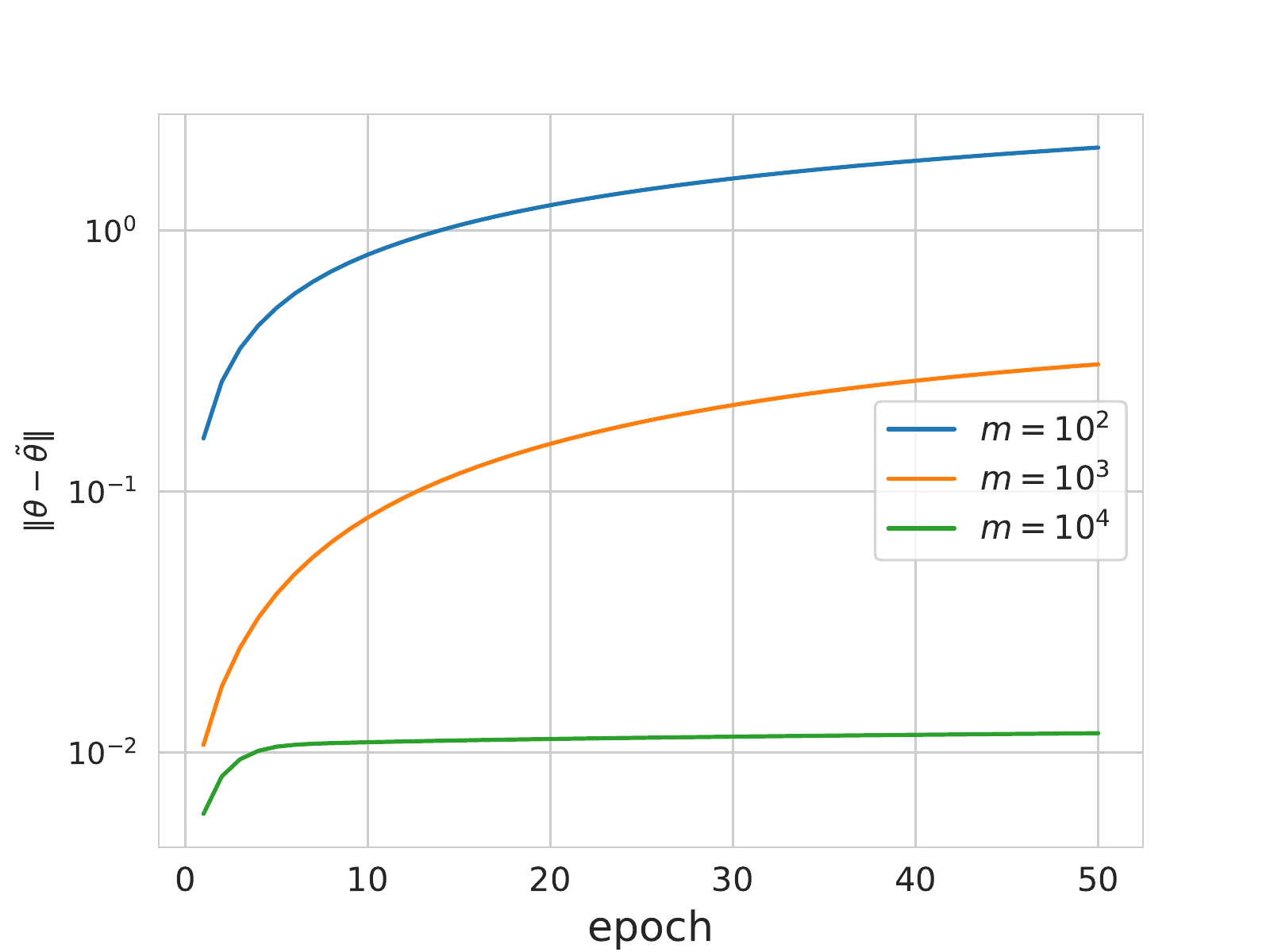} & \includegraphics[width=0.45\textwidth]{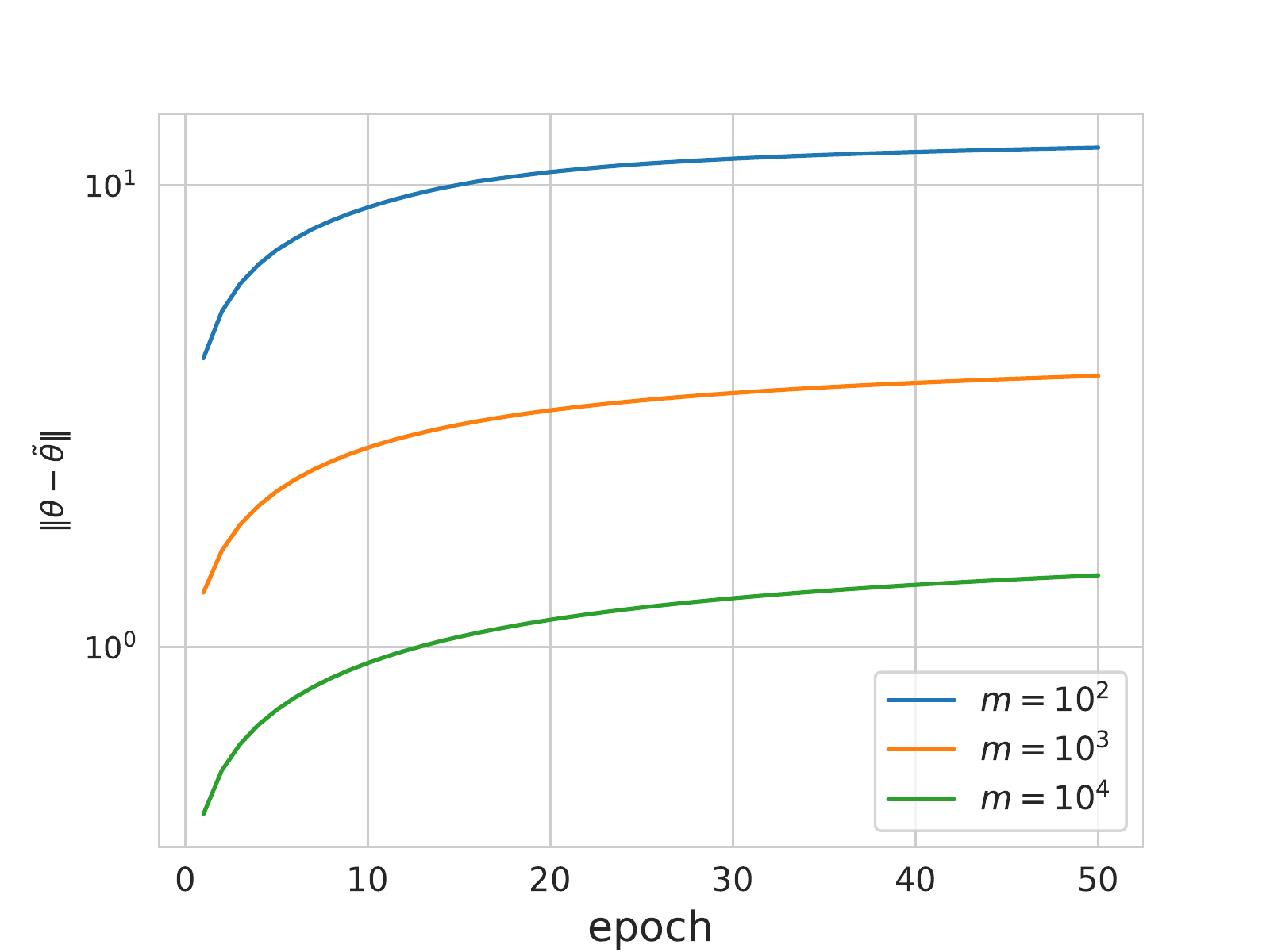} \\
        \small{(a) $\eta_u/\eta_\theta=m/\alpha$} & \small{(b) $\eta_u/\eta_\theta=1$}
    \end{tabular}

    \caption{Impact of $m$ for NN optimization ($\alpha = \sqrt m$)}
    \label{fig:ntk_m}
\end{figure*}

Another factor that determines the accuracy of NTK approximation is the number of hidden units $m$ which measures the degree of overparameterization. Figure~\ref{fig:ntk_m} shows that when $m$ increases, the solution of the neural networks becomes closer to the initialization, and this phenomenon happens both with the practical learning rate $\eta_u/\eta_\theta=m/\alpha$ and with the NTK theoretical learning rate $\eta_u/\eta_\theta=1$. From this experiment, it is reasonable to speculate that as $m$ approaches infinity, the first-order approximation of NN (NTK) will become more reliable.  

\begin{figure*}
	\centering
	\includegraphics[width=0.45\textwidth]{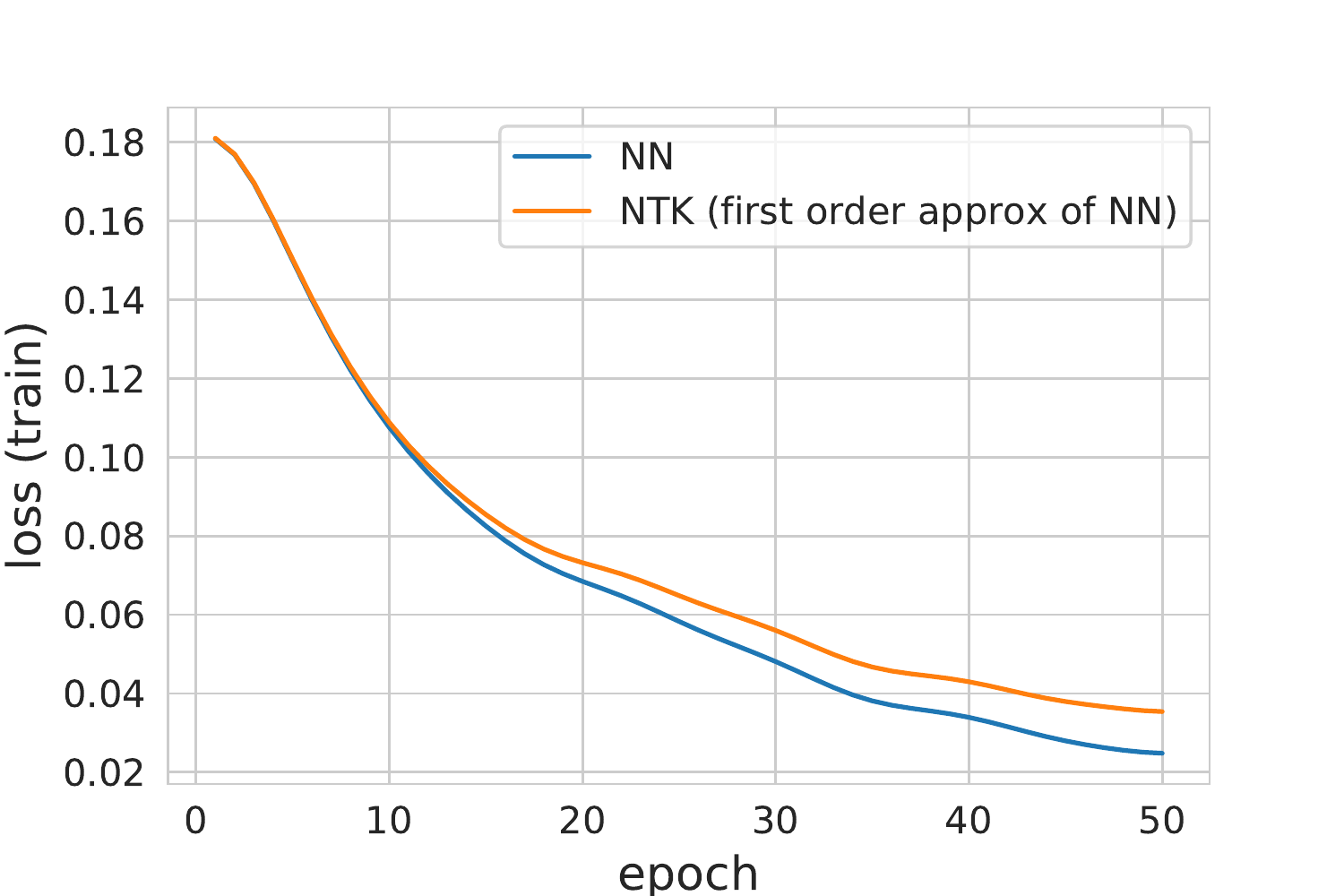}
	\includegraphics[width=0.45\textwidth]{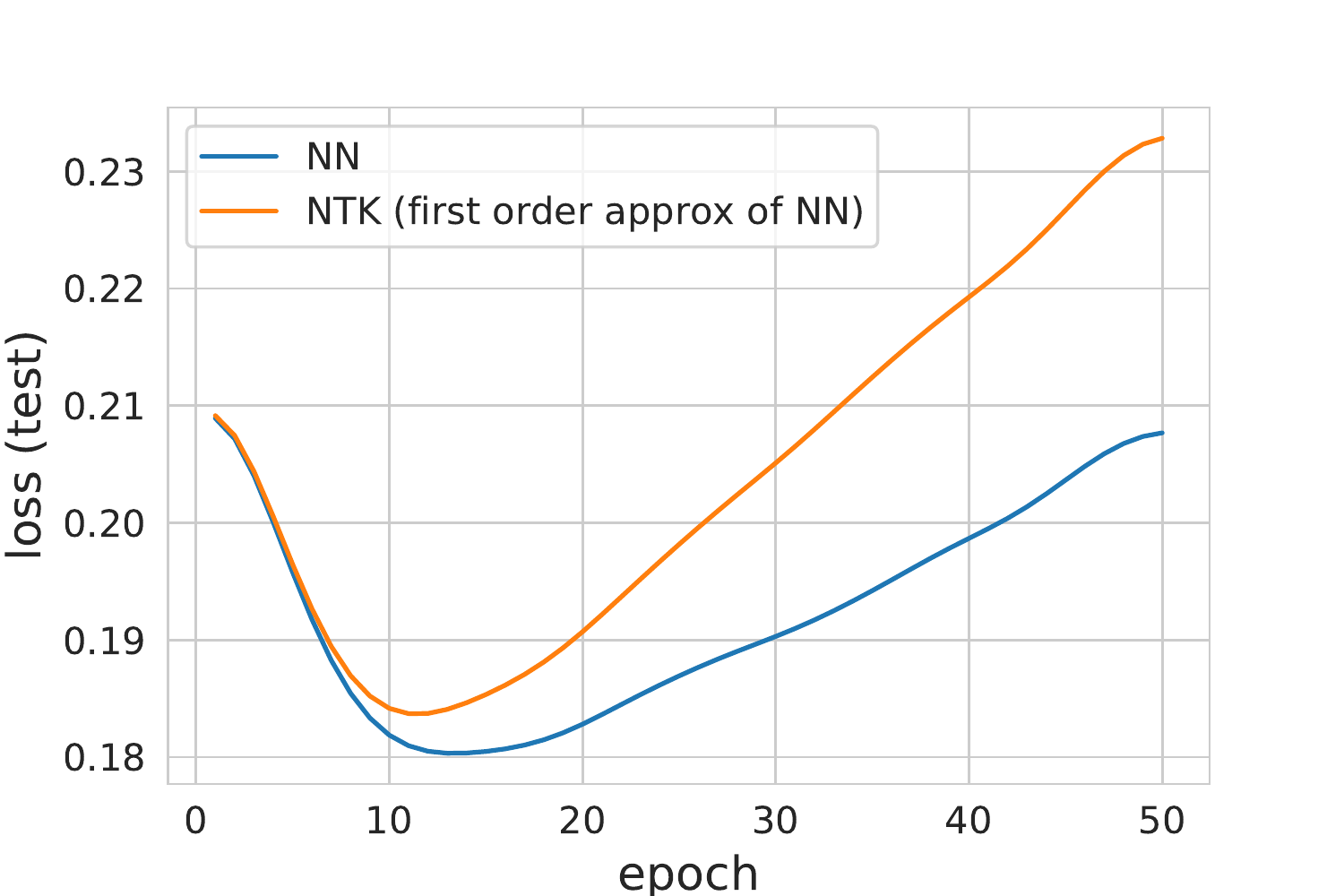}
	\includegraphics[width=0.45\textwidth]{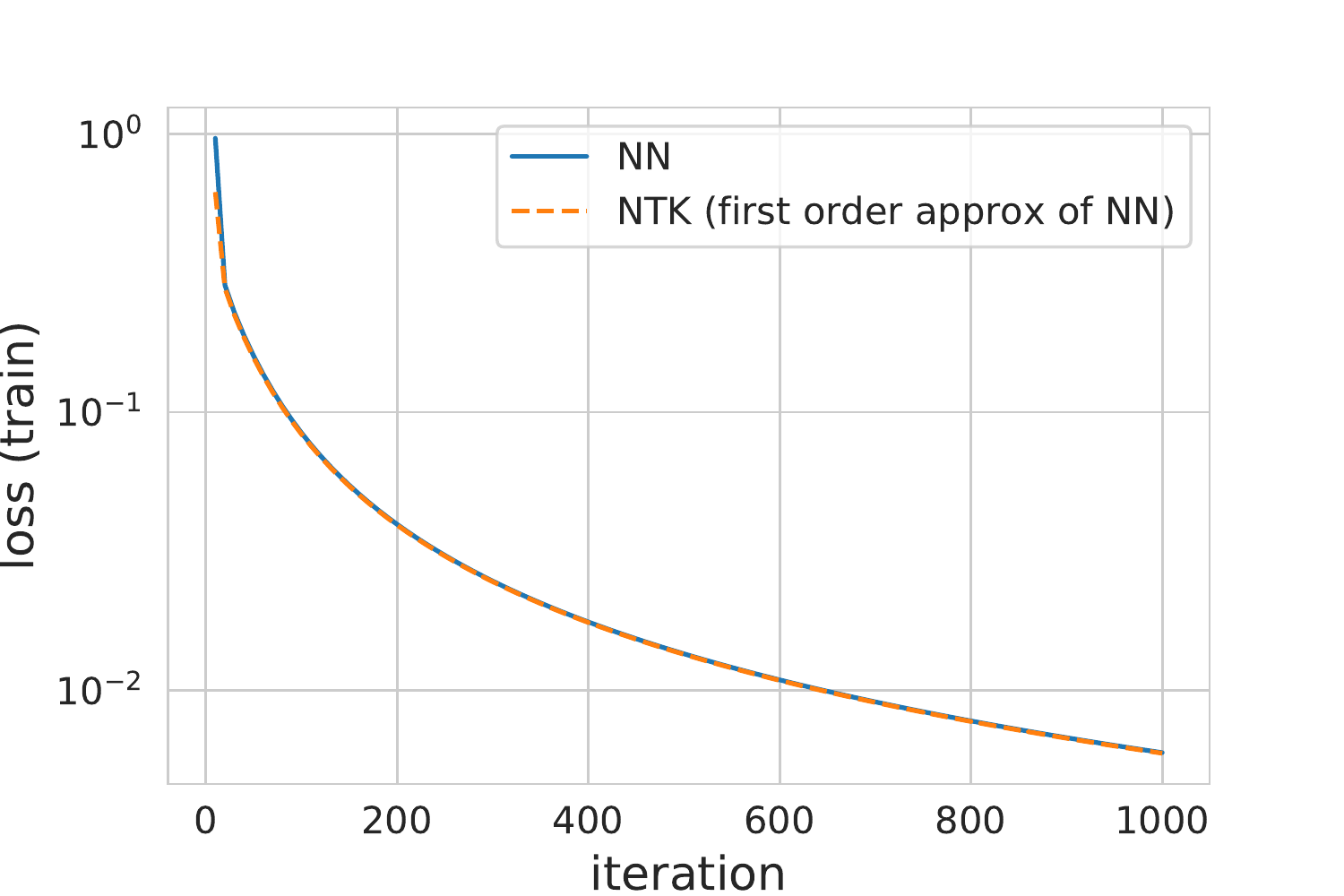}
\includegraphics[width=0.45\textwidth]{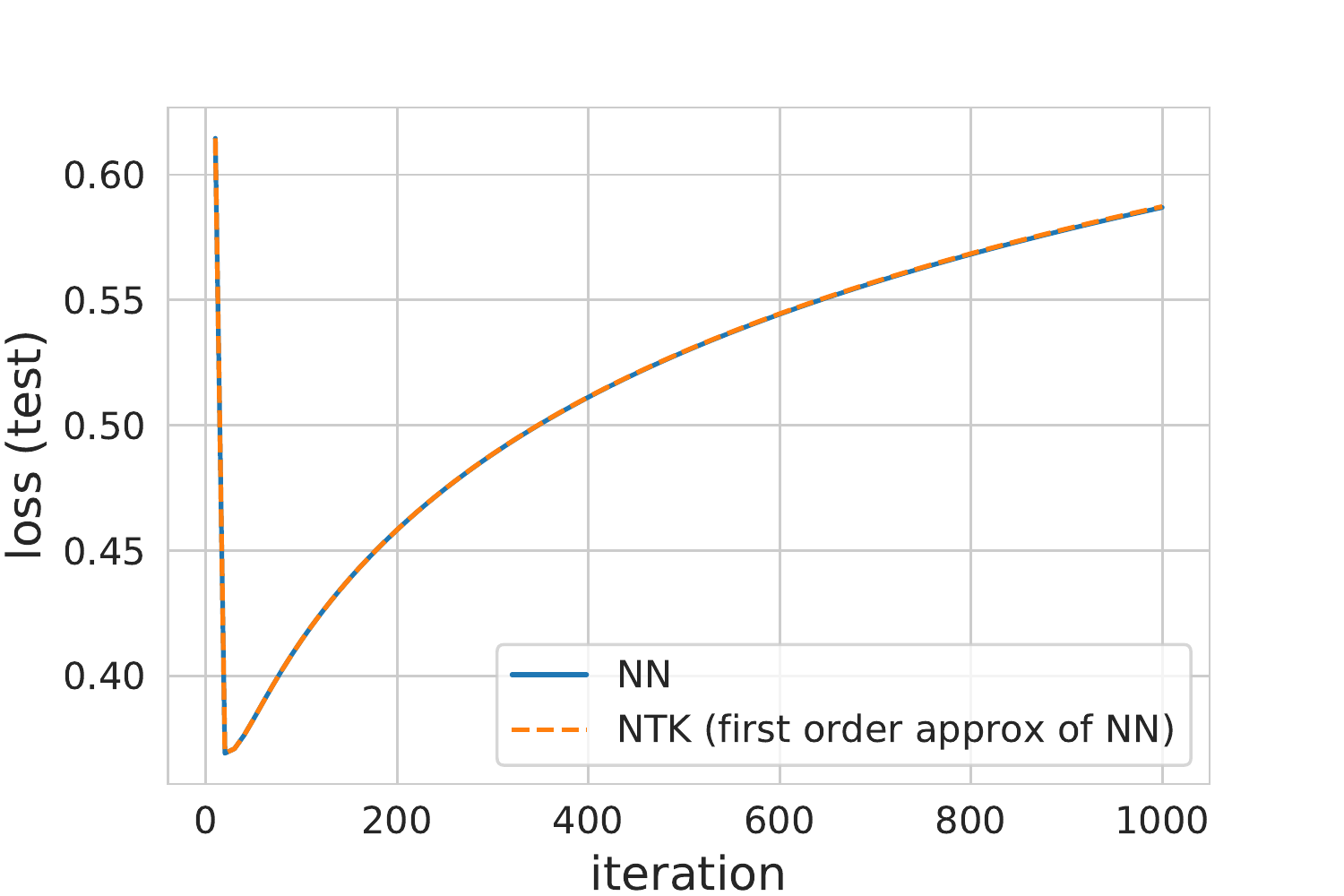}
	\caption{Linearization of NN (First Row: MNIST; Second Row: synthetic data)}
	\label{fig:linear}
\end{figure*}

It is also useful to point out that NTK approximation works better on simple datasets, and fails more easily on complex datasets. Figure~\ref{fig:linear} compares the objective functions of the NN (both on training and on test data) to its linearized NTK approximations during the training process.  We compare the linearization error both on the MNIST dataset and on a simpler 100-dimension synthetic dataset made by  \emph{make\_classification} in sklearn. For an NN with $m=1000$ and $\eta = 0.1$, we can notice that the linearization of NN on the simpler dataset is almost perfect, which means that the entire training process is well-approximated by NTK. However, there is a noticeable discrepancy between NTK and the actual NN on the MNIST data, which means that NTK does not fully explain the actual NN training process well in this case. This experiment implies that  for relatively complex datasets, the NTK approximation requires a much wider NN. 

\begin{figure*}
	\centering
    \begin{tabular}{ccc}
        \includegraphics[width=0.30\textwidth]{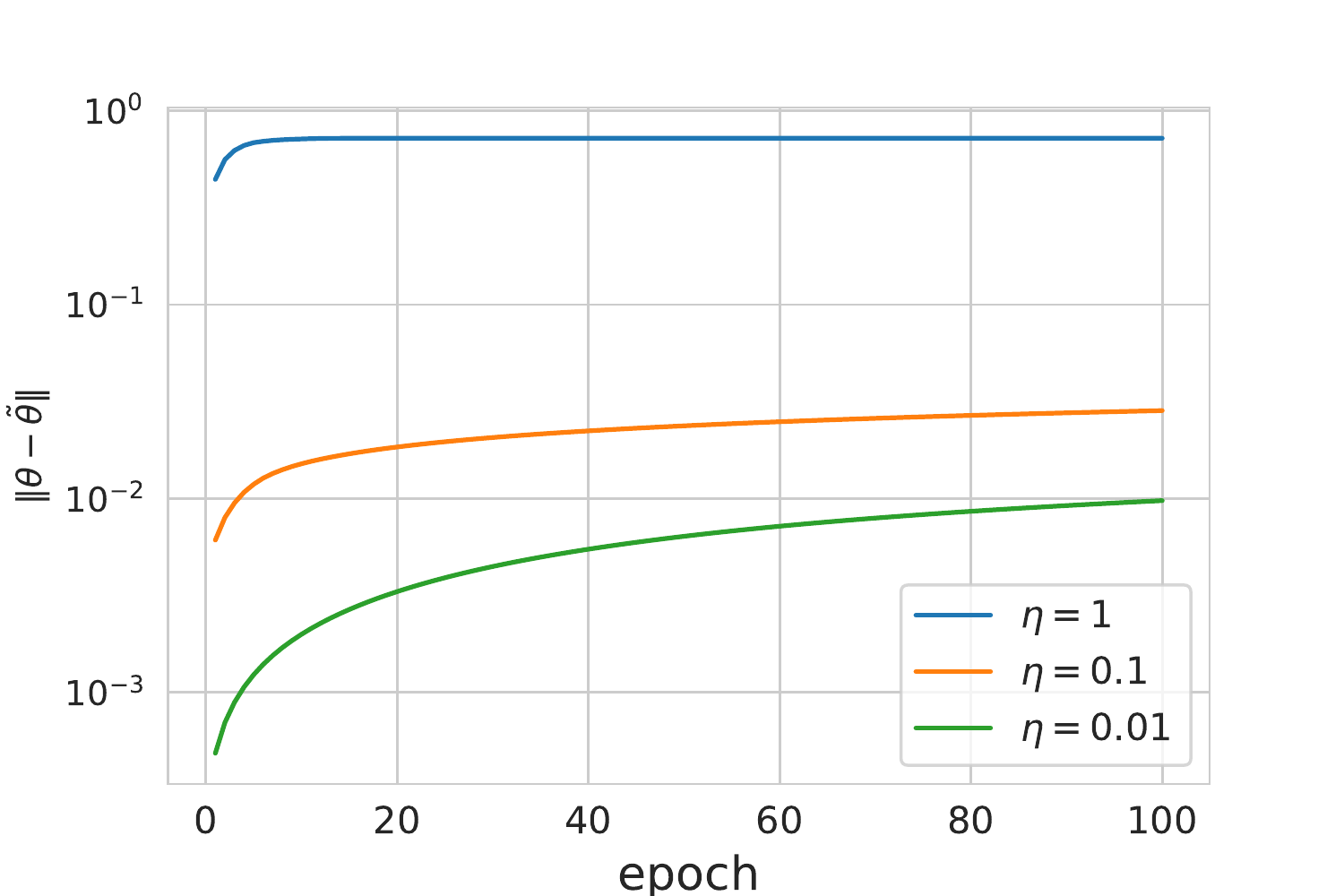} & \includegraphics[width=0.30\textwidth]{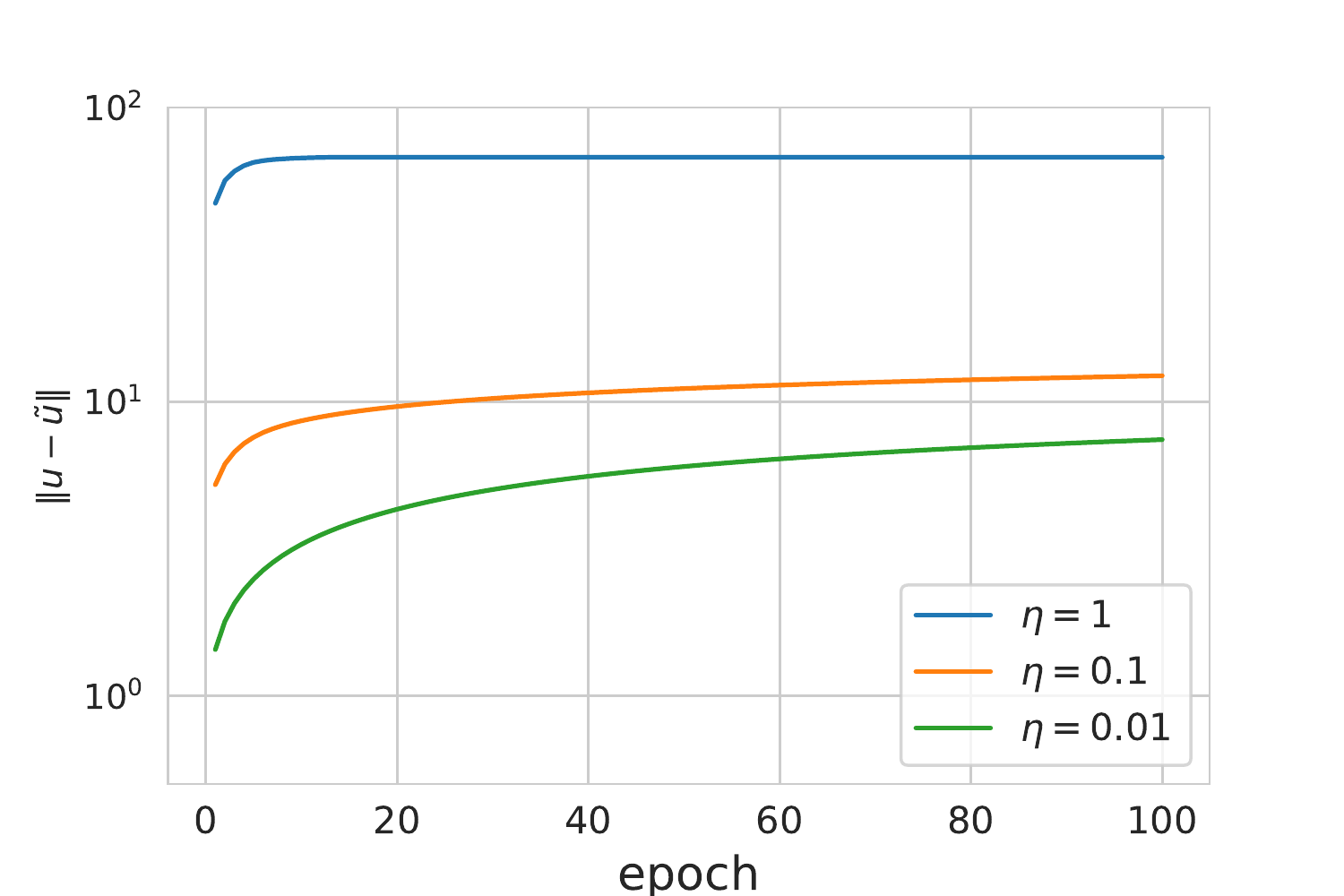}&
        \includegraphics[width=0.30\textwidth]{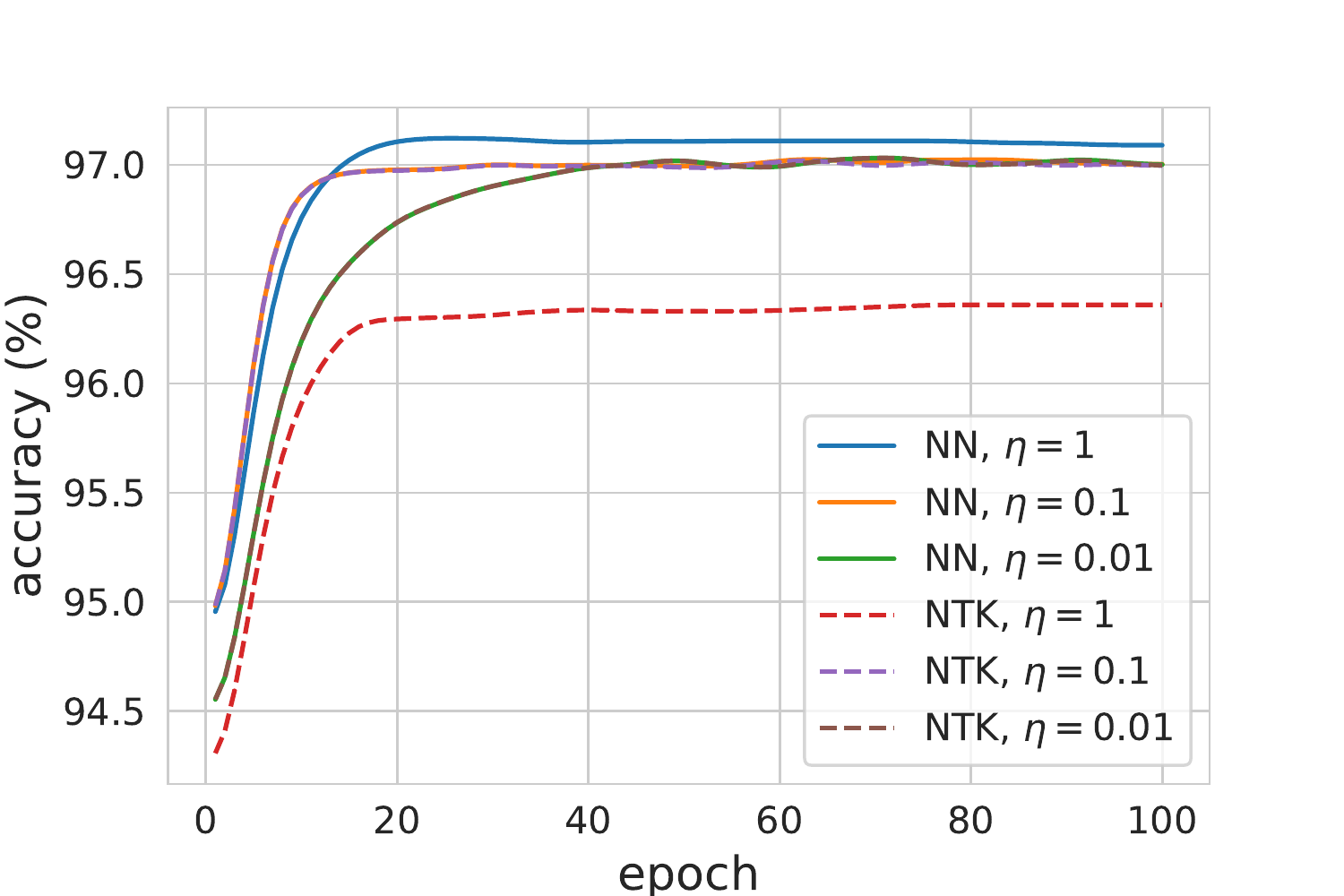}\\

        \includegraphics[width=0.30\textwidth]{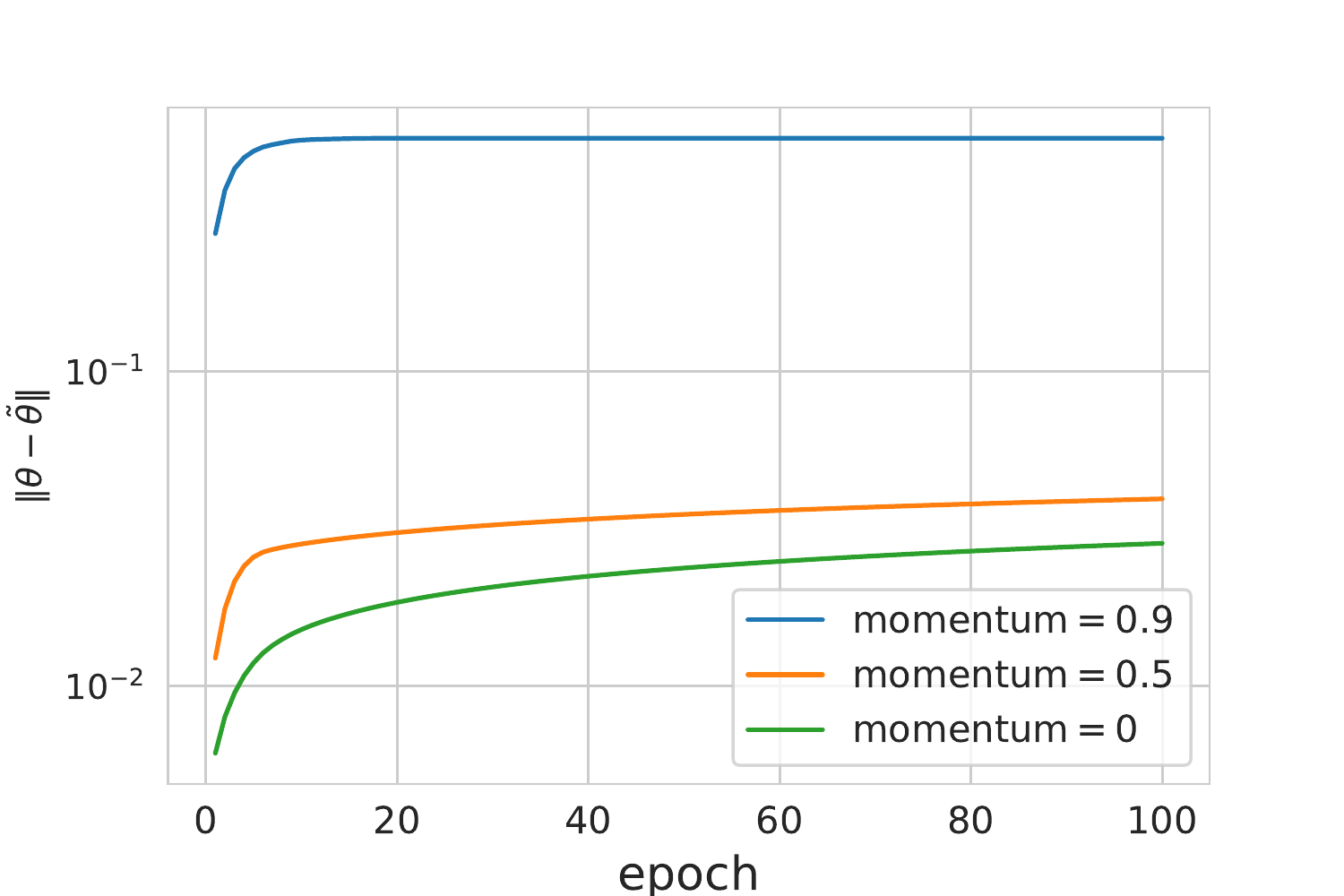}&
        \includegraphics[width=0.30\textwidth]{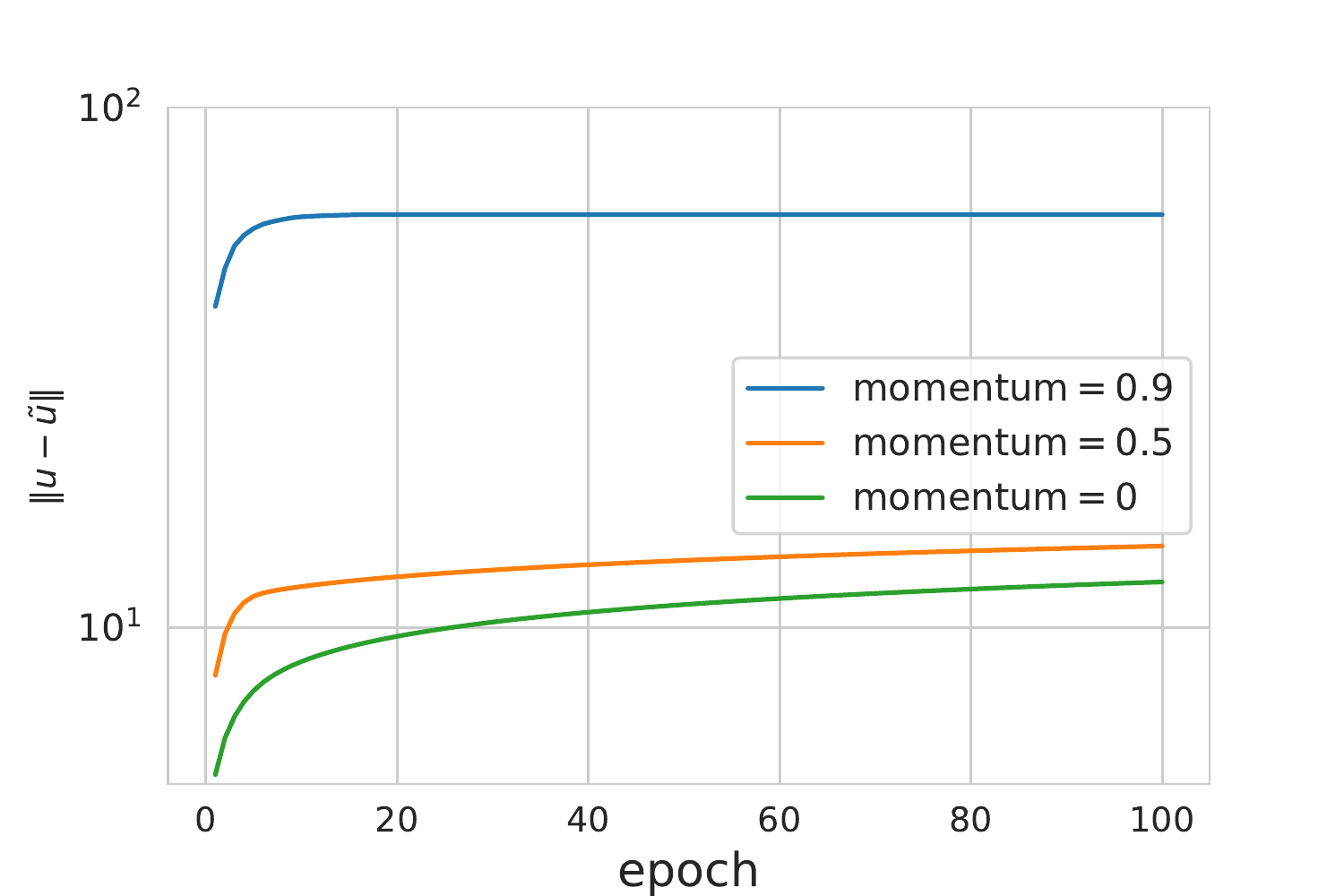}&
        \includegraphics[width=0.30\textwidth]{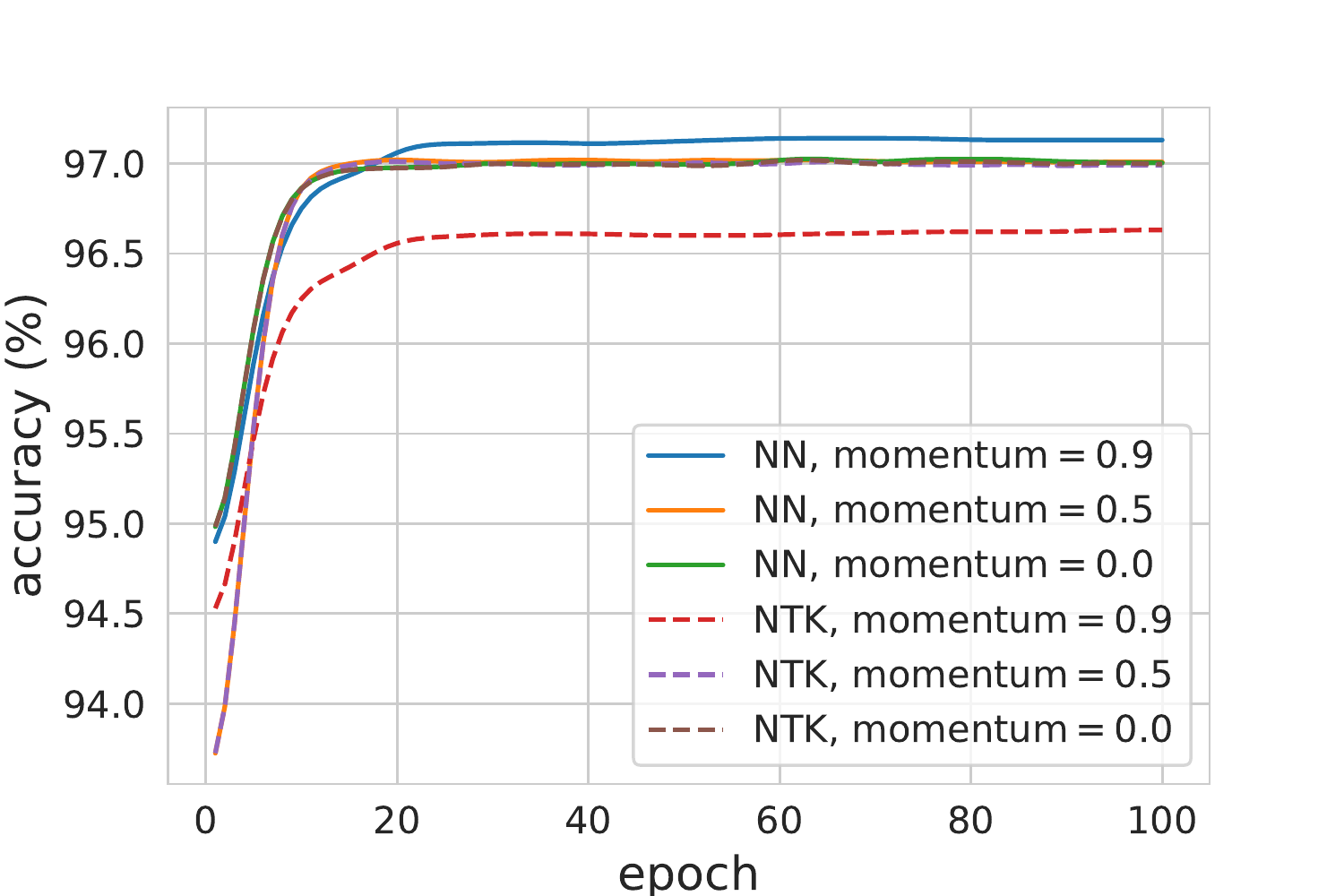}\\
        \includegraphics[width=0.30\textwidth]{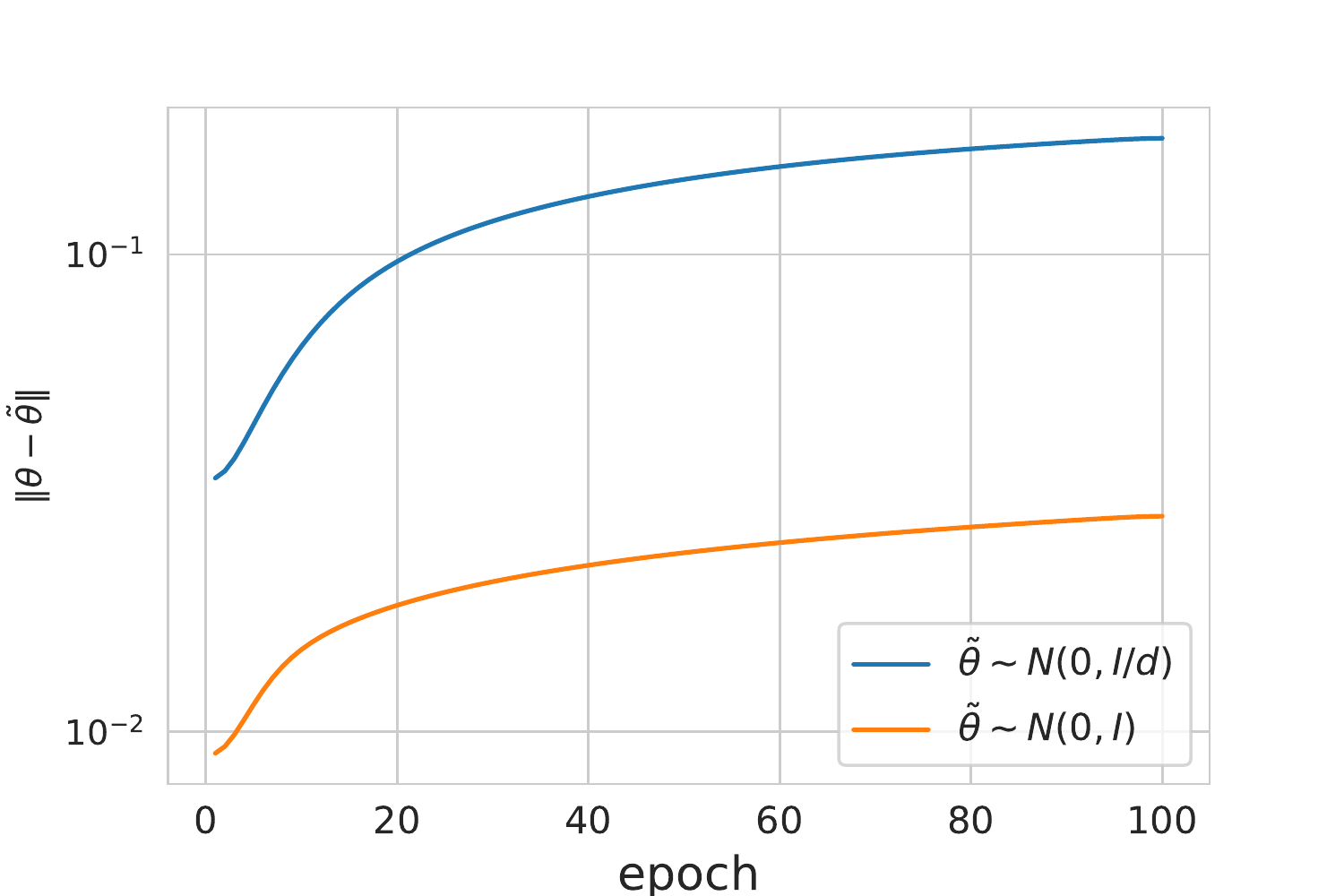}&
        \includegraphics[width=0.30\textwidth]{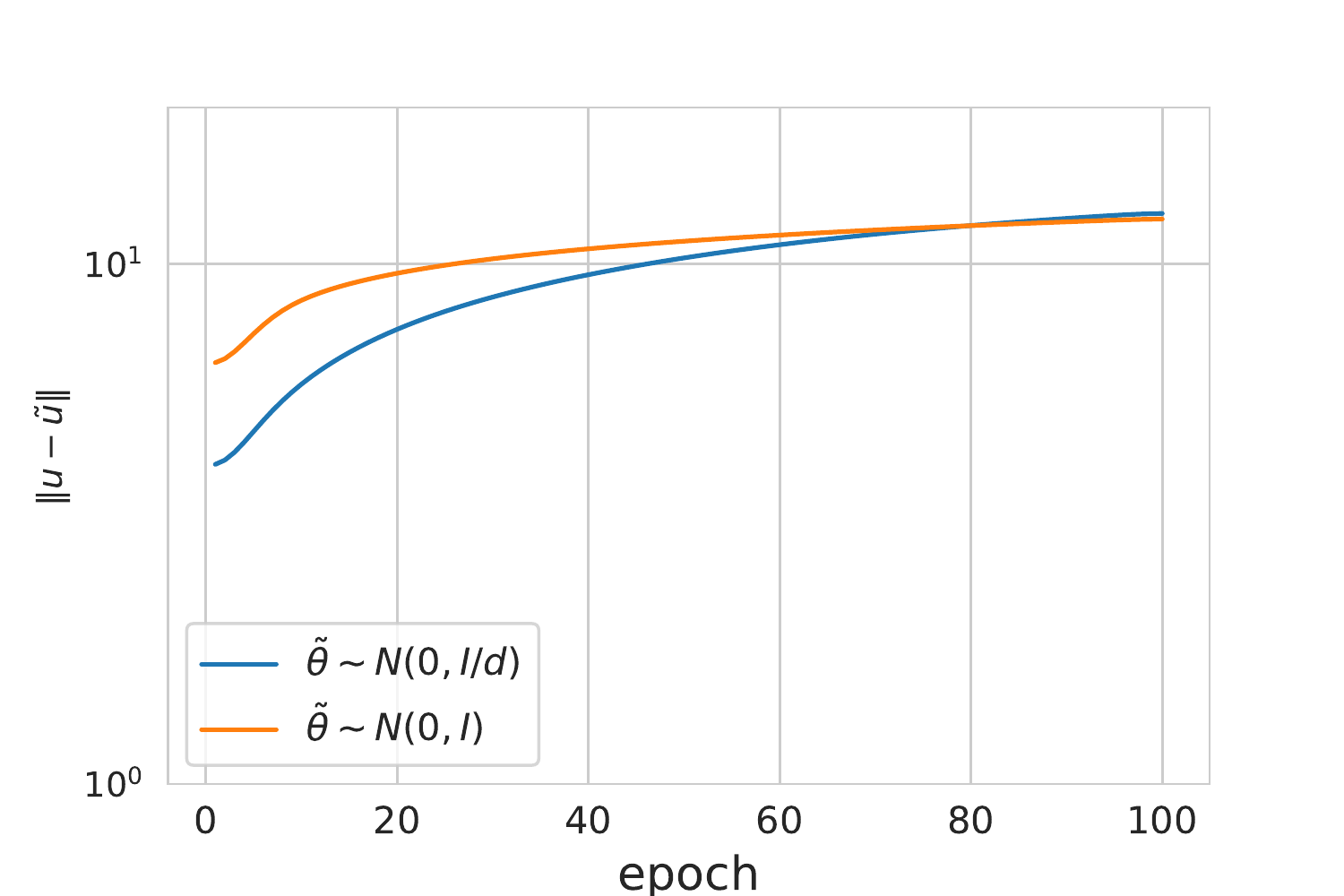}&
        \includegraphics[width=0.30\textwidth]{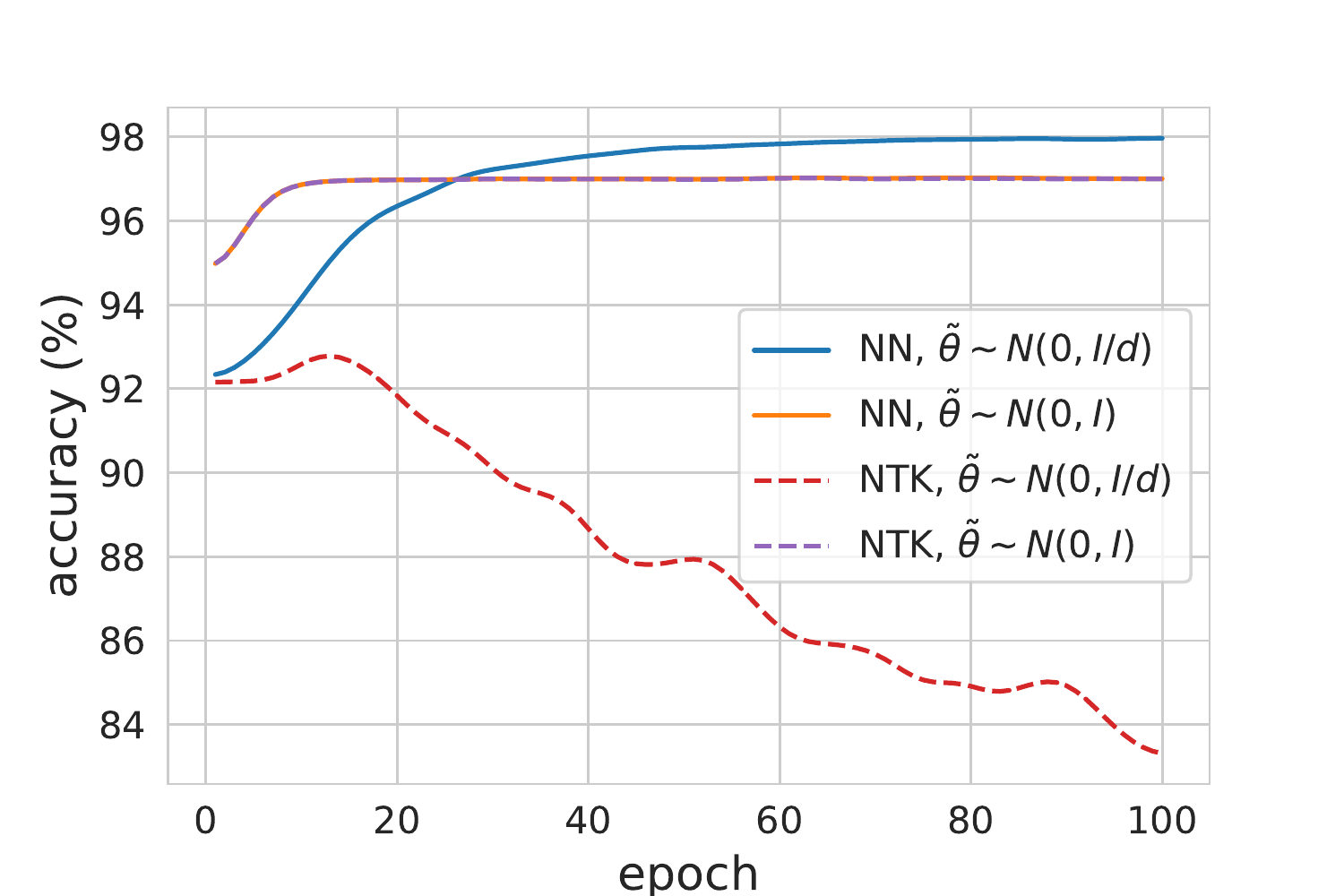}
        
        \\
        \footnotesize{(a) change of $\theta$} & \footnotesize{(b)   change of $u$}& \footnotesize{(c) NN and NTK performance} \\
    \end{tabular}
    
	\caption{Factors to break neural tangent kernel regime ($m=5000$)}
	\label{fig:break_ntk}
\end{figure*}


In summary, when $\alpha \to \infty$ and $m \to \infty$, and the formulation does not contain regularization, then
we have the so-called NTK regime, with the following properties:
\begin{itemize}
\item Initialize NN with a certain scaling
\item Network is sufficiently large
\item The formulation does not contain regularization
\item Learning rate is sufficiently small
\item NN solution path remain close to the initialization, and can be linearized around the initialization
\item The linearization induces a kernel (NTK) which is convex
\end{itemize}
The implications of the NTK view is as follows:
\begin{itemize}
\item The solution is in a tiny neighborhood of the initial point
\item Problem becomes convex, which can be solved efficiently
\item The solution path is reproducible in kernel representation
\end{itemize}

The theory of NTK applies to a specialized regime with special initialization.
It also assumes small learning rate so that the learned parameter does not escape from the NTK region. Although this is a nice mathematical model, there are several problems, making it unsuitable for a general theory of neural networks.

\fangcong{ Note that  RF can be considered as a two-layer NN where the bottom layer is fixed.  The model is linear with respect to the top-layer, and one can incorporate regularization to improve generalization. In contrast, in the NTK regime, we perform GD  to optimize NN weights in all layers, with random features generated in the tangent space around the initialization. In order to ensure that the solution of \eqref{eq:opt} can be approximated by \eqref{eq:opt-ntk}, the resulting optimization problem \eqref{eq:opt-ntk} of NTK cannot include non-trivial regularizers that will pull the training process out of the tangent space.} 
It is natural to extend \eqref{eq:opt-ntk} by adding regularization, where we may consider the following more general formulation of NTK in \eqref{eq:opt-ntk} to the following regularized NTK method,  which may be regarded as an NTK motivated new learning algorithm, although it may not be a good approximation for two-layer NNs anymore. 
\begin{equation}
\min_\beta \frac1n \sum_{i=1}^n L(f_\infty([\beta^u,\beta^\theta],x_i),y_i)+  R([\beta^u,\beta^\theta]),\label{eq:opt-ntk-reg}
\end{equation}
where
\begin{align*}
R([\beta^u,\beta^\theta])=&
\frac{\lambda}{2} \sum_{i=1}^n\sum_{\ell=1}^n \left[((\beta_i^u)^\top \beta_\ell^u)k_\infty^u(x_i,x_\ell)\right.\\
&\left.+ (\beta_i^\theta)^\top k_\infty^\theta(x_i,x_\ell)\beta_\ell^\theta
\right] .
\end{align*}
Although this regularized formulation is natural in the kernel learning setting, it is not equivalent to two-layer NN due to the addition of the regularization term.

A major problem of the NTK view is that the practical performance of the NTK solution from \eqref{eq:opt-ntk-reg} is often inferior to that of the fully trained NNs, despite the equivalence which can be proved under certain theoretical assumptions. Even an infinitely wide NTK cannot achieve state-of-the-art performance.
In the following, we explain why this happens in practice and what can break the NTK view for neural network learning. We show that the NTK regime is broken by standard tricks in NN learning.

Although the random parameter initialization with large scaling $\alpha$ is used by practitioners, and the choice is consistent with that required by the NTK view point, practitioners use a large initial learning rate 
while the small learning rates is required by the theoretical analysis. Consequently the optimized NN by using practical SGD procedures goes out of the NTK regime. Because of this, the NTK linear approximation fails. 

Moreover, in practice, neural networks are not infinite-width, and thus the large $m$ theory does not exactly match the practical behavior of neural network learning. Another theoretical condition for the NTK view is to not impose nontrivial regularization\footnote{\fangcong{We note that the NTK regime can still incorporate very small regularizers. For example, one can add a regularizer as $\mu(\| \theta\|^2+\|u\|^2)$, where $\mu$ is much smaller than $\frac{1}{m}$. In this case,  there is still a solution in the neighborhood of the initialization. Moreover, we may add regularizers centered at the initialization $\tilde{\theta}$ such as $\|\theta-\tilde{\theta}\|_2^2$, although they are not used by practitioners.}} because regularization  automatically pushes the solution away from the initialization, which violates the NTK regime. However, regularization (or weight decay) is frequently used by practitioners. \fangcong{Besides, there is currently no analysis of NNs that can incorporate batch-normalization in the NTK regime.}

One of the key reasons that NTK does not perform well in practice is that the NTK method is very similar to RF, in that it also employs random features. Compared to RF of Section~\ref{sec:random}, NTK contains an additional kernel corresponding to the random features $\tilde{u}\nabla_\theta h(\tilde{\theta},x)$. In fact, the mathematical theory of NTK relies mostly on the modification of $\theta$ associated with random features $\tilde{u}\nabla_\theta h(\tilde{\theta},x)$ to reduce training loss, while the mathematical theory of RF relies on the modification of $u$ associated with random features $h(\tilde{\theta},x)$ to reduce training loss. This is why the theoretical analysis of NTK relies on a small learning rate $\eta_u$. Nevertheless, the additional random features used by NTK provides extra information over RF. 

Since NTK is still a random feature based method, it does not learn feature representations.
In contrast, it is well-known by practitioners  that a key benefit of NN learning is the ability to learn useful feature representations.
The theory of NTK completely fails to explain the benefit of feature learning by neural networks. We will investigate this issue further in Section~\ref{sec:exp}.

The NTK view is also inconsistent with many technical tricks used in practical NN training, which benefit learning performance, such as large initial learning rate and momentum, which may push the model parameters out of the initialization neighborhood. Here we show some cases in real practice to demonstrate these phenomenons.

As shown in Figure~\ref{fig:break_ntk}, when we use a large learning rate or employ momentum, the difference between the initial parameters and the final solution increases. In such situation, the first order approximation \eqref{eq:first-order} used by NTK fails to capture the dynamics of NN. The gap between NN and NTK is significant, and the performance of NN is better with these tricks. A few other works have also mentioned the phenomenon that a large initial learning rate leads to better NN solution \cite{li2019towards,jastrzkebski2017three}, which cannot be explained by the NTK view very well.

Note that most of our NTK experiments employ the random initialization $\tilde \theta_j\sim N(0,1)$. In the analysis, the variance $\sigma^2$ is fixed as a constant, which does not influence asymptotic behavior significantly. However, the actual value of $\sigma^2$ plays a noticeable role when $m$ is not large enough  compared to the input dimension $d$. Figure~\ref{fig:break_ntk} shows that when we use $\tilde \theta_j\sim N(0,1/d) $ (which is a standard initialization technique with good practical performance, referred as He initialization \cite{he2015delving}), the NTK regime can be broken more easily.  Therefore, the gap between the first order approximation of NTK and actual NN training cannot be ignored in practice.



\section{Mean Field View}\label{sec:meanfield}

In order to overcome the limitations of the NTK view which we explained above, other theoretical models have been developed to investigate overparameterized two-layer neural networks. In this section, we introduce another line of research, which applies the mathematical tools of the mean-field analysis from statistical physics to study two-layer neural networks
\cite{meie7665,chizat2018global,sirignano2019mean, sirignano2020mean, rotskoff2018neural, dou2020training,  wei2018margin,dou2020training,hu2019mean,zhang2020can}.

In order to motivate the mean field analysis for overparameterized NNs, it is instructive to first investigate the continuous dynamics of infinitely wide NNs, known as the {\em mean field limit},  and then consider the finite width neural networks as its approximation. In this paper, we call the corresponding analysis the mean field view (MF).
The idea of studying the mean filed limit comes from statistical physics \cite{engel2001statistical}, which suggests that the mathematical model of a large number of interacting neurons can be simplified using the probability distribution that represents the average effect.

Unlike the NTK view, which requires $\alpha \to \infty$ as $m \to \infty$, in the mean field view \cite{meie7665}, we may set the scaling fixed at a constant such as $\alpha=1$ while letting $m \to \infty$. In this case,  the solution is allowed to go far from the initialization,
which remedy the main limitations of NTK. Therefore one may argue that this approach gives a more realistic mathematical model for practical behaviors of neural networks. 

In the limit of $m \to \infty$ with fixed $\alpha$, we may consider the continuous limit of two-layer neural networks in \eqref{eq:twolayer-nn} as
\begin{equation}
f(\rho,x) = \int \alpha u h(\theta,x) d \rho(u,\theta) , \label{eq:twolayer-nn-cont}
\end{equation}
where $\rho(u,\theta)$ is a probability distribution over $[u,\theta]$. In this continuous formulation, we can regard the probability
measure $\rho$ as the model parameter.
Therefore the original model parameters  $\{[u_j,\theta_j]\}$ in the discrete NN formulation can be viewed as a discrete probability distribution on the model parameter space $[u,\theta] \in \real^k \times \real^d$, and this discrete probability distribution puts a mass of $1/m$ at each point $[u_j,\theta_j]$ ($j=1,\ldots,m$). In the continuous limit of $m \to \infty$, 
this discrete probability distribution naturally converges to the distribution parameter
$\rho$ in the continuous NN formulation \eqref{eq:twolayer-nn-cont}. It is easy to see that the function represented by two-layer NN becomes linear in $\rho$.

In the continuous limit, the training objective \eqref{eq:opt} for the discrete NN becomes
\begin{align}
  \phi(\rho) =& \frac1n \sum_{i=1}^n L(f(\rho,x_i),y_i) + R(\rho) , \label{eq:opt-cont}\\
  R(\rho) =& \int r(u,\theta) d \rho(u,\theta) \nonumber
\end{align}
for the continuous NN, where $r(u,\theta)$ is a regularizer of $[u,\theta]$ such as the $L_2$ regularization
\[
r(u,\theta) = 
\frac{\lambda_u}{2} \|u\|_2^2
+ \frac{\lambda_\theta}{2} \|\theta\|_2^2 .
\]
It follows that the training objective \eqref{eq:opt-cont}
is convex with respect to $\rho$ if both the loss function $L(\cdot)$, and the regularizer $R(\cdot)$ are convex.
In fact, the global optimal solution of $\rho$ satisfies the first order optimality condition:  for
all probability measures $\rho'(u,\theta)$:
\begin{equation}
 \int g(\rho,u,\theta) d \rho'(u,\theta) \geq \int g(\rho,u,\theta) d \rho(u,\theta) , \label{eq:first-order0}
\end{equation}
where
\begin{equation}
g(\rho, u,\theta)= \frac1n \sum_{i=1}^n \alpha L_1'(f(\rho,x_i),y_i) u h(\theta,x_i) + r(u,\theta) \label{eq:grad}
\end{equation}
is the derivative of $\phi(\rho)$ with respect to the component $\rho(u,\theta)$ by regarding the distribution $\rho$ as an infinite dimensional vector
$\rho=\{\rho(u,\theta)\}$.
Here $L_1'(v,y)= \nabla_v L(v,y)$.
Note that if we can find $\rho$ such that
\begin{equation}
g(\rho,u,\theta)=c \label{eq:first-order}
\end{equation}
for a constant $c$, then
\eqref{eq:first-order0} is satisfied. This is because in this case, we have for all $\rho'$:
\[
\int g(\rho,u,\theta) d \rho'(u,\theta) = c= \int g(\rho,u,\theta) d \rho(u,\theta) .
\]

In the mean field view, we may take the following connection of the discrete NN versus continuous NN when $m$ is large. The hidden units $[u_j,\theta_j]$ of the discrete NN \eqref{eq:twolayer-nn} can be viewed as $m$ particles sampled
from the distribution $\rho(u,\theta)$. In the training process, we move each particle $[u_j,\theta_j]$ using SGD, which is the derivative of the objective function with respect to each particle. In the continuous limit, we have  infinitely many particles, and each particle $[u,\theta]$ also moves according to the gradient of the objective function with respect to the parameter. In the literature, such a gradient is often referred to as {\em gradient flow} \cite{ambrosio2008gradient}, which characterizes the learning dynamics of the continuous formulation. In the following, we will present a more mathematical description. 

In the continuous formulation, a hidden unit can be regarded as a particle indexed by a parameter $z_0$ sampled from
a distribution $\nu_0(z_0)$. Here $z_0$ only plays the role of discrete index $j$ in the discrete formulation, and its own value is of no significance. The initial distribution $\nu_0(z_0)$ is introduced for convenience so that we can sample over the index $z_0$. In the discrete setting, it is simply the uniform distribution over $j=1$ to $j=m$. 

Each particle indexed by $z_0$ also has a parameter $[u,\theta]$ which will be trained. 
We assume that at time $t$, we move each particle during the training process, so that the model parameter becomes $[u(t,z_0),\theta(t,z_0)]$. If we take $z_0 \in \real^{k+d}$, with $\nu_0(z_0)$ as a Gaussian distribution, then we may simply initialize $[u,\theta]$ as $[u(0,z_0),\theta(0,z_0)]=z_0$. 
Since $z_0$ is sampled from $\nu_0(z_0)$, the particles  $[u(t,z_0),\theta(t,z_0)]$ induces
a probability measure $\rho_t(u,\theta)$ on $[u,\theta] \in \real^k \times \real^d$.
Here the time dependent parameters $u(t,z_0)$ and $\theta(t,z_0)$ are obtained via training over the time. 

Using the above terminology, the optimization of $\rho(u,\theta)$ in \eqref{eq:twolayer-nn-cont} leads to a distribution $\rho_t(u,\theta)$ at training time $t$, which is by moving hidden unit parameters $[u,\theta]$ via gradient flow with respect to the objective function
$\phi(\rho_t)$. More precisely, the corresponding particle movement in the continuous limit obey the gradient flow equation \cite{chizat2018global}:
\begin{equation}
\begin{cases}
  \frac{\partial u(t,z_0)}{\partial t}  &= -\eta(t) \nabla_u g(\rho_t,u(t,z_0),\theta(t,z_0)) ,\\
  \frac{\partial \theta(t,z_0)}{\partial t}  &= -\eta(t) \nabla_\theta g(\rho_t,u(t,z_0) ,\theta(t,z_0)) ,
  \end{cases}\label{eq:grad-flow}
\end{equation}
where the particle gradient $g(\rho,u,\theta)$ for a particle $[u,\theta]$ is defined in \eqref{eq:grad}. The gradient flow direction of a particle $[u,\theta]$ is the gradient of $g(\rho,u,\theta)$, which is equivalent to the gradient of the objective with respect to each particle parameter $[u_j,\theta_j]$ in the discrete NN formulation.
Therefore \eqref{eq:grad-flow} is the continuous version of gradient descent method with respect to the model parameter $[u,\theta]$ associated with the hidden units, and this continuous version of gradient descent method tries to minimize the objective function \eqref{eq:opt-cont}.

The gradient flow equation \eqref{eq:grad-flow}  implies a partial differential equation for
the probability measure $\rho_t$ as
\begin{equation}
  \frac{\partial \rho_t(u,\theta)}{\partial t}   = \eta(t) \nabla \cdot \left[  \rho_t(u,\theta)  \nabla g(\rho_t,u,\theta) \right] , \label{eq:rho-t}
\end{equation}
with its solution interpreted in the weak sense. This equation describes the dynamics of the objective function parameter $\rho$ of \eqref{eq:opt-cont} under the continuous gradient descent method of \eqref{eq:grad-flow}.
Here we use the simplified notation
\[
\nabla g(\rho,u,\theta)=[\nabla_u g(\rho,u,\theta), \nabla_\theta g(\rho,u,\theta)].
\]
The differential equation of $\rho_t$ in \eqref{eq:rho-t} characterizes the dynamics of $\rho_t$ in the neural network training process, and it can be shown that the objective value reduces according to the following ordinary differential equation:\fangcong{
\begin{align}
 \frac{d\phi(\rho_t)}{d t}  &=  \int  \frac{\delta \phi(\rho_t)}{\delta \rho_t} \cdot  \frac{\partial \rho_t}{\partial t} d \theta du \notag\\
 &=   \int g(\rho_t, u, \theta) ~ \eta(t)~ \nabla \cdot \left[  \rho_t(u,\theta)  \nabla g(\rho_t,u,\theta) \right] d \theta du \notag\\
  &=- \eta(t)
  \int \| \nabla g(\rho_t,u,\theta)\|_2^2 \; d \rho_t (u,\theta) .\label{eq:obj-dynamics}
\end{align}}
The derivation of the first equation has used the calculus of variations, which may be considered as the functional gradient of $\phi$ with respect to $\rho_t$ by treating $\rho_t$ as an infinite dimensional vector indexed by $(\theta,u)$. The functional gradient is given by $g(\rho_t,u,\theta)$, which leads to the second equation. In the third equation, we have used the integration by parts, and with a slight abuse of notation, we have used the notation $d \rho_t(u,\theta)=\rho_t(u,\theta) d\theta d u$, which does not differentiate measure $\rho_t$ and its corresponding density representation.
This equation is the key to prove global convergence in the MF approach. It shows that the gradient descent method of \eqref{eq:grad-flow} reduces the objective function of \eqref{eq:opt-cont}, and the result is stated using the probability measure $\rho_t$, which is what we want to learn in the continuous NN formulation. 

\begin{figure*}
	\centering
	\begin{tabular}{cc}
	 \includegraphics[width=0.40\textwidth]{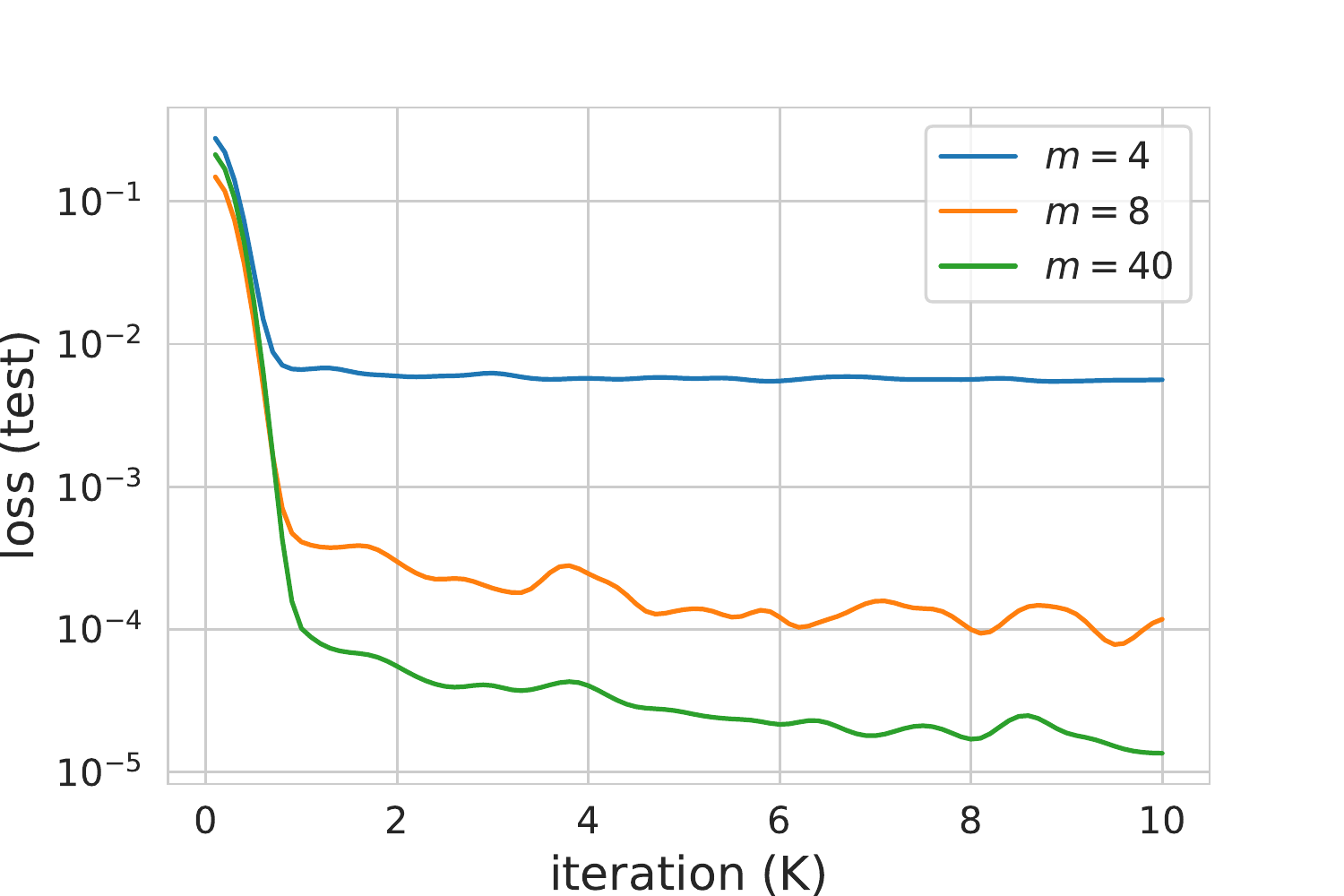} & \includegraphics[width=0.40\textwidth]{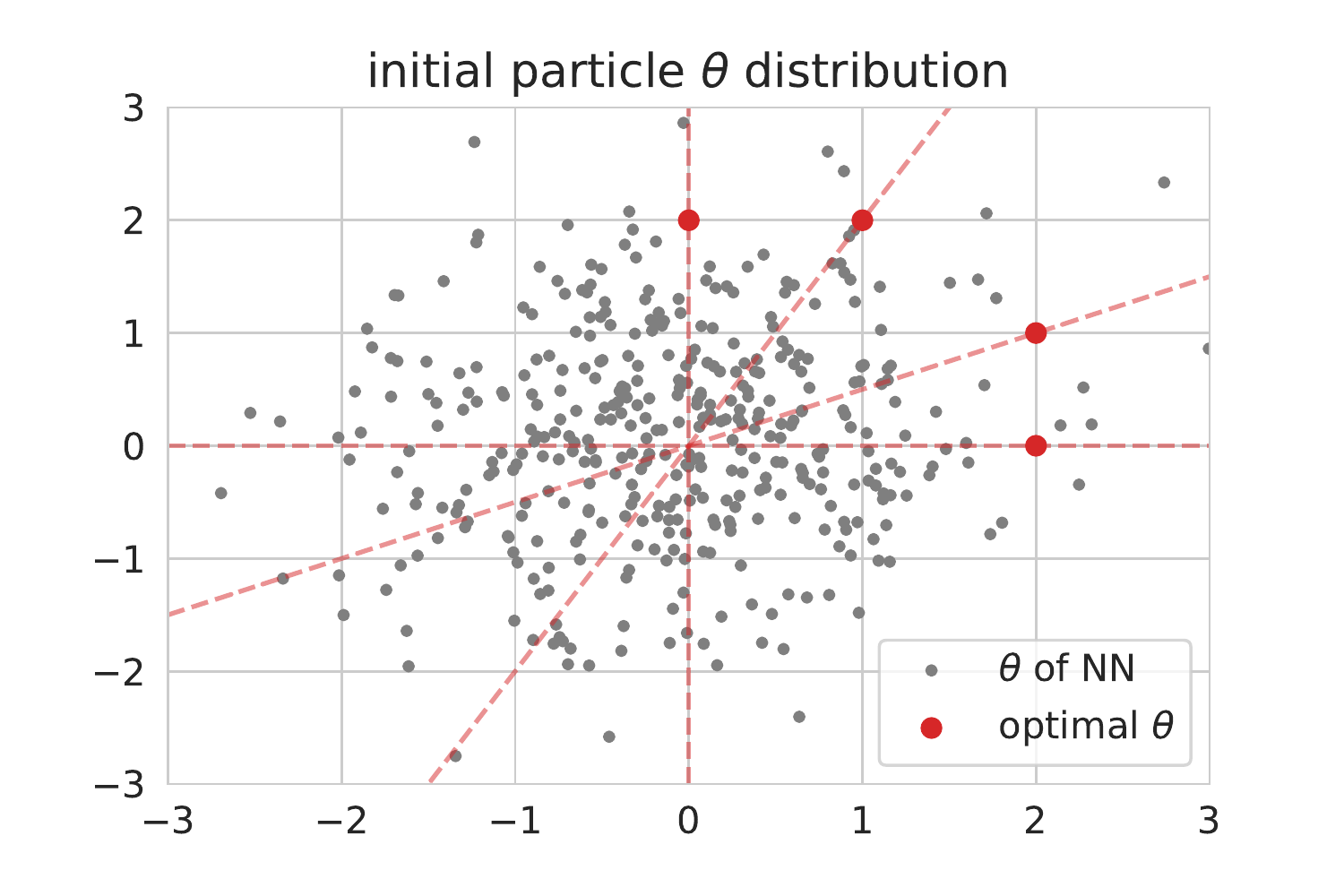}\\
	 \footnotesize{(a) change of loss}&\footnotesize{(b) initial $\tilde \theta$}
	\end{tabular}
    \begin{tabular}{ccc}
        \includegraphics[width=0.30\textwidth]{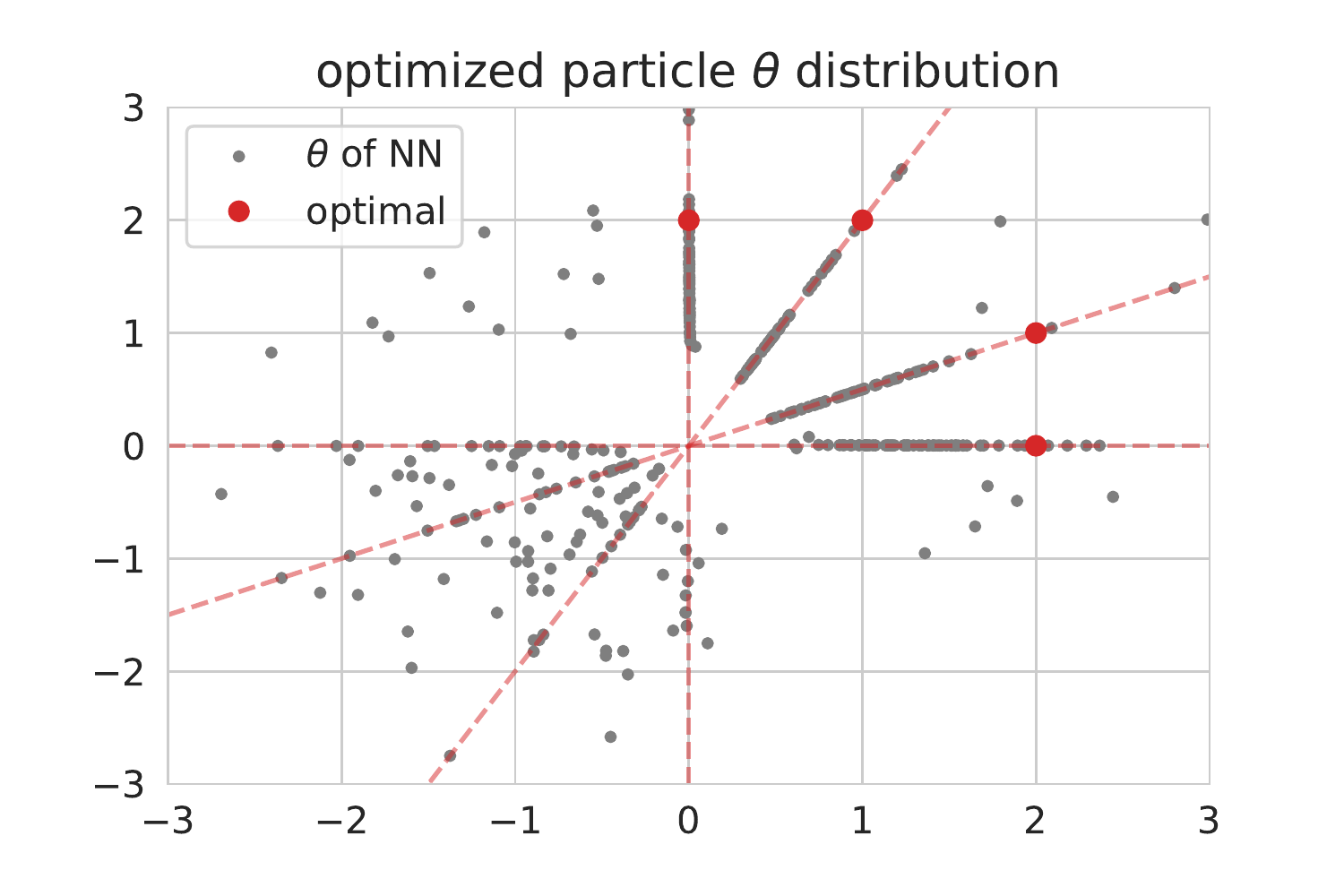}&\includegraphics[width=0.30\textwidth]{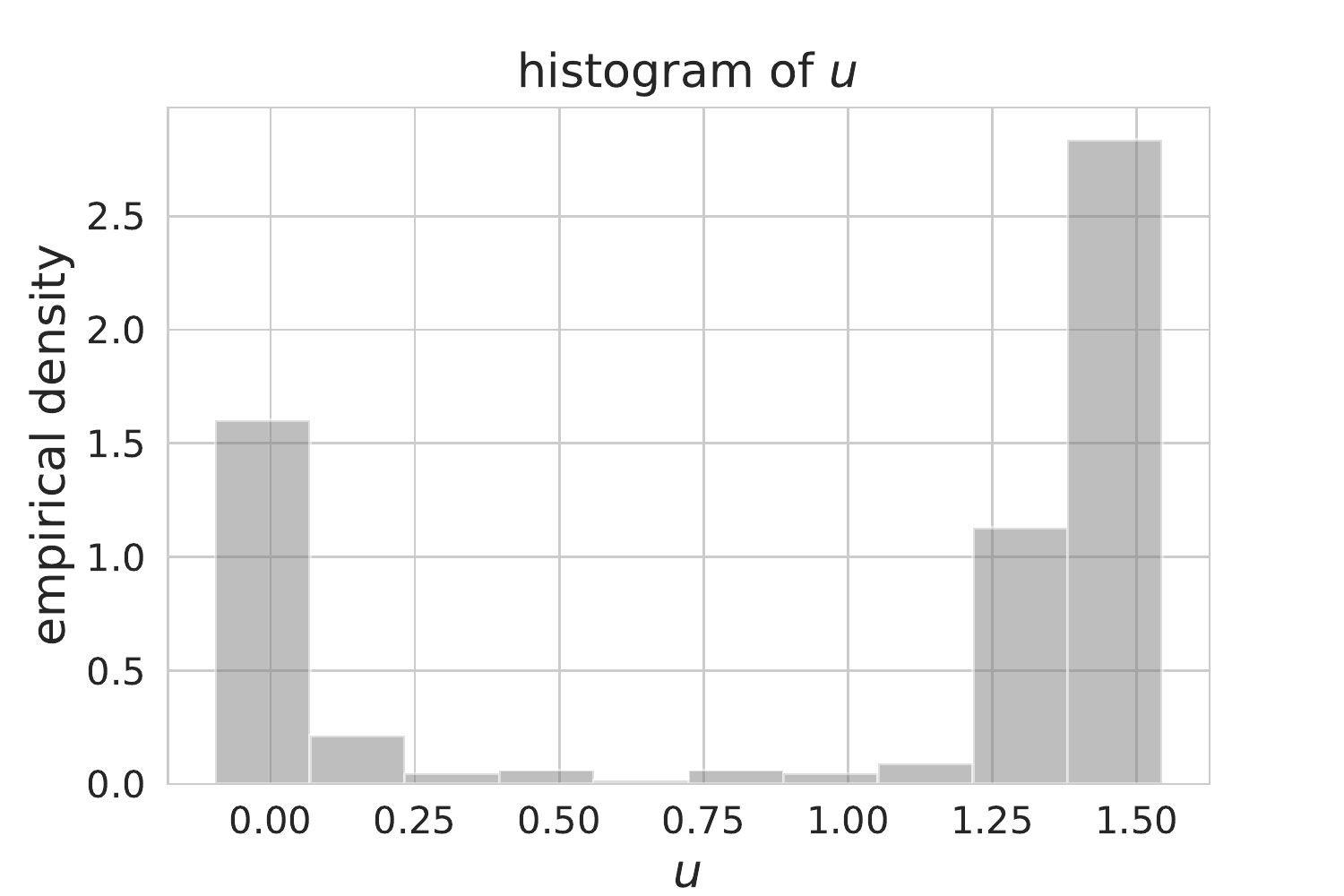} &\includegraphics[width=0.30\textwidth]{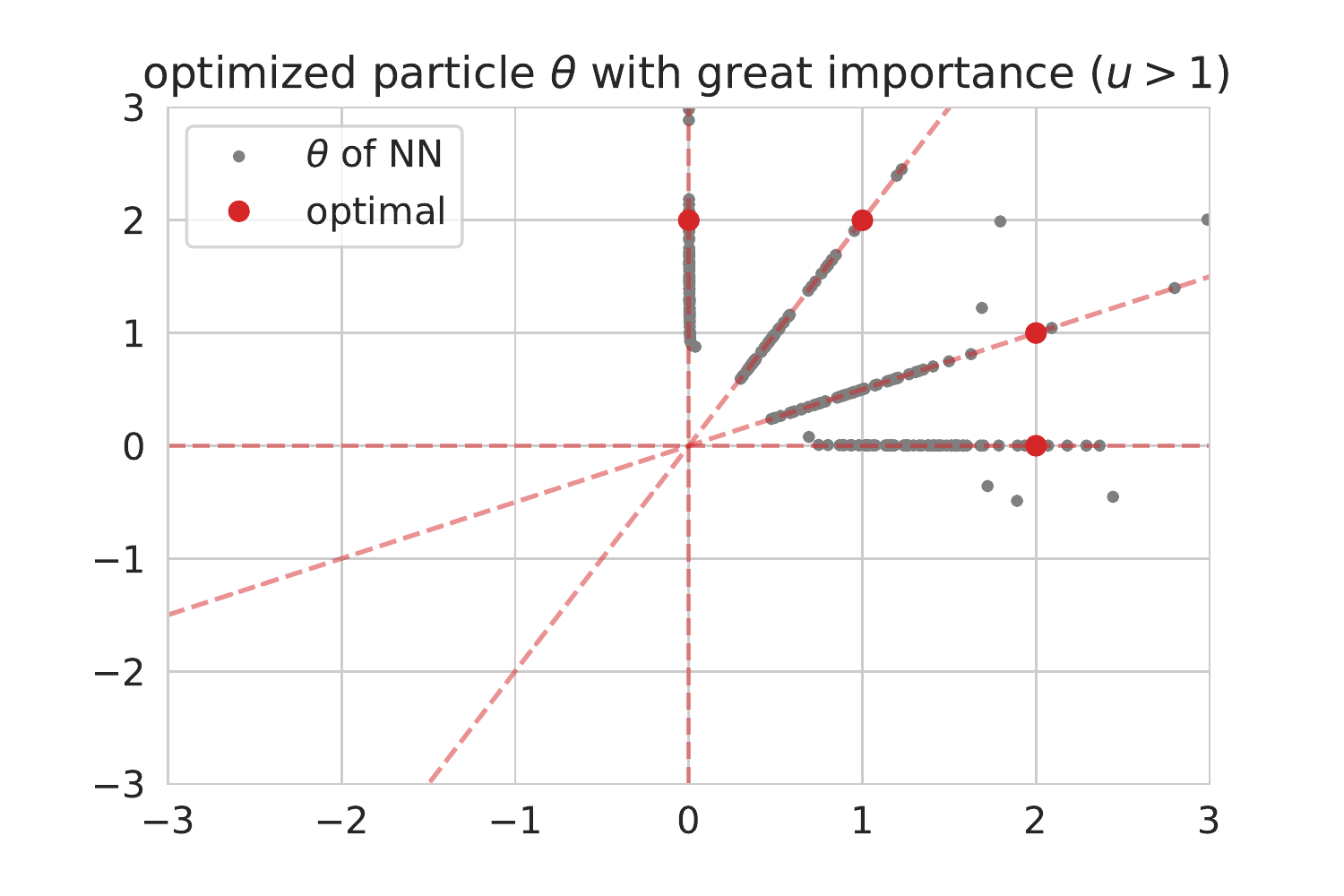}\\
        \footnotesize{(c) optimized $\theta$}&\footnotesize{(d) normalized histogram of $u$}&\footnotesize{(e) optimized $\theta$ with large $u$} \\
\includegraphics[width=0.30\textwidth]{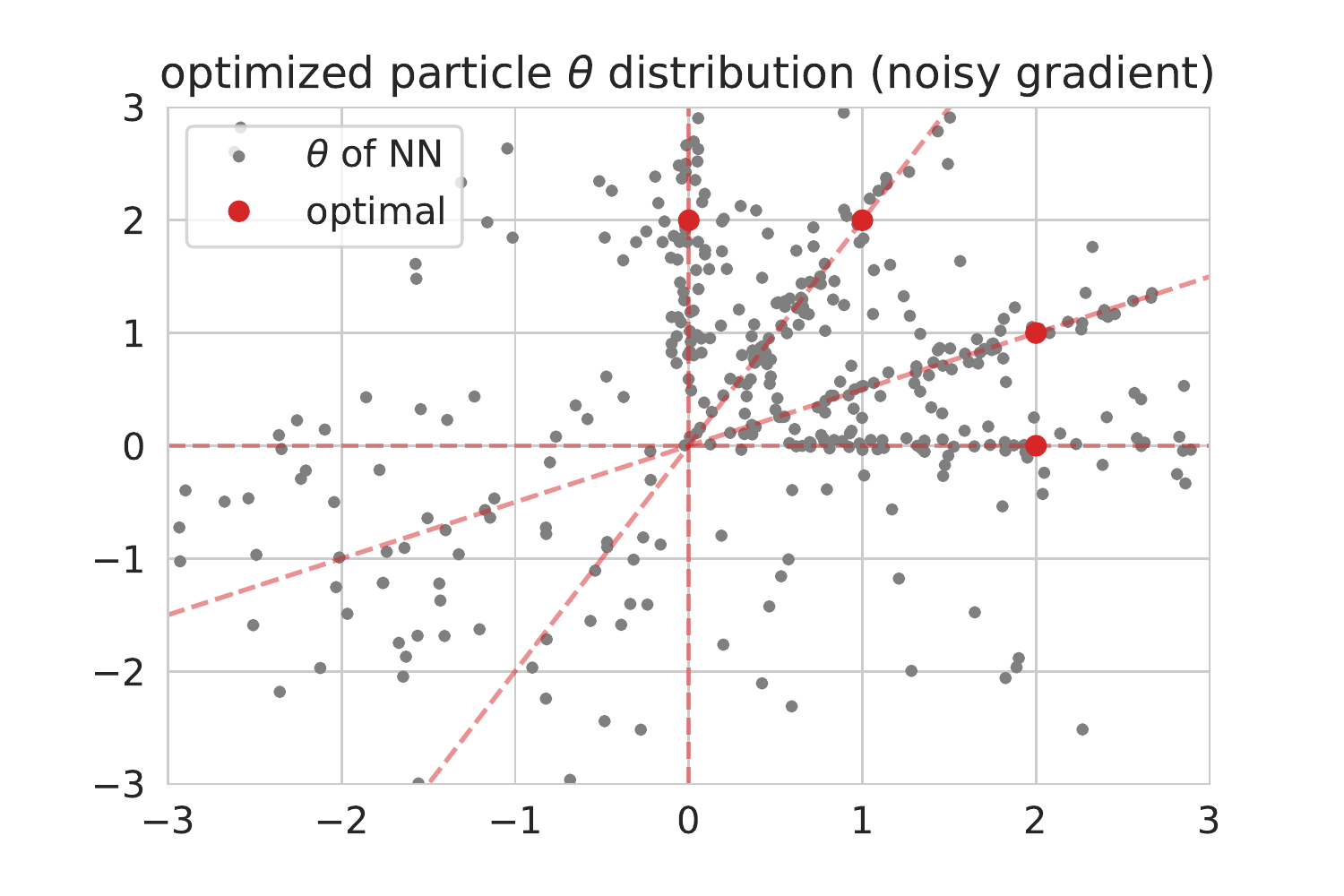}
        &\includegraphics[width=0.30\textwidth]{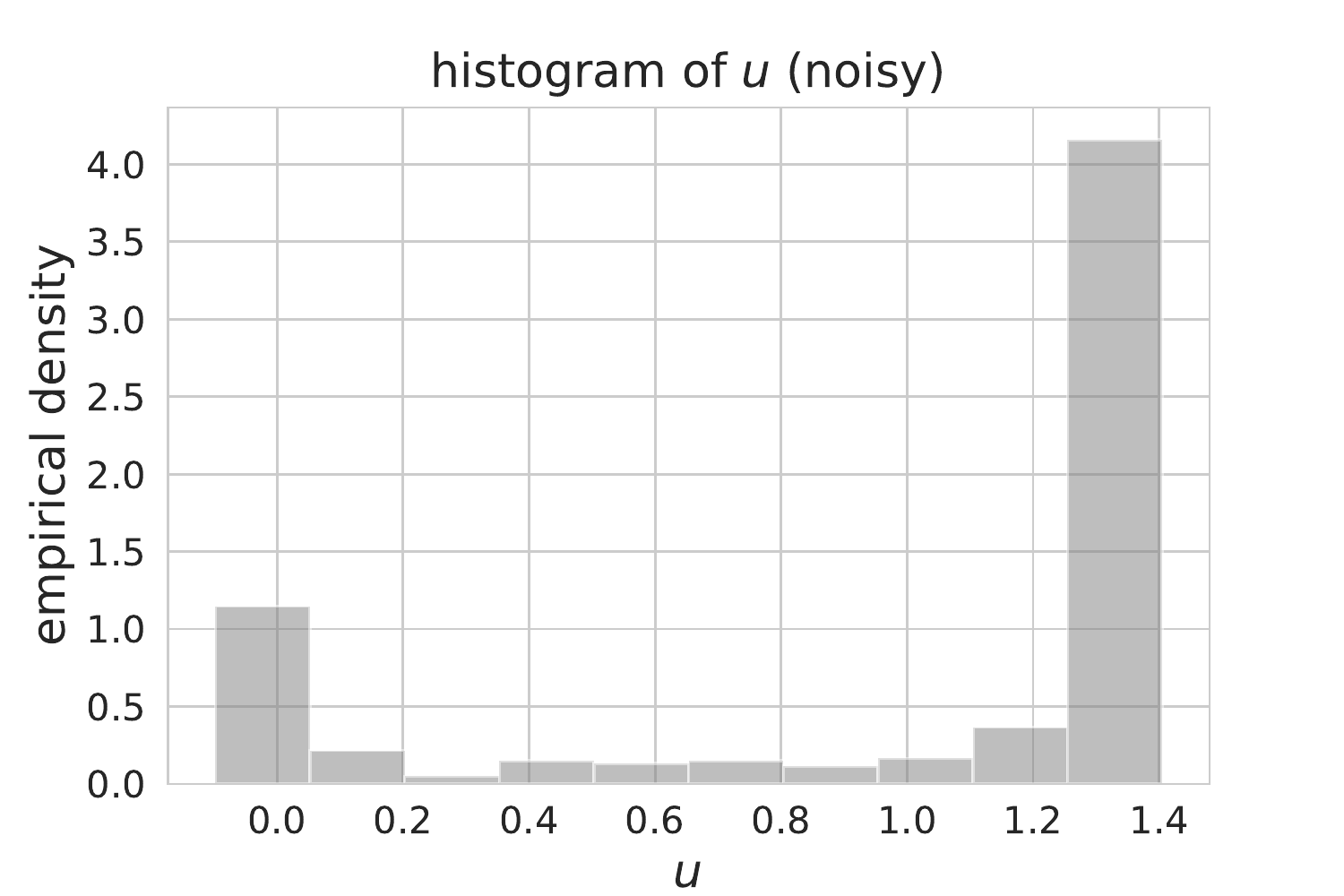}
        &\includegraphics[width=0.30\textwidth]{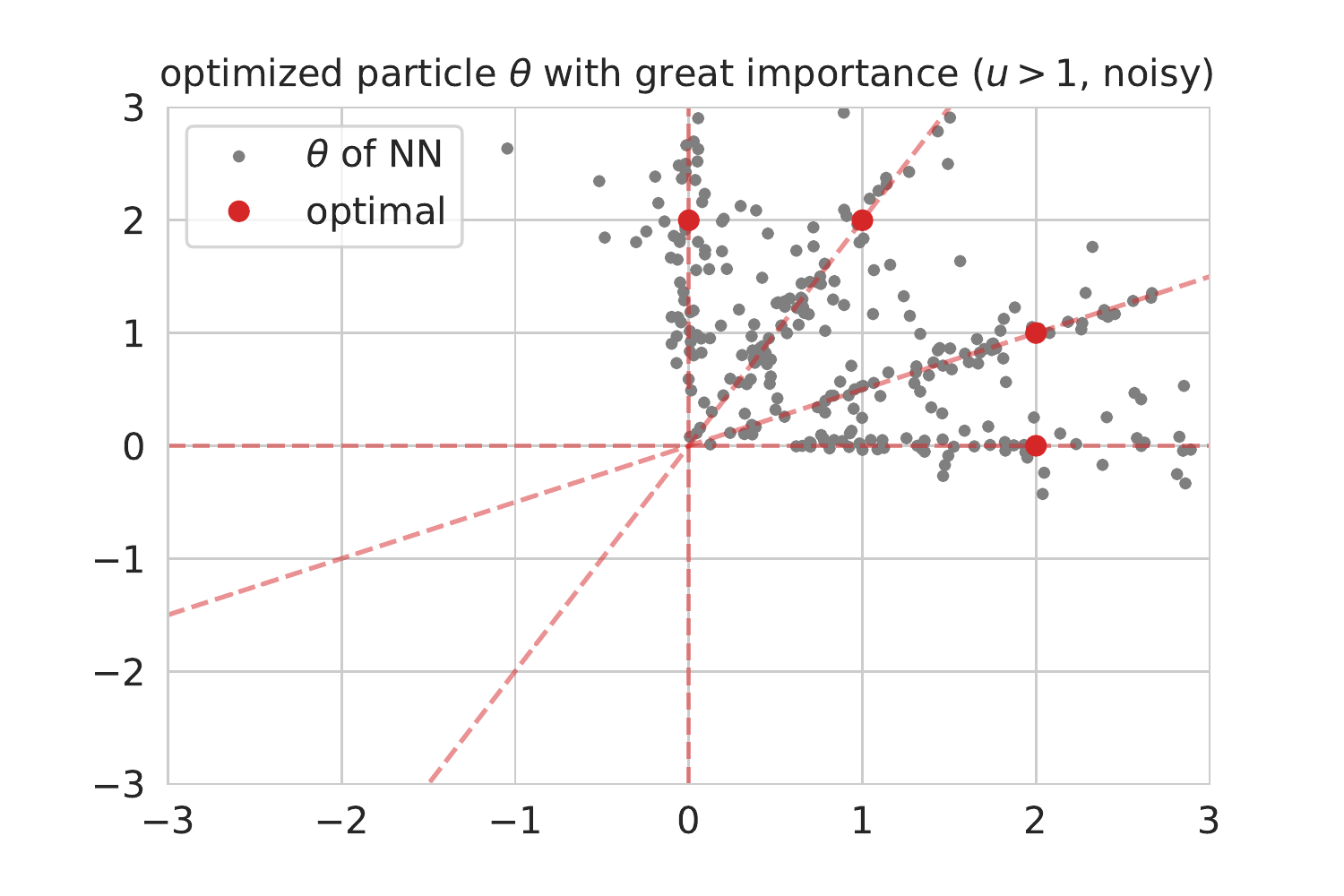}
        \\
       \footnotesize{(f) optimized $\theta$ with noisy gradient} &
    \footnotesize{(g) normalized histogram of $u$ (noisy)}
       &\footnotesize{(h) optimized $\theta$ with large $u$ (noisy)}  \\
    \end{tabular}
	\caption{Neural network optimization from a mean field perspective}
	\label{fig:nn_mf}
\end{figure*}

Since the objective function is bounded from below, from \eqref{eq:obj-dynamics} we can obtain that as $t \to \infty$, we must have $d \phi(\rho_t)/dt \to 0$. It follows that
\begin{equation}
  \lim_{t \to \infty} \int \|\nabla g(\rho_t,u,\theta)\|_2^2 d \rho_t(u,\theta) = 0  . \label{eq:vanishing-grad}
\end{equation}
However, this does not ensure that the objective function reaches the global minimum, unless additional conditions are imposed. Next, we shall present an intuitive explanation first, and then describe more rigorous results. 

From \eqref{eq:vanishing-grad}, if we can show $d \rho_t(u,\theta) \neq 0$ for all $[u,\theta]$, then we have 
$\|\nabla g(\rho_t,u,\theta)\|_2^2=0$ for all $[u,\theta]$. In this case, from $\nabla g(\rho_t,u,\theta) \to 0$, we obtain $g(\rho_t,u,\theta) \to c$ for a constant $c$, which implies the first order condition \eqref{eq:first-order}.
 This result implies that GD training converges to the global optimal solution of \eqref{eq:opt-cont} in the continuous setting. 

A more rigorous treatment of the above reasoning was presented in \cite{meie7665}, which  considered a formulation with an additional entropy regularization term in $R(\rho)$. This entropy regularizer ensures that $d \rho_t(u,\theta) \neq 0$ for all $[u,\theta]$. In fact, with this regularization, the measure $\rho(u,\theta)$ always has a density: $d \rho(u,\theta) = p(u,\theta) d u d \theta$, and we can write the regularizer as
\[
  R(\rho) = \lambda_p \int p(u,\theta) \log p(u,\theta) d u d \theta + \int r(u,\theta) p(u,\theta) d u d \theta ,
\]
which modifies the regularizer in \eqref{eq:opt-cont} by adding an extra entropy term. 
Using this regularizer, it can be shown that there is a unique global solution that satisfies \eqref{eq:first-order}. Moreover, under mild conditions, we have $\rho_t(u,\theta)$ converges (weakly) to a distribution $\rho_\infty(u,\theta)$ that can be lower bounded by a normal distribution \cite{meie7665}. Then by using the Poincar\'e inequality for Gaussian random variables (which state that if $X$ is a standard normal random variable, and $f(X)$ is a real-valued function, then $\mathrm{Var}(f(X)) \leq \mathbb{E} \|\nabla f(X)\|_2^2$),
we may obtain from \eqref{eq:vanishing-grad} that $g(\rho_\infty,u,\theta)=c$ almost everywhere for some constant $c$. This implies that the first order condition \eqref{eq:first-order} holds. It follows that as $t \to \infty$, the solution converges to the unique global optimal solution. 

In a practical implementation of the gradient descent rule in \eqref{eq:grad} with entropy regularization, we need to compute the gradient $\nabla \log p(u,\theta)$ in $\nabla g(\rho,u,\theta)$. It can be shown that an equivalent implementation  is to add a random noise, and the corresponding gradient flow equation of \eqref{eq:grad-flow}
becomes a stochastic partial differential equations (SDE) with $t \geq 0$:
\begin{align*}
d\, [u(t,z_0),\theta(t,z_0)]
=& - \eta(t) \nabla g(\rho_t,u(t,z_0),\theta(t,z_0)) d t\\
&+ \sqrt{2 \lambda_p \eta(t) } d \, B(t) ,
\end{align*}
where $\{d B(t)\}_{t \geq 0}$ is the standard Brownian motion in $\real^{k+d}$, and $g(\cdot)$ is defined in \eqref{eq:grad}.

In the GD (or SGD) implementation of the Brownian motion component of this SDE, we simply add a Gaussian noise of $N(0,2\lambda_p \eta(t))$ to each GD update step with learning rate $\eta(t)$.
This method is referred to as {\em noisy gradient descent} in the literature. With the help of entropy regularization, it can be shown that noisy gradient descent for continuous NN converges to the unique global optimal solution, and overparameterized discrete NN with a sufficiently large $m$ approximately reaches this solution. This result can be used to explain why training of overparameterized NN is easier in practice, and why the (idealized) two-layer neural network training process can reach a good solution with consistent performance. 

\fangcong{The benefits of entropy regularization are three-fold: (i) When we supplement the loss function with entropy regularization, the overall learning problem becomes strictly convex. Thus a unique global minimum can be guaranteed. (ii) The implementation of the entropy regularization is a very simple addition of noise.  In practice, one can often observe that adding noise helps us to find a better solution. (iii) After injecting noise, $d\rho_t\neq 0$ for all $[\mu, \theta]$. This fact, combined with \eqref{eq:vanishing-grad}, implies the point-wise vanishing of $\nabla g(\rho_t,u,\theta)$, which implies that the global minimum is achievable.} 

On the other hand,  without the extra entropy regularization, the objective function of \eqref{eq:opt-cont} may not have a unique global optimal solution. That is, there can be more than one solutions that satisfy the first order condition \eqref{eq:first-order0}. However, we can still achieve the global convergence to the optimal objective function value of \eqref{eq:opt-cont}, although the final solution the training process converge to may not be unique. To analyze this situation (without the extra entropy regularization),  the authors in \cite{chizat2018global} considered a different assumption with homogeneous activation functions (such as ReLU) and homogeneous regularizers. Under such assumptions, it is possible to show that  the solution $\rho_t(u,\theta)$ converges to a global optimal solution (which may not be unique) that satisfies \eqref{eq:first-order0} as $t \to \infty$. 

In the mean field approach, learning the distribution $\rho$ can be viewed as learning effective feature representations. The ability of NN to learn feature representations is consistent with empirical observations. This perspective also explains why fully trained NN is better than RF and NTK, both of which employ random feature representations that are not learned. We will further discuss this aspect in Section~\ref{sec:exp}.

To visualize the process of learning $\rho$, we conduct an experiment to reproduce an $m=4$ sigmoid activated NN ($u_i=1$, $i=1,2,3,4$, denoted as $F_4$) by NNs with different width $m$. Note that the process of learning the target function can be recognized as the process of learning the target optimal $\rho_* =\sum_{i=1}^4 \delta_{\theta_i}/4$, where $\delta_x$ is dirac delta function at point $x$ and $\theta_i$ is the weight of the NN to be reproduced. Although we know that the target function can be represented by 4 neurons, Figure~\ref{fig:nn_mf} shows that using a larger $m$ leads to better learning. This is consistent with the theory of overparameterization.  In Figure~\ref{fig:nn_mf} (a), we use mean squared error loss to measure the difference between target and optimized NN. The training values are $F_4(x), x\sim N(0,100^2I)$ blurred with noise $\epsilon\sim N(0,0.1^2)$ (the reason to use large scale input is to improve the reconstruction difficulty).
It can be seen that with $m=4$, we will get stuck at a local minimum and cannot learn the correct target function. When we increase $m$, we achieve more and more accurate learning of the target function.
Figure~\ref{fig:nn_mf} (b) and (c) show the distributions of $\theta$ at initialization, and at the optimal solution when training convergence. We can see that they differ significantly, and thus in this case, NN training goes out of the NTK regime. In the end, the distributions of the neurons are scattered, with a large number of neurons become aligned with the target $\{\theta_i\}$ represented by the four red dots. Since the target function does not have a unique representation, therefore we cannot recover the parameters $\{\theta_i\}$, but only recover the function value represented by each $\{\theta_i\}$ using multiple $\theta$ parameters distributed over the lines of the targets. Therefore we can learn the target function reliably when $m$ is large, although we do not necessarily learn the four target parameters $\{\theta_i\}$. This is consistent with the analysis in \cite{chizat2018global}, where the NN training reaches the minimal training error, but not necessarily unique. 
 Figure~\ref{fig:nn_mf} (d) shows that many particles have a very small $u$, which means they are "wasted neurons" that do not affect the function value. If we remove these wasted neurons with small $u$, then we can display the effective neurons in (e), which are well-aligned with the target neuron directions, and they can approximately recover the functions represented by the four target neurons. 
The phenomenon of wasted neurons in (c) is because the target function is not strongly convex in $\rho$. Therefore there can be many solutions that achieve global optimal. 
The analysis in \cite{chizat2018global} demonstrates that under suitable conditions, the training process will converge to the global optimal, although the solution may not be unique. However, if we add the entropy regularization as in \cite{mei2019mean}, then the global solution becomes unique. Since the entropy regularization can be implemented using noisy gradient, we show in (f) the effect of using noisy gradient on this problem. It shows that with this regularization, there is a 
significant reduction of wasted neurons, and the final solution are nearly aligned with the directions of the four target neurons. 
Because the function is not uniquely represented by the four neurons,
we still do not recover the four target neurons. Instead, the final
solution converges to a unique optimal distribution $\rho_*$ which is
a smooth distribution around the directions of $\{\theta_i\}$ in the
continuous limit. In real finite width NNs, some neurons may still get stuck in the low density
regions. We can thus observe a small portion of wasted neurons, which will decrease with wider NNs or larger
noises.

\begin{figure}
    \centering
    \includegraphics[width=0.5\textwidth]{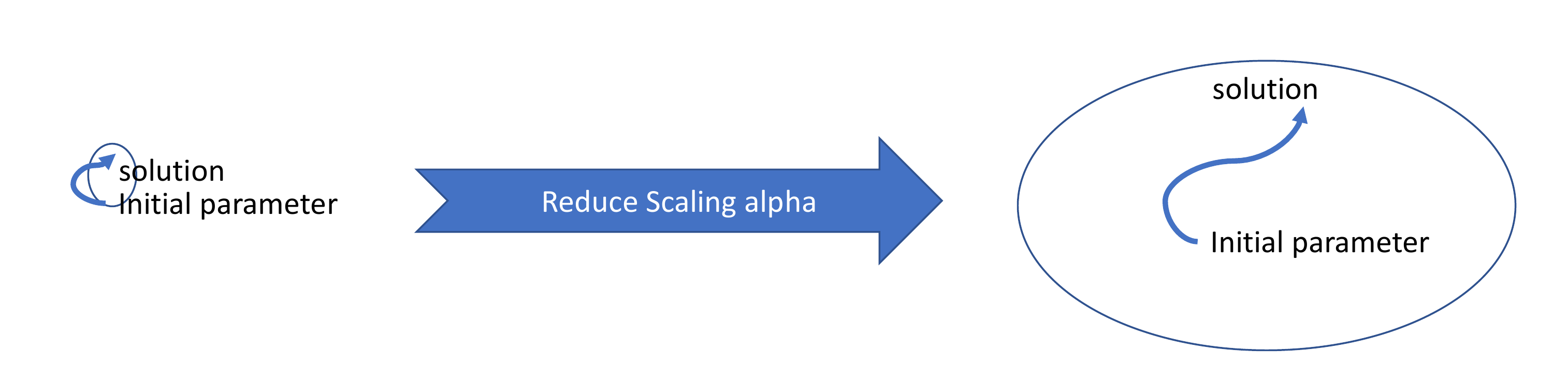}
    \caption{Scaling factor $\alpha$ controls NN behavior: NTK versus MF}
    \label{fig:ntk-mf}
\end{figure}

We may summarize some key points of the MF approach as follows.
\begin{itemize}
\item The method learns a distribution $\rho$ which behaves like learning effective feature representations.
\item GD or noisy GD (SGD) over model parameters define gradient flows with dynamics characterized by partial differential equations (PDE)
\item Under appropriate assumptions, the solution of the underlying PDE converges to the optimal solutions in $\rho$ that satisfies the first order condition \eqref{eq:first-order}.
\item The optimal solution can be far from the initial parameter, leading to more realistic model for neural network learning than NTK. 
\end{itemize}
It can be shown that as $\alpha \to \infty$, the dynamics of MF becomes similar to that of NTK under suitable conditions, and the solution becomes closer and closer to the initialization \cite{mei2019mean,chen2020mean}.
When we reduce $\alpha$, the final solution becomes farther apart from the initialization.
This migrates from the NTK regime to the MF regime. 
This phenomenon is illustrated in Figure~\ref{fig:ntk-mf}. 
We also summarize the relationship between NTK and MF in Table~\ref{tab:ntk-mf}.

\begin{table}
	
	\caption{Comparison of the NTK view and the MF view}
	\centering
\begin{tabular}{|c|c|}
 \hline
NTK regime & MF regime \\ \hline
Large initial parameter & 
Small initial parameter
\\
Solution close to initialization
& Solution far from initialization
\\
Learning model parameters&
Learning feature distributions
\\ 
Without weight-decay regularization &
Can incorporate regularization\\
\hline
\end{tabular}

\label{tab:ntk-mf}
\end{table}

While MF for two-layer NN is well-understood,  compared to NTK, it is significantly more difficult to generalize
 MF to handle deep neural network structures. It is also more difficult to obtain concrete complexity results using MF, which requires study both the discretized differential equations, and the convergence rate in terms of letting $m \to \infty$.

\section{The Importance of Feature Learning}
\label{sec:exp}

\begin{table*}[tb]
\centering
\caption{ Pros and Cons of RF, NTK, and MF. Note that though RF and NTK can be applied on DNN, they are still a linear (one-layer) model and do not fully explore the hierarchical architecture. See discussion on further directions in Section \ref{sec:conclusion}. }
\fangcong{
\begin{tabular} {c |c |c|c }\hline
Properties &RF & NTK&MF \\\hline\hline
 Special initialization &  ~~~yes~~~ &  ~~~yes~~~  &  ~~~no~~~ \\\hline 
 Incorporate regularization& ~~~yes~~~  & ~~~no~~~  & ~~~yes~~~ \\\hline
Random/discriminative feature& ~~~random~~~ &  ~~~random~~~  & ~~~discriminative~~~ \\\hline
Quantitative computational  complexity result&  ~~~yes~~~ &  ~~~yes~~~  &  ~~~open ~~~ \\\hline
Applicable on deep neural network& ~~~yes~~~ &  ~~~yes~~~ & ~~~open ~~~ \\\hline
\end{tabular}}
\label{comparison}
\end{table*}

In the previous sections, we presented three mathematical models closely related to two-layer neural networks: RF, NTK, and MF.  \fangcong{A summary of the pros and cons for the three models are shown in Table \ref{comparison}}. The first two approaches, RF and NTK, employ simplified mathematical models by treating two-layer NNs as linear models with random features. The MF view, on the other hand,  directly model the feature learning dynamics of NN. It was argued by \cite{fang2019over} that a theoretical understanding of feature learning is the key to explain the success of NN. Following the argument of \cite{fang2019over}, this section compares the three models empirically from the feature learning perspective. 

It was pointed out in \cite{fang2019over} that when $m$ is large, the hidden units of a discrete NN in \eqref{eq:twolayer-nn} can be regarded as $m$ (nearly) independent samples from a distribution $\rho$, which is the distribution of the corresponding continuous NN in \eqref{eq:twolayer-nn-cont}. 
If we treat the function value $f(\rho,x)$ of the continuous NN as the target, then it follows that the error of discretize NN is caused by the variance of sampling $m$ hidden units from $\rho$, which converges to
\begin{align}
&\|f([u,\theta],x)-f(\rho,x)\|_2^2 \nonumber \\
\approx& \frac{1}{m}
\int \|\alpha u h(\theta,x)-f(\rho,x)\|_2^2 d \rho(u,\theta) , \label{eq:var}
\end{align}
where $f([u,\theta],x)$ represents the discrete NN of \eqref{eq:twolayer-nn}, with each hidden unit $j$ sampled (independently) from $\rho$, and $f(\rho,x)$ is the corresponding continuous NN of \eqref{eq:twolayer-nn-cont}.

Since under suitable conditions, the continuous representation $f(\rho,x)$ can reach a global optimal solution via training, it can be regarded as the target function we try to learn with discrete NN. A good feature representation of the target is thus a feature distribution $\rho$ so that its continuous NN can be well approximated by the corresponding discrete NN via \eqref{eq:var}. This means that the variance on the right hand side of \eqref{eq:var} should be small. If we consider using the $L_2$ regularization for $u$, and assume that $h(\theta,x)$ is batch-normalized as 
\[
\frac1n\sum_{i=1}^n h(\theta,x_i)^2 =1
\]
for all $\theta$, 
then it is shown in \cite{fang2019over} that when fully optimized, $\|u\|_2$ is nearly a constant with respect to the distribution $\rho(u,\theta)$. It implies that the variance of \eqref{eq:var} achieved by NN training is nearly minimized among all $\rho'$ such that $f(\rho',x)=f(\rho,x)$. Therefore for a fixed $m$, we will achieve the smallest error with discrete NN and the learned probability measure $\rho(u,\theta)$. 
We thus conclude from this result that after NN training, the discrete NN can efficiently represented the target function by learning an effective feature representation characterized by the feature distribution $\rho(u,\theta)$. 

If we compare this learned feature representation to the random feature approaches (RF or NTK), the feature representation learned by NN leads to more efficient discrete representation by sampling from the distribution. This efficiency explains the superiority of NN over the random feature approach. 
A consequence of the optimal feature representation point of view in \cite{fang2019over} is the possibility to use a generative model to learn such a distribution $\rho$, and then use this generative model to replace the initial random features (i.e. random Gaussian distributions) in RF and NTK to generate hidden units of the neural network. If we consider the random features sampled from this learned distribution, instead of random features at the initialization, more effective RF and NTK can be obtained. 
This was illustrated in \cite{fang2019over,atanov2018deep}, which we present here as well.

\begin{figure*}
	\begin{center}
		\begin{tabular}{ccc} 
			\includegraphics[width=0.3\textwidth]{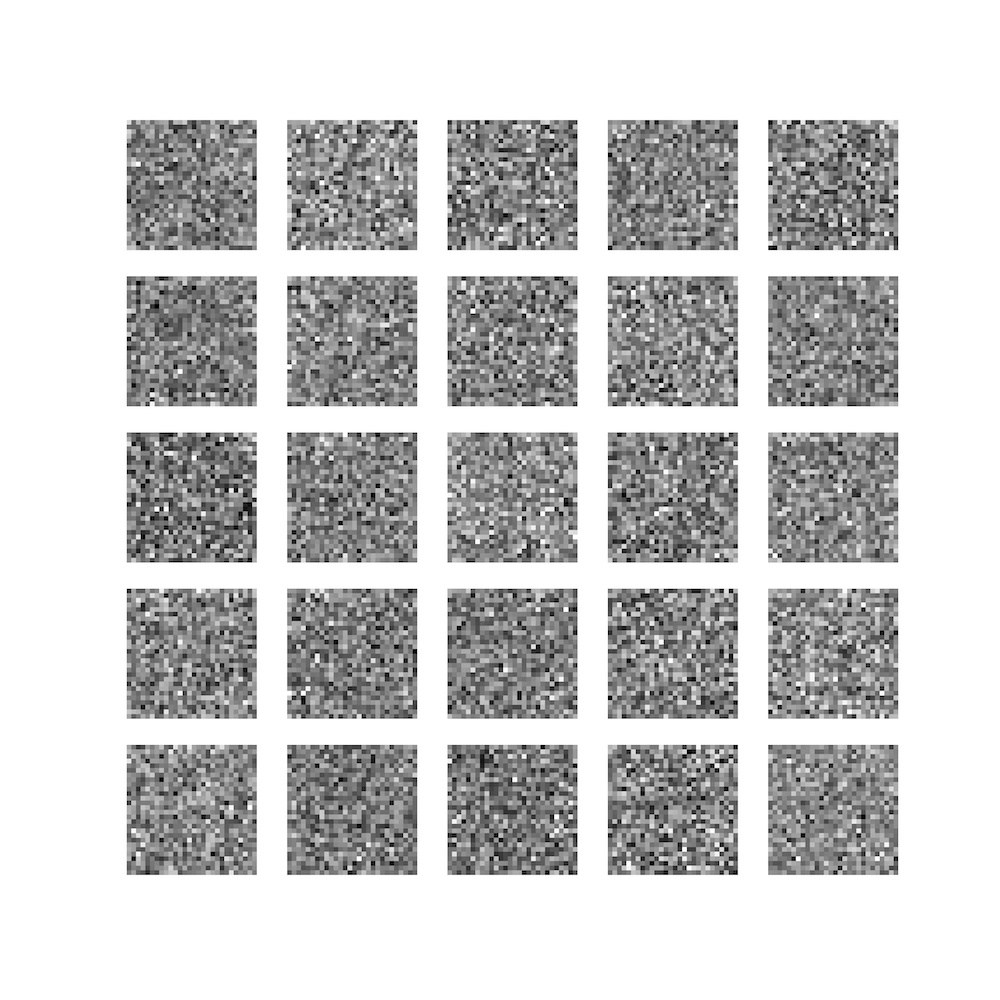}&
			\includegraphics[width=0.3\textwidth]{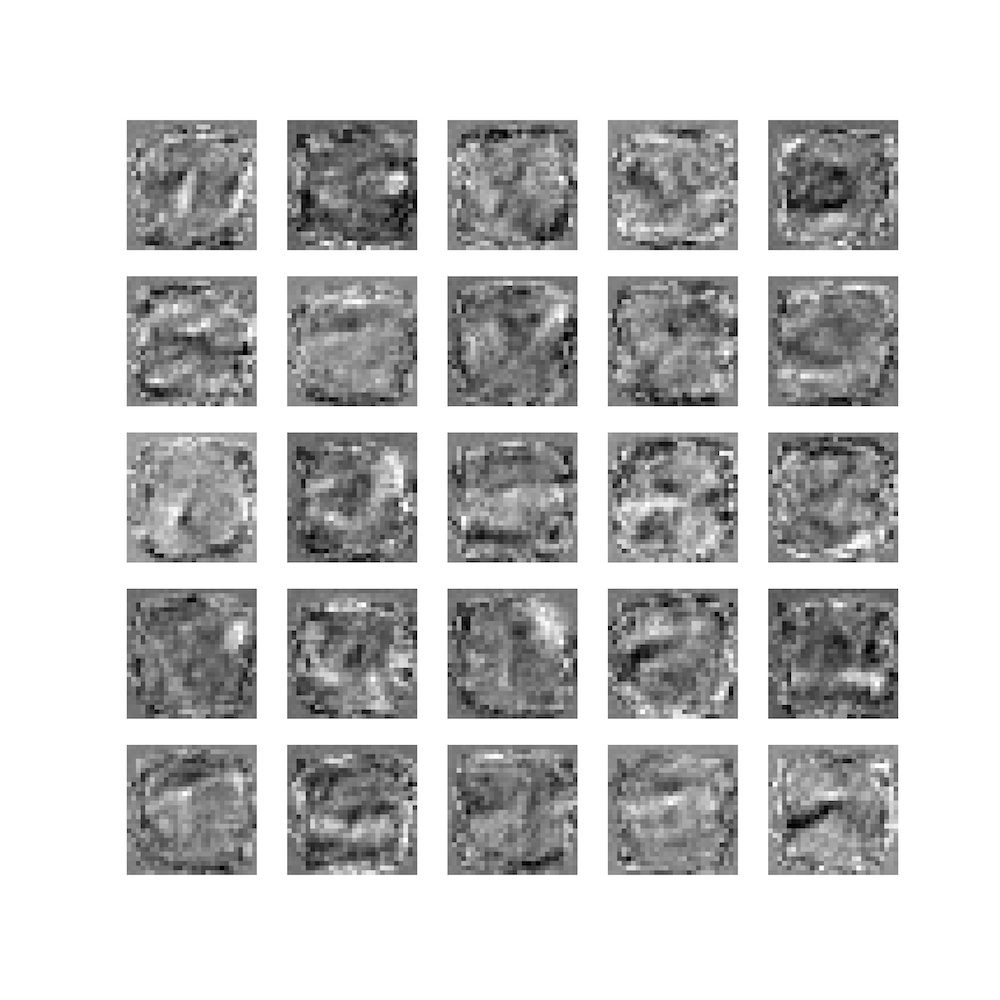}&
			\includegraphics[width=0.3\textwidth]{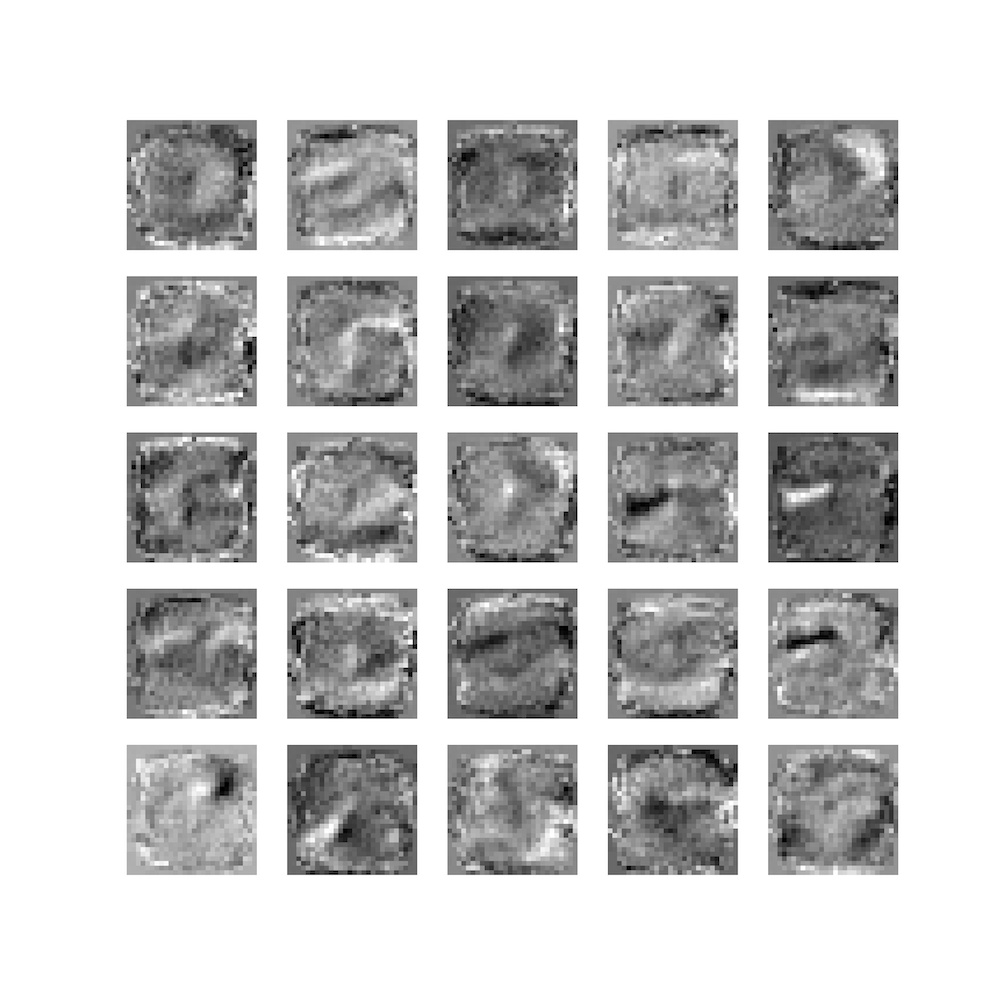}\\
			(a) Random & (b) Optimized & (c) Generated
		\end{tabular}
	\end{center}
	\caption{Visualization of weights on MNIST dataset (reshaped to $28\times 28$)}
	\label{fig:visualization}
\end{figure*}

For Convolutional Neural Networks (CNN), the phenomenon of learning features is a consensus among practitioners \cite{erhan2009visualizing,atanov2018deep}. 
\hanze{A visualization of this phenomenon is shown in Figure \ref{fig:visualization}. When we use a Variational Auto-Encoder (VAE) \cite{kingma2013auto} to learn the optimized $\rho$ distribution from samples of pre-trained models, we observe meaningful patterns not found at the initialization. In particular, to obtain samples from $\rho_*$, we prepare $1,000$ pre-trained two-layer NNs ($m=100$) with different initializations. Note that the weight of the pre-trained NNs can be regarded as samples $\{\theta_i\}$ from $\rho_*$. We can then use the generative model (VAE) to learn the transform from the standard normal distribution to the target distribution $\rho_*$ with these samples $\{\theta_i\}$.
 }

\begin{figure*}
	\centering
	\includegraphics[width=0.30\textwidth]{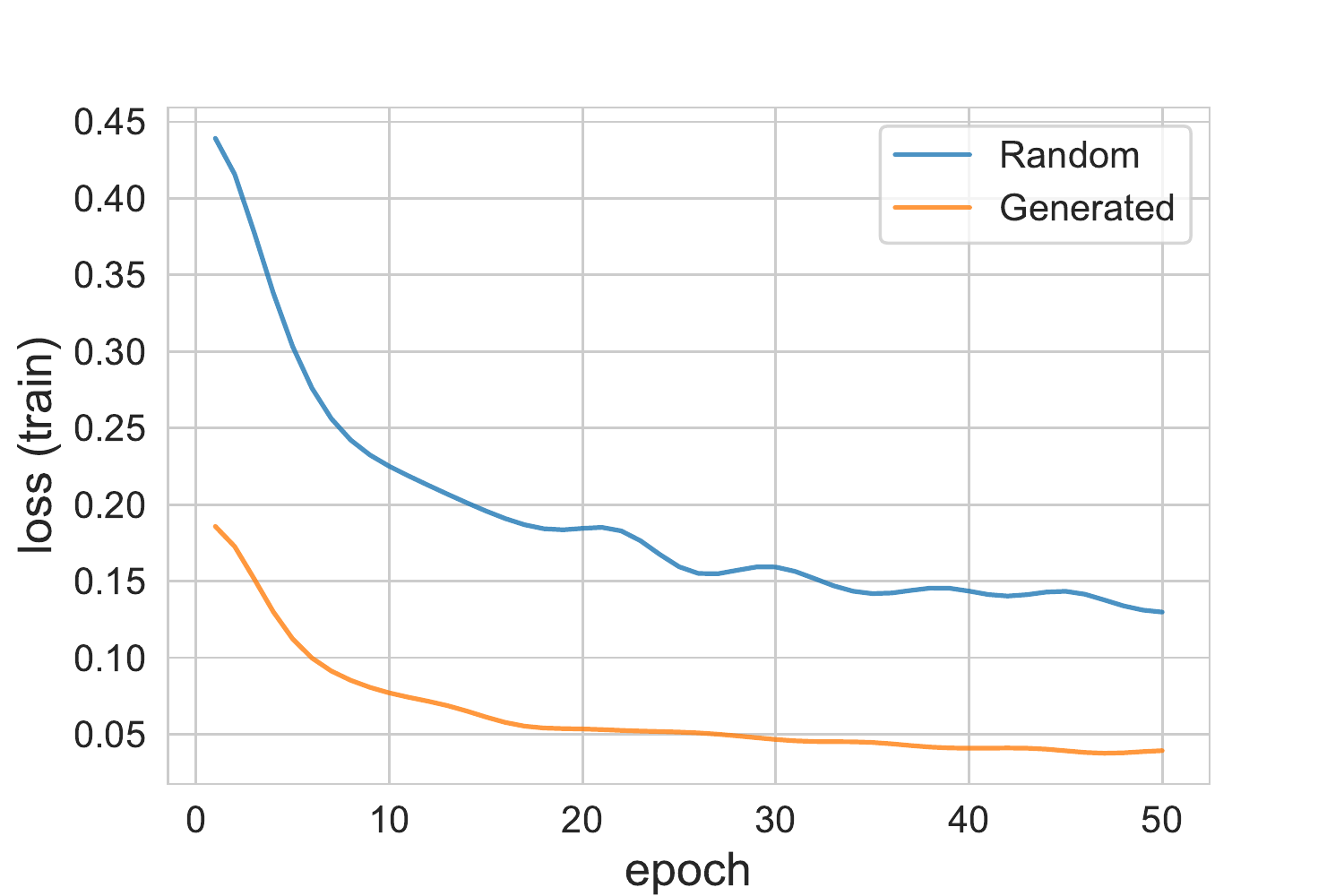}
	\includegraphics[width=0.30\textwidth]{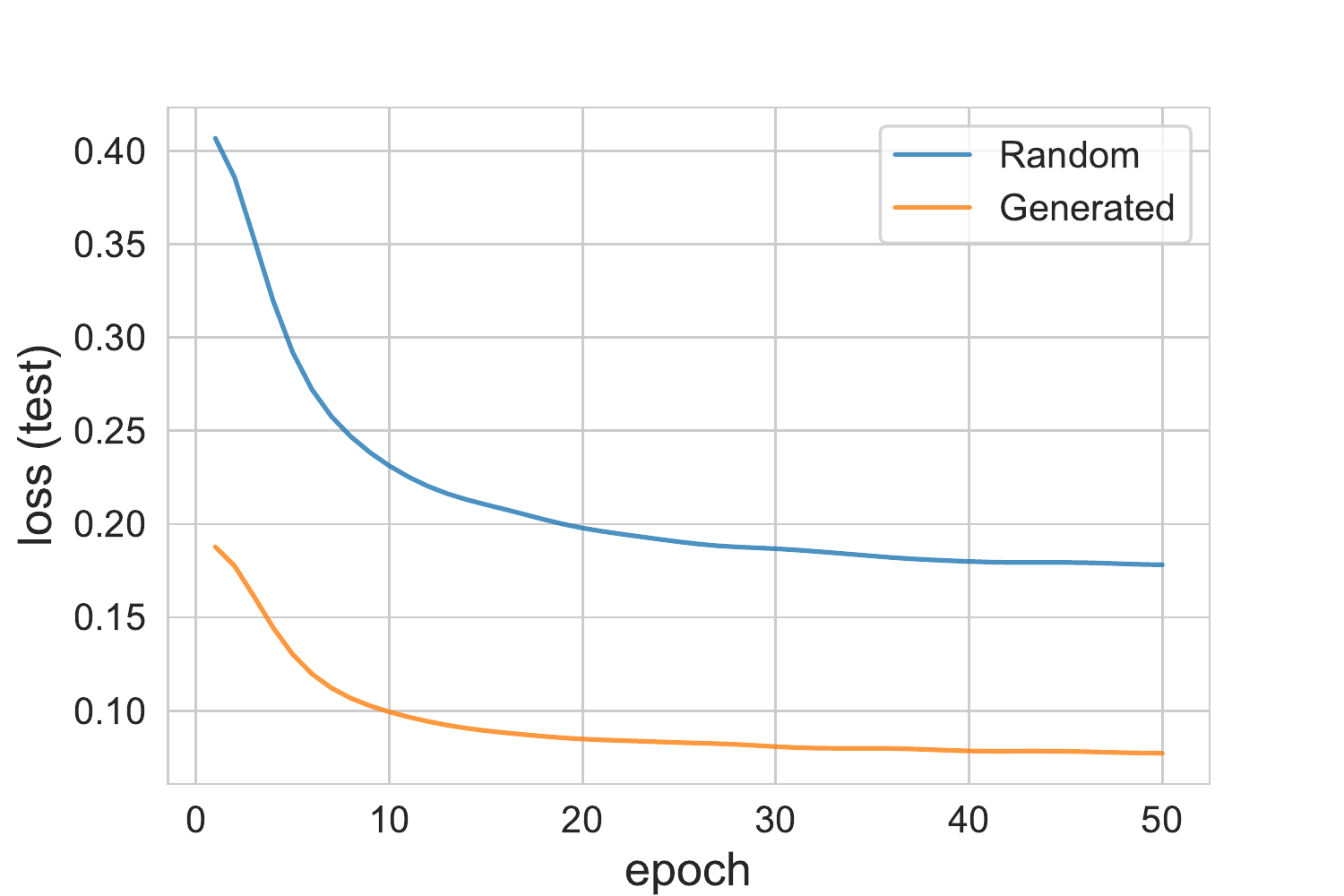}
	\includegraphics[width=0.30\textwidth]{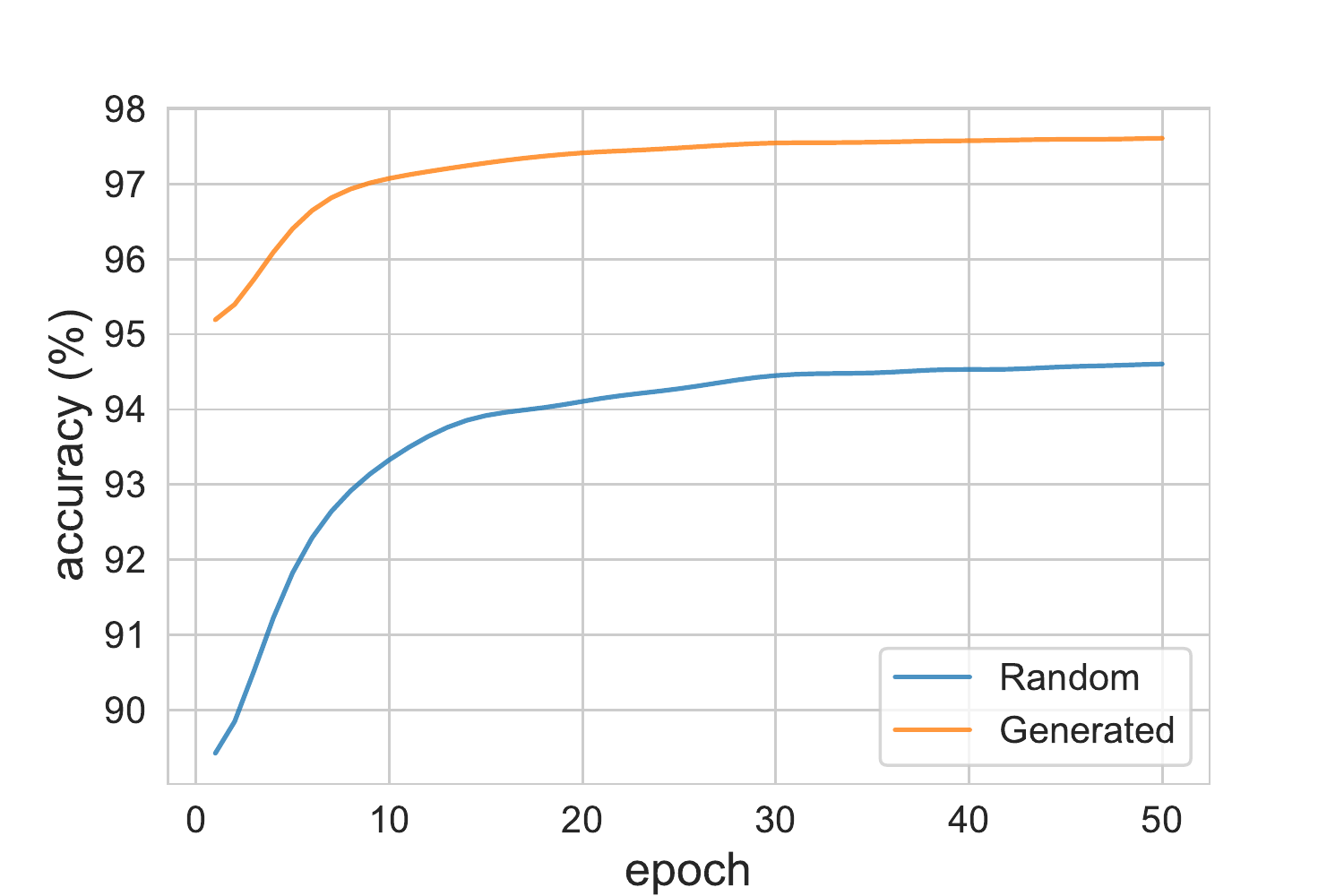}
	\caption{Random feature versus Repopulated feature ($m=10^3$)}
	\label{fig:mnist_repop}
\end{figure*}

\begin{figure*}
	\centering
	\includegraphics[width=0.30\textwidth]{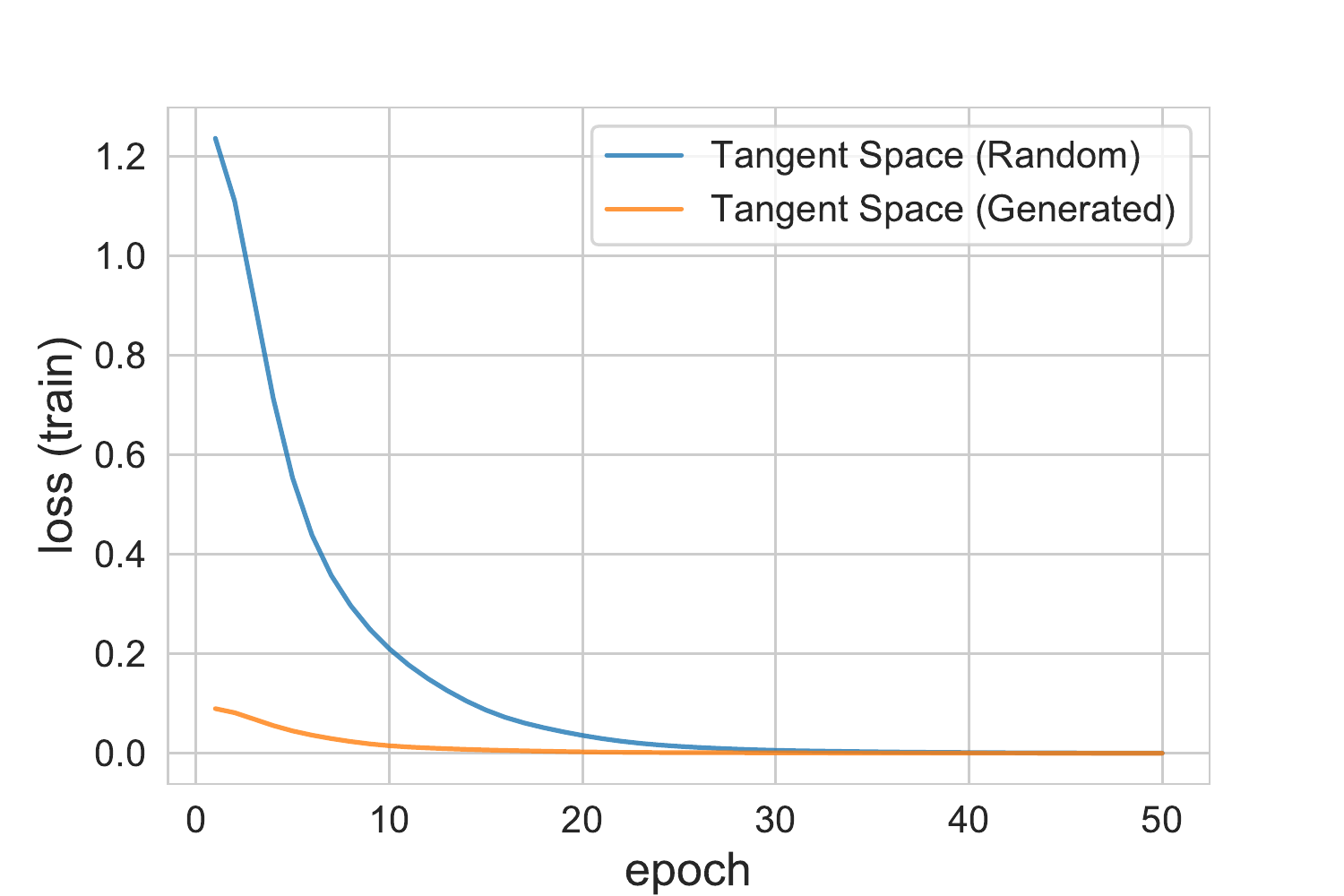}
	\includegraphics[width=0.30\textwidth]{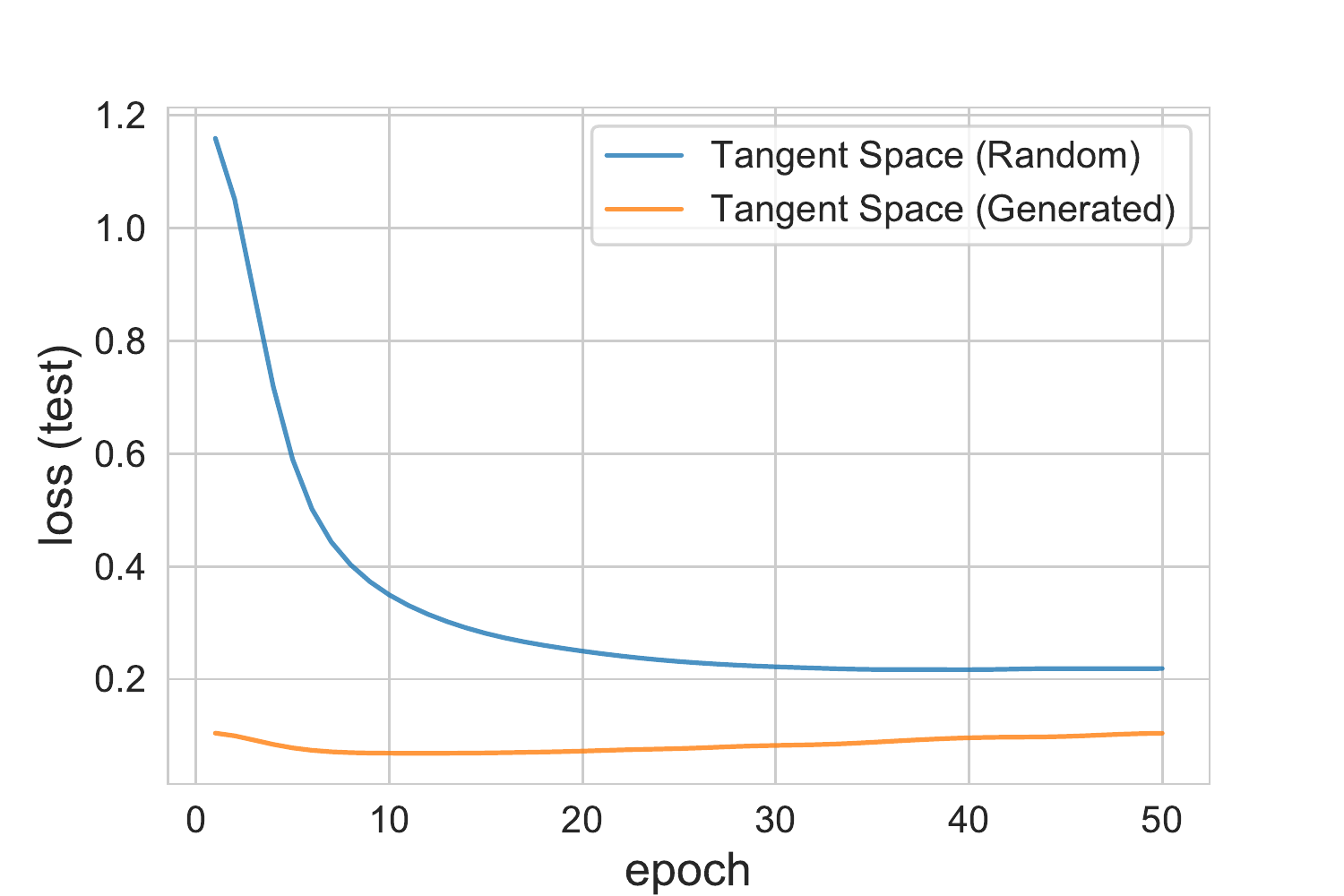}
	\includegraphics[width=0.30\textwidth]{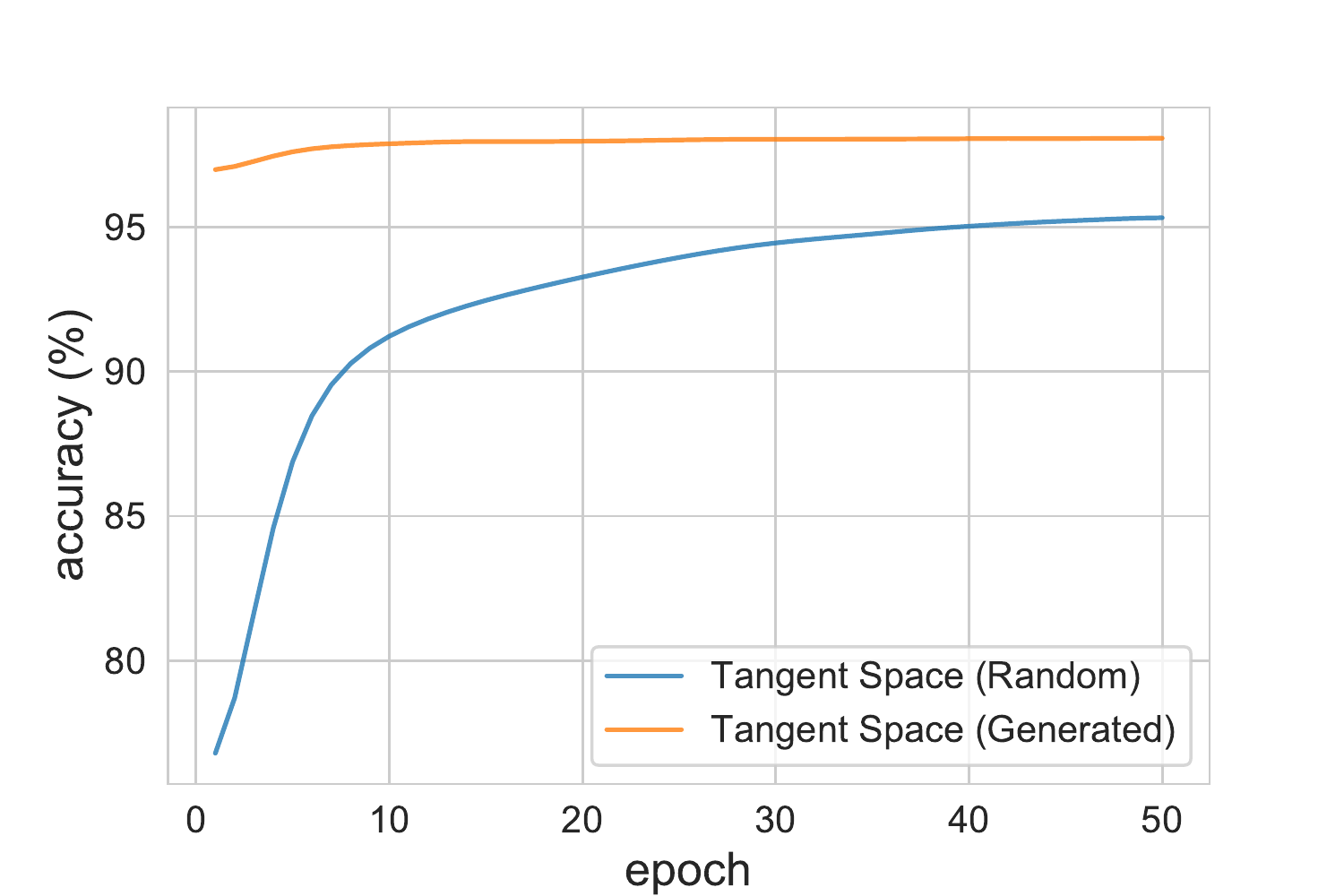}
	\caption{Tangent space comparison ($m=10^3$)}
	\label{fig:tangent}
\end{figure*}

\begin{figure*}
	\centering
	\includegraphics[width=0.30\textwidth]{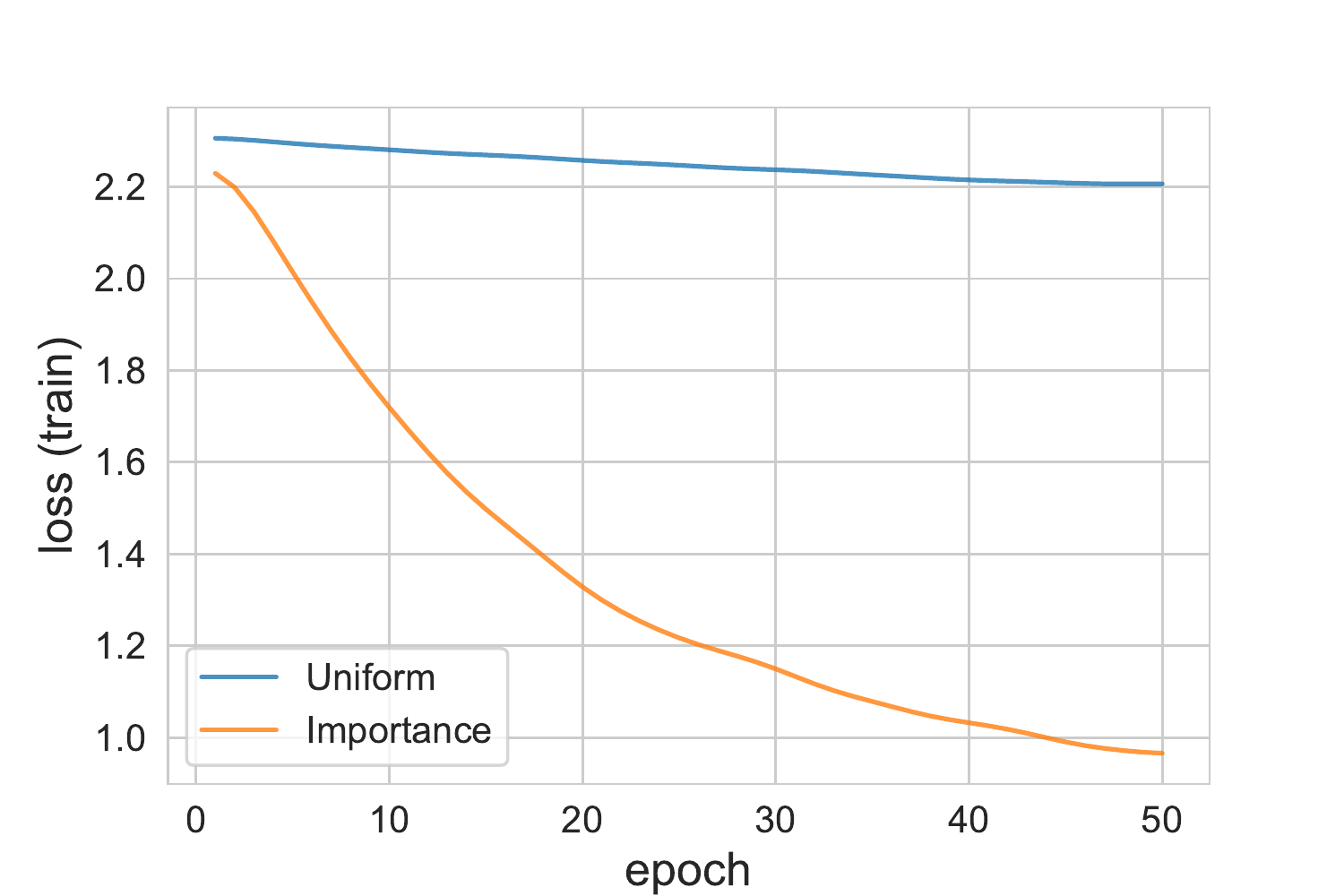}
	\includegraphics[width=0.30\textwidth]{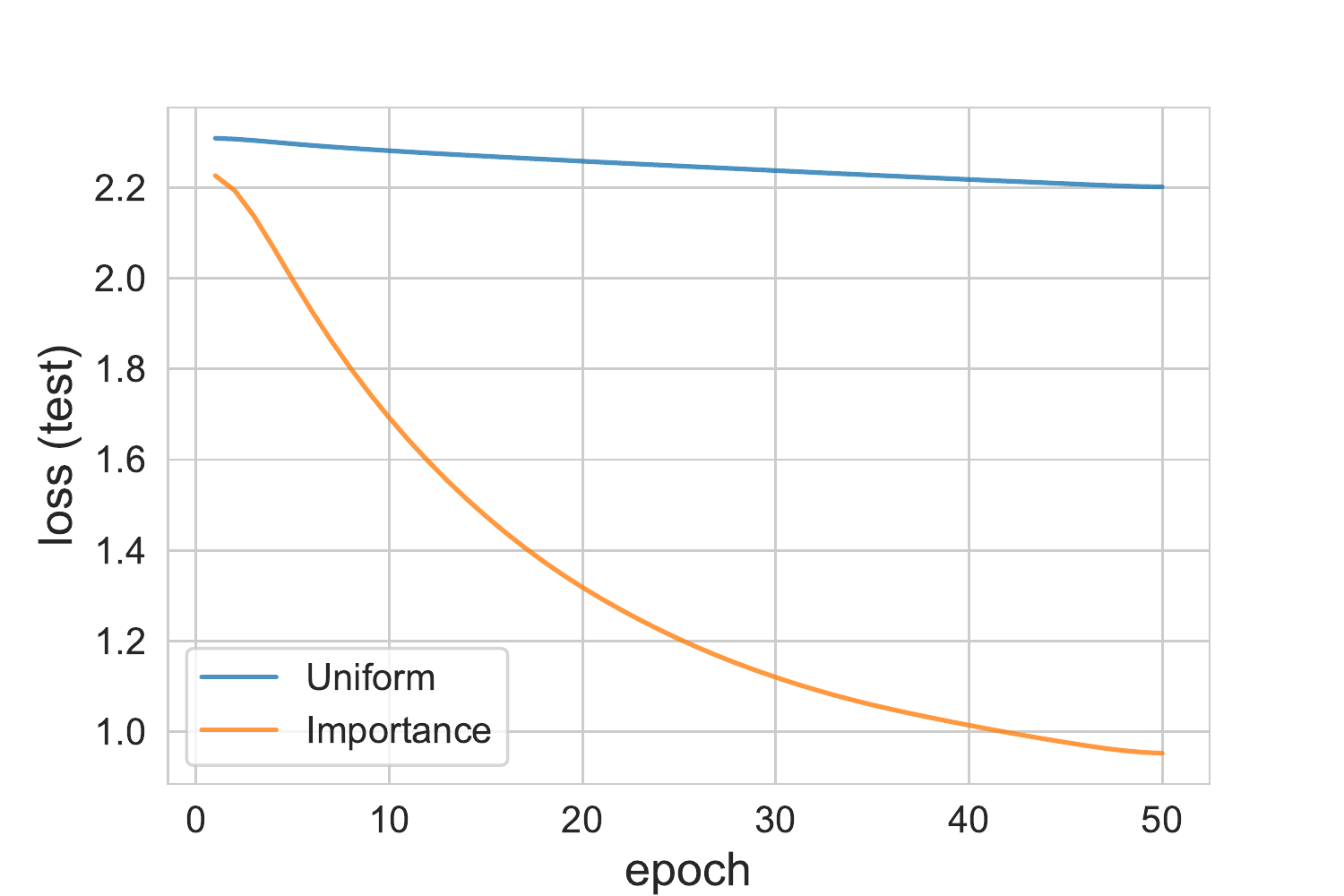}
	\includegraphics[width=0.30\textwidth]{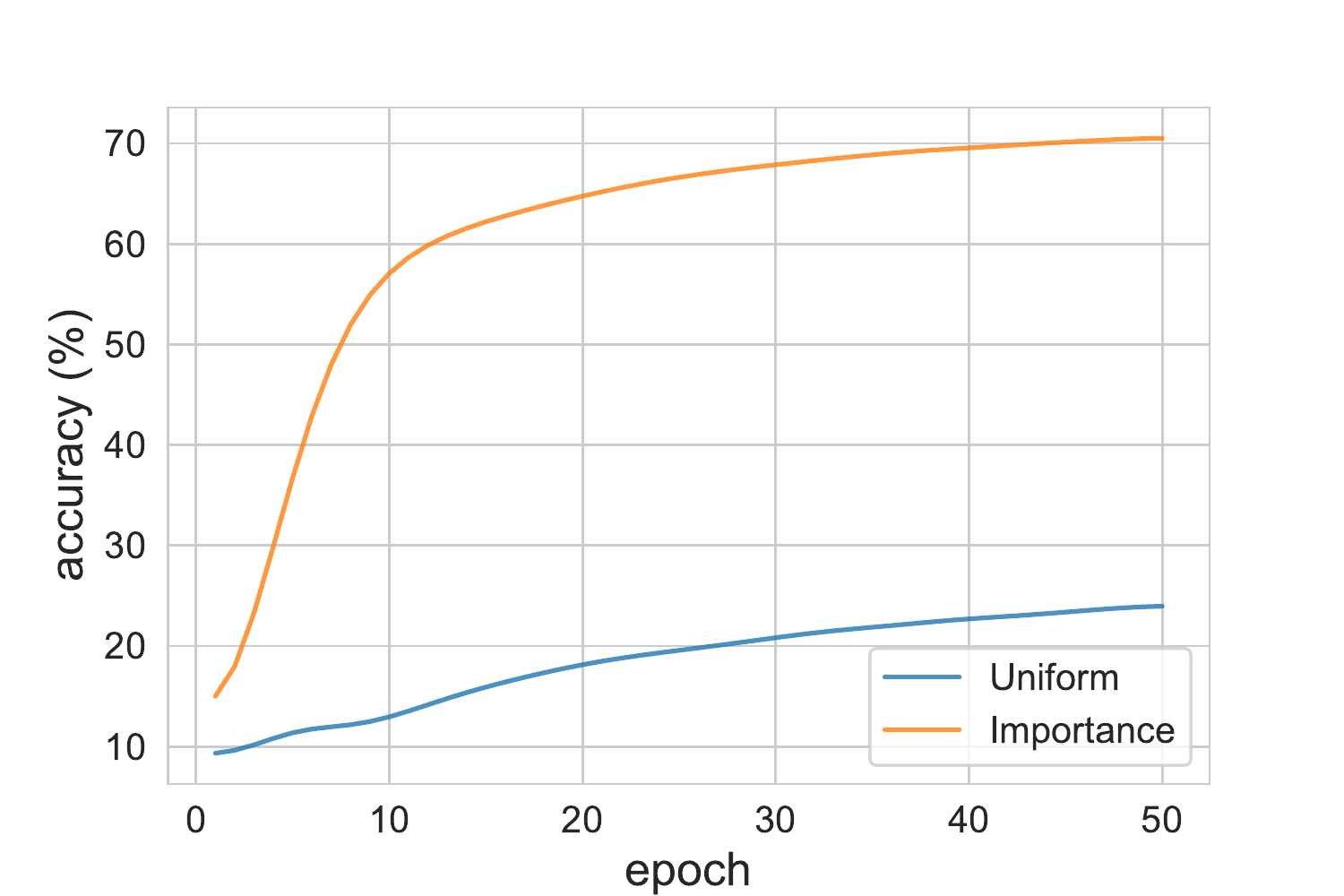}
	\caption{Importance sampling from large NN to remove ``wasted" neurons ($m=10$)}
	\label{fig:is}
\end{figure*}




Figure \ref{fig:mnist_repop} illustrates the {\em repopulation phenomenon} in neural networks. 
In the repopulation process, we use VAE to learn the feature distribution $\rho$ and then sample weights from the learned generative model. 
We then fix the generated features and learn parameter $u$ only as in \eqref{eq:model-rand}, just like the RF method. From this experiment, we can see that 
the performance of the repopulated features outperforms that of the initial random features. This means the random features learned by NN are superior to the Gaussian random features at initialization. This is consistent with the theory of \cite{fang2019over}.

Another approach to examine the effectiveness of $\rho$ is to compare the tangent spaces at the initial and the final solutions using the linear approximation \eqref{eq:lin-approx}. This scenario has also been investigated in \cite{ghorbani2019limitations}. If the representation power of NTK matches that of NN in practice, then the performance using the tangent space at the initialization should be similar to that of the learned distribution $\rho$. We compare random weights and generated ones in Figure \ref{fig:tangent} on the MNIST dataset. Note that the generated weights are learned by VAE at the final solution. In training, both approaches achieve very small errors. However, the generalization ability differs significantly: the learned $\rho$ provides a more robust model in the testing stage. 
Many analysis of NTK investigated the training loss which can become almost zero due to the effectiveness of tangent space. However, the restricted space cannot perform as well as the full NN in terms of the generalization ability.

As indicated by Figure~\ref{fig:nn_mf} (d), many neurons of NN can be ``wasted", and it can be identified by $u$ with proper  regularization. Therefore it is possible to perform importance sampling to select effective weights from a very wide NN, which can also be regarded as an approach of pruning. We train a large NN ($m=10000$) with regularization $10^{-3}$, which leads to many ``wasted" neurons. We want to prune the NN by choosing only $10$ effective neurons, and finetune the weight $u$. There are two strategies to select the neurons: (1) uniform sampling, which does not distinguish the importance of neurons; (2) importance sampling, which take the corresponding $u$ as the importance of neurons. After selecting neurons, we fix the first layer $\theta$ and train $u$. Note that the optimization of $u$ is a convex problem. Figure~\ref{fig:is} shows that the performance of importance sampling outperforms uniform sampling significantly. This also confirms that in a wide network, some neurons may get stuck due to the non strong convexity of the formulation.

 Since random features of NTK are not learned during training, there have been several works that tried to investigate the difference between the lazy training condition in NTK and the actual training process of NNs \cite{ wei2018margin,fang2019over,yehudai2019power,ghorbani2019limitations, ghorbani2019linearized, ghorbani2020neural}.
 Notably,     \cite{yehudai2019power}  showed that random features
cannot be used to learn even a single ReLU neuron unless the number of the  hidden units is  exponentially large in $d$. The authors in \cite{ghorbani2019limitations} considered the quadratic activation function and showed that Gradient Descent  achieves  a lower prediction  risk in the actual training process when the number of neurons is small. As discussed earlier, \cite{fang2019over} showed that  with appropriate regularization, NN can learn optimal feature representations that are superior to random features.

\fangcong{
Because MF outperforms NTK in the feature learning perspective, in many cases, better generalization bounds can be obtained for MF than those of NTK. In particular, shown by \cite{fang2019over} and \cite{wei2018margin}, learning a two-layer NN with an $\ell$-$2$ norm regularizer  on the weights is equivalent to solving an $\ell$-$1$ norm regularized problem in the feature space. This is consistent with the empirical observation that MF learns meaningful features because $\ell$-$1$ regularization has a strong capability for feature selection and sparse representation learning.
In contrast, the kernel methods typically consider an $\ell$-$2$ norm regularizer. Moreover, in \cite{wei2018margin} a simple $d$-dimensional distribution was constructed, for which  MF needs $O(d)$ samples to learn. However, kernel methods (including NTK) require at least $\Omega(d^2)$ samples, which demonstrated the superiority of MF in terms of generalization.  Recently, \cite{chizat2020implicit} obtained an interesting result which shows that even without a regularizer, Gradient Descent can implicitly converges to the $\ell$-$1$ norm regularized solution in the mean-field limit.}

\section{Overparameterized Deep Neural Networks}\label{sec:deepnn}

We have explained the concepts of NTK and MF using two-layer neural networks.
A number of papers have considered extensions of these models to deep neural networks. 

\subsection{NTK}

In general, NTK can be generalized to  DNNs without much difficulty, e.g. \cite{allen2019convergence, du2018gradient,zou2018stochastic}, and the technique can also be generalized to handle more complex  topological structures, such as  Recurrent NNs \cite{allen2019can} and Residual NNs  \cite{du2018gradient}. 

In these approaches, with proper initialization, we can linearize the nonlinear NN models at the initialization, similar to what we have done for two-layer NNs. By showing that the training process with small learning rates leads to zero-training error within a small neighbor of the initialization, the entire NN train lies in the so-called NTK (or lazy-learning) regime, and the linear approximation is effective throughout training. Similar to the situation of two-layer NN, this requires specialized initialization and specialized learning rate which are often different from what are used by practitioners. 

One difficulty with the NTK approach for deep neural networks is that it cannot satisfactorily explain the benefit of using deeper structures.
This because the NTK view essentially corresponds to a linear model using an infinite dimensional random feature representation that defines the underlying NTK. Although with deeper structures, we add more and more random features, similar to the situation of the two-layer NNs, these features are not learned. 

If we want to apply NTK to real problems, efficient computation of the NTK kernel is necessary, which may require special design. For example,
an efficient exact algorithm to compute Convolutional NTK was proposed in \cite{arora2019exact}. In practice, kernel methods have a quadratic complexity with respect to the number of training data, and the computational cost can be prohibitive for big data applications. Various algorithms have been investigated to alleviate this problem in the traditional kernel learning literature. We refer the readers to \cite{SchSmo18} and references therein.

\subsection{MF}

Unlike NTK, it is nontrivial to generalize MF to deep neural networks. 
There were a number of recent works that attempted to generalize MF \cite{ sirignano2019mean2,fang2019convex, araujo2019mean,nguyen2020rigorous,chen2020mean, fang2020modeling,chizat2020implicit}. This is still an active research area which has not matured. We will thus describe some of the challenges and the latest results.

First, it is not easy to formulate the continuous limit of DNNs. Consider a three-layer NN as an example.  The hidden units of the upper layer are functions of hidden units of the lower layer. However, if we allow the number of hidden units of the lower layer go to infinity (as we do in the two layer NN), then there are infinitely many features for every hidden unit of the upper hidden layer. If we let the number of hidden units of the upper layer to go to infinity, then there are infinitely many such functions, each with infinitely many features (each feature corresponds to a hidden unit of the lower layer). It is nontrivial to model these functions mathematically. One of the attempted approaches is to model DNNs with nested measures (also known as multi-layer measures \cite{dawson1982wandering,dawson2018multilevel}). However, as mentioned in \cite{sirignano2019mean2}, the mathematical limit may not be well-defined.  Another  approach considered the  continuous limit of DNNs under special conditions.  For example,  \cite{araujo2019mean,nguyen2020rigorous} investigated the continuous limit of DNNs under the initialization that all weights were i.i.d.~realizations of a \emph{fixed} distribution (with finite variance) independent of the number of hidden units. Unfortunately, in such setting, all neurons in a middle layer will have the same output value at initialization, and this property holds during the entire training process. It is clearly not an appropriate mathematical model for general DNNs. 
In real applications, initialization strategies, e.g.,  \cite{glorot2010understanding,he2015delving} sample the NN weights from  $\mathcal{N}(0,O(m))$, with variance approaching infinity as the number of hidden nodes $m$ goes to $\infty$. More recently,   \cite{fang2020modeling} designed a new mean-field framework for DNNs, in which  a  DNN is represented by probability measures and functions over outputs of the hidden units   instead of the neural network parameters.  This new representation   overcomes the   degenerate situation exited in some earlier attempts, where all the hidden units essentially have only one  meaningful hidden unit in each middle layer.

A second difficulty is that  a DNN cannot be regarded as a linear model with respect to the distribution of the parameters. Unlike the case of two-layer NN, which is convex with respect to a reparameterization of the model using the corresponding feature distribution, it is much harder to derive a convex formulation of DNN with appropriate reparameterization. Therefore, the global minimum  is hard to be identified and Gradient Descent  potentially leads to sub-optimal solutions. Recently,   \cite{nguyen2020rigorous} and \cite{fang2020modeling} showed that Gradient Descent can find a global minimal solution for three-layer and multi-layer DNNs,  respectively. Notably,  they assumed that  no regularization is imposed and the activation function can achieve universal approximation. Under such conditions,  the global minimum can be  identified as $0$.  Another remarkable work  is \cite{fang2019convex} and the closely related study \cite{ZGFLZ2020landscape}, in which the authors introduced a new technique, called neural feature repopulation (NFR), to reparameterize the DNNs. Using the NFR technique, one can decouple the  distributions of the features from the loss function and their impact can be integrated into the regularizer. Surprisingly,  with suitable regularizers,  it can be shown that the overall objective function under the special reparameterization is convex, which is analogous to the case of two-layer NNs. Moreover,  they   proposed  a  new  optimization  process to find the global minimal solution under such  regularizers.
It remains an open theoretical question to show that gradient descent type of algorithms can find a global optimal solution for the associated convex formulation.


\section{Complexity Analysis for overparameterized NNs}\label{sec:complexity}

The theoretical properties of the linearized system in the NTK view is much easier to analyze. Therefore it is possible to prove rigorous convergence and statistical complexity bounds under the NTK regime, and polynomial convergence rates can be obtained  under various conditions. 

For two-layer NNs, for example, by assuming that the  minimum eigenvalue  of the kernel matrix for the  training data is positive, denoted as  $\lambda_0$, it was shown in \cite{du2019gradient}  that when the number of hidden units is greater than $n^{6}\lambda_0^{-4}\delta^{-2}$, with a learning rate  of $\eta =O(\lambda_0n^{-2})$,  then with probability at least $1-\delta$, the gradient descent method finds an $\epsilon$-global minimum in $O\left(\eta^{-1}\lambda_0^{-1}\log(\epsilon^{-1})\right)$ steps.  Before \cite{du2019gradient}, \cite{li2018learning} studied a different data assumption.   They showed that a polynomial convergence rate can be  achieved  under appropriate separability conditions of the data.

The above results can be generalized to DNNs. For example,  in \cite{du2018gradient}, the authors   showed that as long as the number of hidden units  is larger than $\tilde{\Omega}( \text{poly}(\lambda_0, n) 2^L)$, Gradient Descent finds a global minimal solution in  $\tilde{O}( \text{poly}(\lambda_0, n) 2^L)$ steps for standard $L$-layer DNNs, where $\tilde{\Omega}$ and $\tilde{O}$ hide poly-logarithmic terms. Moreover, for the  Residual NNs, the exponential dependencies on $L$ can be reduced to polynomial dependency. Similarly, the authors  in  \cite{allen2019convergence, zou2018stochastic} adopted  the  data assumption in  \cite{li2018learning} and  achieved polynomial complexities.

Some other researchers, e.g., \cite{allen2019can,hanin2019finite,bai2019beyond, huang2019dynamics}, have  tried to model NNs beyond linear approximation of NTK, typically second-order approximation. 
For example, \cite{allen2019can}, \cite{bai2019beyond} and \cite{chen2020towards} proposed a training procedure with randomization techniques to extract the  second-order approximation,   sharpening  complexity bounds. In general, the  second-order approximation  satisfies the so-called \emph{strict saddle property} \cite{ge2015escaping},  thus are  solvable efficiently by saddle-escaping algorithms, e.g. \cite{jin2017escape,fang2018spider,fang2019sharp}.  Specifically, \cite{bai2019beyond} showed that  complexity bounds for learning polynomials on
uniform distributions are lower than those of NTK by a factor of $O(d)$.

\section{Other Mathematical Models of NNs}\label{sec:othermodel}

A number of recent works have considered approximation properties of neural networks, leading to better understanding of why deep neural networks are superior to shallow networks in terms of function approximation. It is well known that two layer neural networks are universal approximators \cite{leshno1993multilayer,barron1993universal}. However, for certain functions that can be represented by deep neural networks with a small number of nodes, exponentially number of nodes are needed to represent them with shallow neural networks \cite{liang2017deep,telgarsky2016benefits}. Related results show that deep neural networks can represent any function with a constant number of nodes per layer 
\cite{lu2017expressive,hanin2019universal}, which suggest a trade-off between depth and width in terms of universal approximation. More generally, in order to represent a complex function, we can either increase a network's width, or its depth. It was observed in practice that it is beneficial to increase both depth and width simultaneously to balance the trade-off \cite{tan2019efficientnet}.

Before the development of recent mathematical models of overparameterized NN such as NTK and MF, which tried to formulate the NN optimization procedure as convex optimization, there were developments in the machine learning research community that focused on the non-convex optimization aspect of NNs. 

In order to understand the NN training process, a number of earlier works studied the loss landscape of NNs. For example, several researchers observed that NN's generalization ability is related to the sharpness/flatness of the local minimal solution resulted from training, and discussed different methods to characterize flatness \cite{hochreiter1997flat,keskar2016large,li2018visualizing}. There are also works, e.g., \cite{papyan2018full,ghorbani2019investigation}, which attempted to understand the Hessian matrices of  neural networks.
With the help of restrictive assumptions, or for specialized models, 
a number of  earlier works, e.g., \cite{kawaguchi2016deep,soudry2016no,hardt2016identity, freeman2016topology, safran2017spurious,nguyen2017loss,zhou2017critical,du2018power},  studied the theoretical characterizations of NN  landscapes. For example,  under the assumption that the input follows Gaussian distribution or the activation function is linear or quadratic.  
These results, in general, showed that for any NN that satisfies strict saddle property, standard saddle-escaping algorithms can converge to a global minimal solution.  We also refer
the readers to the review \cite{sun2020global} and the references therein  for the  global landscape of NNs.

Related studies of neural network training were investigated from the generic non-convex optimization point of view, where a main issue was the complexity of stochastic optimization algorithms such as SGD to escape saddle points and converge to local minimal solutions  \cite{ge2015escaping,jin2017escape}. This question was resolved satisfactorily for general non-convex problems, where both the convergence rate of SGD and that of the optimal stochastic algorithm were known \cite{fang2018spider,fang2019sharp,arjevani2019lower}.

The kernel representation in the RF/NTK view has a natural connection to Gaussian processes, which has a Bayesian statistics interpretation.  The earliest study of overparameterized infinite-width NN was motivated by this Bayesian interpretation~\cite{Neal95-thesis,Williams97-nips,lee2017deep}, where the relationship of infinite-width NN and Gaussian processes were investigated, which is only based on random feature.  More recently, some paper also investigate the kernel form of NTK regime to perform Bayesian inference \cite{lee2019wide}, which has a larger function class than NF.
Moreover, the Bayesian interpretation can be used to derive uncertainty estimation for neural networks. For example, it was argued in \cite{galgha16-icml} using the Gaussian process point of view that dropout \cite{SHKSS-dropout} can be used to obtain uncertainty estimation for neural networks.

\section{Conclusion and Further Directions}\label{sec:conclusion}

Neural network has become an essential tool in machine learning and artificial intelligence, with a wide range of applications. Although there have been significant empirical progresses, theoretical understanding is rather limited, due to the complexity of the non-convexity in NN modeling. It has been noted by practitioners that overparameterized NNs are easier to optimize, and the solutions are often reproducible with good performances that are difficult to explain from a non-convex optimization point of view. To explain this mystery, there have been numerous works to develop mathematical models for overparameterized neural networks in recent years. Due to these efforts, we begin to understand how neural network works, especially in the continuous limit of overparameterized NNs. Surprisingly, under these models, overparameterized neural networks behave more like convex systems, which can explain why they lead to reproducible results observed in practice. 
\fangcong{
There are many research activities in developing better mathematical theories of DNNs. We outline some of the current directions that we feel are particularly promising. 
\begin{itemize}
    \item For two-layer NNs, NTK can achieve a polynomial computational cost, although a relatively weak generalization result.  In comparison, MF  achieves better generalization, but lacks quantitative computational results under general conditions.  
    Therefore we need more sophisticated analysis of two-layer NNs showing better generalization behavior than NTK (especially in terms of feature learning), with a polynomial computational complexity. 
\item The understanding of deep NNs is still quite limited. Although it can be shown that GD converges globally in the NTK regime, in the existing analysis, the weight updates can be ignored except for the second to the last layer. This is clearly inconsistent with practice. Moreover, because NTK is effectively a  linear (single-layer) model  with respect to random features, existing results on DNN approximation imply immediately that representations requiring deep structures can not be learned efficiently by NTK. As an example,  the random features cannot even represent a single ReLU neuron unless there are  exponentially number of hidden units \cite{yehudai2019power}. Although this gap can be potentially addressed by the MF view, its analysis is even further behind in that we still do not have a satisfactory theory of MF for DNNs. Even if a satisfactory theory of MF is developed, quantitative results on generalization and optimization remain open. Therefore for DNNs, both optimization and generalization analysis require further study. As one interesting work in this direction, \cite{allen2020backward}  
established a principle called ``backward feature correction" and showed that Gradient Descent can learn hierarchical features when the activation function is quadratic.  We expect to see more results of this type that can truly illustrate the benefits of deep structures beyond the inherently shallow structure of NTK.
\item  The success of NNs has been largely attributed to their abilities to learn discriminative features. Related topics of transfer learning, pre-training, semi-supervised learning, etc have been actively studied by practitioners with great successes. We expect more and more theoretical investigations of these topics in the future, which will inevitably lead to better understanding of NNs ability to learn feature representations. This may inspire the development of more robust algorithms for representation learning.  
\item In recent years, many specialized NN architectures and components are designed by practitioners that are effective in various different tasks. For example, architectures such as ResNet, CNNs, Transformers, and components such as attention, batch normalization etc have become widely used. We start to see theoretical analysis for such architectures and components.
For example, in \cite{huang2020deep}, the authors provided theoretic justification for Res-Net, and showed  that Res-Nets generalize better than DNNs by comparing the neural tangent kernels.  We expect more sophisticated theoretical analysis of special NN structures can lead to better understanding,  and eventually more effective neural architecture design. 
\item Practitioners have made many interesting empirical observations of NNs that remain to be explained theoretically.  For example, \cite{nakkiran2019deep} showed that NN learning exhibits a {\em double descent  phenomenon}, where when we increase the model size or the number of training epochs, the test performance deteriorates first, then becomes better. In another work, \cite{he2019local} observed the so-called {\em local elasticity} phenomenon where the prediction of a datum $x'$ will not be significantly affected after a stochastic gradient descent update at a datum $x$ which is not close to $x'$. As another example, \cite{papyan2020prevalence} observed a phenomenon called {\em Neural Collapse}, which states that predicted class means collapse to the vertices of a Simplex Equiangular Tight Frame at the final training stage. Developing theoretical explanations of such practical phenomena can lead to better understanding of how NN works. 
\end{itemize}
}
Finally, we conclude that  the study of overparameterized NNs is still in its infancy.  We expect deep mathematical insights obtained from these theoretical investigations will help us to develop solid theoretical foundations for NNs, and motivate effective algorithms and architectures  in the coming years.

\section{Code availability}

Codes for illustration is available at Github\footnote{ \url{https://github.com/hendrydong/NTK-and-MF-examples}}.

\bibliography{myrefs.bib}

\begin{thebibliography}{100}
\providecommand{\url}[1]{#1}
\csname url@samestyle\endcsname
\providecommand{\newblock}{\relax}
\providecommand{\bibinfo}[2]{#2}
\providecommand{\BIBentrySTDinterwordspacing}{\spaceskip=0pt\relax}
\providecommand{\BIBentryALTinterwordstretchfactor}{4}
\providecommand{\BIBentryALTinterwordspacing}{\spaceskip=\fontdimen2\font plus
\BIBentryALTinterwordstretchfactor\fontdimen3\font minus
  \fontdimen4\font\relax}
\providecommand{\BIBforeignlanguage}[2]{{%
\expandafter\ifx\csname l@#1\endcsname\relax
\typeout{** WARNING: IEEEtran.bst: No hyphenation pattern has been}%
\typeout{** loaded for the language `#1'. Using the pattern for}%
\typeout{** the default language instead.}%
\else
\language=\csname l@#1\endcsname
\fi
#2}}
\providecommand{\BIBdecl}{\relax}
\BIBdecl

\bibitem{krizhevsky2012imagenet}
A.~Krizhevsky, I.~Sutskever, and G.~E. Hinton, ``Imagenet classification with
  deep convolutional neural networks,'' in \emph{Advances in Neural Information
  Processing Systems}, 2012.

\bibitem{he2016deep}
K.~He, X.~Zhang, S.~Ren, and J.~Sun, ``Deep residual learning for image
  recognition,'' in \emph{Proceedings of the IEEE conference on Computer Vision
  and Pattern Recognition}, 2016.

\bibitem{hinton2012deep}
G.~Hinton, L.~Deng, D.~Yu, G.~E. Dahl, A.-r. Mohamed, N.~Jaitly, A.~Senior,
  V.~Vanhoucke, P.~Nguyen, T.~N. Sainath \emph{et~al.}, ``Deep neural networks
  for acoustic modeling in speech recognition: The shared views of four
  research groups,'' \emph{IEEE Signal processing magazine}, vol.~29, no.~6,
  pp. 82--97, 2012.

\bibitem{bahdanau2014neural}
D.~Bahdanau, K.~Cho, and Y.~Bengio, ``Neural machine translation by jointly
  learning to align and translate,'' \emph{arXiv preprint arXiv:1409.0473},
  2014.

\bibitem{SHKSS-dropout}
N.~Srivastava, G.~Hinton, A.~Krizhevsky, I.~Sutskever, and R.~Salakhutdinov,
  ``Dropout: A simple way to prevent neural networks from overfitting,''
  \emph{Journal of Machine Learning Research}, vol.~15, no.~1, p. 1929–1958,
  Jan. 2014.

\bibitem{ioffe15}
S.~Ioffe and C.~Szegedy, ``Batch normalization: Accelerating deep network
  training by reducing internal covariate shift,'' in \emph{International
  Conference on Machine Learning}, 2015, pp. 448--456.

\bibitem{zhang2016understanding}
C.~Zhang, S.~Bengio, M.~Hardt, B.~Recht, and O.~Vinyals, ``Understanding deep
  learning requires rethinking generalization,'' in \emph{International
  Conference on Learning Representation}, 2017.

\bibitem{Neal95-thesis}
R.~Neal, ``Bayesian learning for neural networks,'' Ph.D. dissertation,
  University of Toronto, 1995.

\bibitem{Williams97-nips}
C.~K. Williams, ``Computing with infinite networks,'' in \emph{Advances in
  Neural Information Processing Systems}, 1997.

\bibitem{rahimi2009weighted}
A.~Rahimi and B.~Recht, ``Weighted sums of random kitchen sinks: Replacing
  minimization with randomization in learning,'' in \emph{Advances in Neural
  Information Processing Systems}, 2009.

\bibitem{kleene1951representation}
S.~Kleene, ``Representation of events in nerve nets and finite automata,''
  \emph{Automata studies}, vol.~3, p.~41, 1951.

\bibitem{rahimi2008uniform}
A.~Rahimi and B.~Recht, ``Uniform approximation of functions with random
  bases,'' in \emph{2008 46th Annual Allerton Conference on Communication,
  Control, and Computing}.\hskip 1em plus 0.5em minus 0.4em\relax IEEE, 2008,
  pp. 555--561.

\bibitem{bach2017equivalence}
F.~Bach, ``On the equivalence between kernel quadrature rules and random
  feature expansions,'' \emph{The Journal of Machine Learning Research},
  vol.~18, no.~1, pp. 714--751, 2017.

\bibitem{SchSmo18}
B.~Schlkopf, A.~J. Smola, and F.~Bach, \emph{Learning with kernels: support
  vector machines, regularization, optimization, and beyond}.\hskip 1em plus
  0.5em minus 0.4em\relax the MIT Press, 2018.

\bibitem{daniely2017sgd}
A.~Daniely, ``{SGD} learns the conjugate kernel class of the network,'' in
  \emph{Advances in Neural Information Processing Systems}, 2017.

\bibitem{daniely2016toward}
A.~Daniely, R.~Frostig, and Y.~Singer, ``Toward deeper understanding of neural
  networks: The power of initialization and a dual view on expressivity,'' in
  \emph{Advances in Neural Information Processing Systems}, 2016.

\bibitem{jacot2018neural}
A.~Jacot, F.~Gabriel, and C.~Hongler, ``Neural tangent kernel: Convergence and
  generalization in neural networks,'' in \emph{Advances in Neural Information
  Processing Systems}, 2018.

\bibitem{du2018gradient}
S.~S. Du, J.~D. Lee, H.~Li, L.~Wang, and X.~Zhai, ``Gradient descent finds
  global minima of deep neural networks,'' in \emph{International Conference on
  Machine Learning}, 2019.

\bibitem{chizat2019lazy}
L.~Chizat, E.~Oyallon, and F.~Bach, ``On lazy training in differentiable
  programming,'' in \emph{Advances in Neural Information Processing Systems},
  2019.

\bibitem{li2018learning}
Y.~Li and Y.~Liang, ``Learning overparameterized neural networks via stochastic
  gradient descent on structured data,'' in \emph{Advances in Neural
  Information Processing Systems}, 2018.

\bibitem{allen2019can}
Z.~Allen-Zhu and Y.~Li, ``What can resnet learn efficiently, going beyond
  kernels?'' in \emph{Advances in Neural Information Processing Systems}, 2019.

\bibitem{zou2018stochastic}
D.~Zou, Y.~Cao, D.~Zhou, and Q.~Gu, ``Stochastic gradient descent optimizes
  over-parameterized deep relu networks,'' in \emph{Advances in Neural
  Information Processing Systems}, 2018.

\bibitem{arora2019fine}
S.~Arora, S.~S. Du, W.~Hu, Z.~Li, and R.~Wang, ``Fine-grained analysis of
  optimization and generalization for overparameterized two-layer neural
  networks,'' in \emph{International Conference on Machine Learning}, 2019.

\bibitem{allen2019convergence}
Z.~Allen-Zhu, Y.~Li, and Z.~Song, ``A convergence theory for deep learning via
  over-parameterization,'' in \emph{International Conference on Machine
  Learning}, 2019.

\bibitem{allen2019learning}
Z.~Allen-Zhu, Y.~Li, and Y.~Liang, ``Learning and generalization in
  overparameterized neural networks, going beyond two layers,'' in
  \emph{Advances in Neural Information Processing Systems}, 2019.

\bibitem{yang2019scaling}
G.~Yang, ``Scaling limits of wide neural networks with weight sharing: Gaussian
  process behavior, gradient independence, and neural tangent kernel
  derivation,'' \emph{arXiv preprint arXiv:1902.04760}, 2019.

\bibitem{oymak2018overparameterized}
S.~Oymak and M.~Soltanolkotabi, ``Overparameterized nonlinear learning:
  Gradient descent takes the shortest path?'' in \emph{International Conference
  on Machine Learning}, 2018.

\bibitem{zou2019improved}
D.~Zou and Q.~Gu, ``An improved analysis of training over-parameterized deep
  neural networks,'' in \emph{Advances in Neural Information Processing
  Systems}, 2019.

\bibitem{mei2019mean}
S.~Mei, T.~Misiakiewicz, and A.~Montanari, ``Mean-field theory of two-layers
  neural networks: dimension-free bounds and kernel limit,'' in \emph{Annual
  Conference on Learning Theory}, 2019.

\bibitem{arora2019exact}
S.~Arora, S.~S. Du, W.~Hu, Z.~Li, R.~Salakhutdinov, and R.~Wang, ``On exact
  computation with an infinitely wide neural net,'' in \emph{Advances in Neural
  Information Processing Systems}, 2019.

\bibitem{tzen2020mean}
B.~Tzen and M.~Raginsky, ``A mean-field theory of lazy training in two-layer
  neural nets: entropic regularization and controlled mckean-vlasov dynamics,''
  \emph{arXiv:2002.01987}, 2020.

\bibitem{yang2019theoretical}
Z.~Yang, Y.~Xie, and Z.~Wang, ``A theoretical analysis of deep q-learning,''
  \emph{arXiv preprint arXiv:1901.00137}, 2019.

\bibitem{cai2019neural}
Q.~Cai, Z.~Yang, J.~D. Lee, and Z.~Wang, ``Neural temporal-difference learning
  converges to global optima,'' in \emph{Advances in Neural Information
  Processing Systems}, 2019.

\bibitem{liu2019neural}
B.~Liu, Q.~Cai, Z.~Yang, and Z.~Wang, ``Neural proximal/trust region policy
  optimization attains globally optimal policy,'' in \emph{Advances in Neural
  Information Processing Systems}, 2019.

\bibitem{wang2019neural}
L.~Wang, Q.~Cai, Z.~Yang, and Z.~Wang, ``Neural policy gradient methods: Global
  optimality and rates of convergence,'' \emph{arXiv preprint
  arXiv:1909.01150}, 2019.

\bibitem{zhang2020generative}
Y.~Zhang, Q.~Cai, Z.~Yang, and Z.~Wang, ``Generative adversarial imitation
  learning with neural networks: Global optimality and convergence rate,''
  \emph{arXiv preprint arXiv:2003.03709}, 2020.

\bibitem{liao2020provably}
L.~Liao, Y.-L. Chen, Z.~Yang, B.~Dai, Z.~Wang, and M.~Kolar, ``Provably
  efficient neural estimation of structural equation model: An adversarial
  approach,'' in \emph{Advances in Neural Information Processing Systems},
  2020.

\bibitem{su2019learning}
L.~Su and P.~Yang, ``On learning over-parameterized neural networks: A
  functional approximation prospective,'' in \emph{Advances in Neural
  Information Processing Systems}, 2019.

\bibitem{cao2019generalization}
Y.~Cao and Q.~Gu, ``Generalization bounds of stochastic gradient descent for
  wide and deep neural networks,'' in \emph{Advances in Neural Information
  Processing Systems}, 2019.

\bibitem{ji2019polylogarithmic}
Z.~Ji and M.~Telgarsky, ``Polylogarithmic width suffices for gradient descent
  to achieve arbitrarily small test error with shallow relu networks,'' in
  \emph{International Conference on Learning Representation}, 2020.

\bibitem{du2019gradient}
S.~S. Du, X.~Zhai, B.~Poczos, and A.~Singh, ``Gradient descent provably
  optimizes over-parameterized neural networks,'' in \emph{International
  Conference on Learning Representation}, 2019.

\bibitem{lecun1998gradient}
Y.~LeCun, L.~Bottou, Y.~Bengio, and P.~Haffner, ``Gradient-based learning
  applied to document recognition,'' \emph{Proceedings of the IEEE}, vol.~86,
  no.~11, pp. 2278--2324, 1998.

\bibitem{li2019towards}
Y.~Li, C.~Wei, and T.~Ma, ``Towards explaining the regularization effect of
  initial large learning rate in training neural networks,'' in \emph{Advances
  in Neural Information Processing Systems}, 2019, pp. 11\,674--11\,685.

\bibitem{jastrzkebski2017three}
S.~Jastrzkebski, Z.~Kenton, D.~Arpit, N.~Ballas, A.~Fischer, Y.~Bengio, and
  A.~Storkey, ``Three factors influencing minima in {SGD},'' 2018.

\bibitem{he2015delving}
K.~He, X.~Zhang, S.~Ren, and J.~Sun, ``Delving deep into rectifiers: Surpassing
  human-level performance on imagenet classification,'' in \emph{International
  Conference on Computer Vision}, 2015.

\bibitem{meie7665}
S.~Mei, A.~Montanari, and P.-M. Nguyen, ``A mean field view of the landscape of
  two-layer neural networks,'' \emph{Proceedings of the National Academy of
  Sciences}, vol. 115, no.~33, pp. E7665--E7671, 2018.

\bibitem{chizat2018global}
L.~Chizat and F.~Bach, ``On the global convergence of gradient descent for
  over-parameterized models using optimal transport,'' in \emph{Advances in
  Neural Information Processing Systems}, 2018.

\bibitem{sirignano2019mean}
J.~Sirignano and K.~Spiliopoulos, ``Mean field analysis of neural networks: A
  central limit theorem,'' \emph{Stochastic Processes and their Applications},
  2019.

\bibitem{sirignano2020mean}
J.~Sirignano and K.~Spiliopoulos, ``Mean field analysis of neural networks: A
  law of large numbers,'' \emph{SIAM Journal on Applied Mathematics}, vol.~80,
  no.~2, pp. 725--752, 2020.

\bibitem{rotskoff2018neural}
G.~M. Rotskoff and E.~Vanden-Eijnden, ``Neural networks as interacting particle
  systems: Asymptotic convexity of the loss landscape and universal scaling of
  the approximation error,'' in \emph{Advances in Neural Information Processing
  Systems}, 2018.

\bibitem{dou2020training}
X.~Dou and T.~Liang, ``Training neural networks as learning data-adaptive
  kernels: Provable representation and approximation benefits,'' \emph{Journal
  of the American Statistical Association}, pp. 1--14, 2020.

\bibitem{wei2018margin}
C.~Wei, J.~D. Lee, Q.~Liu, and T.~Ma, ``Regularization matters: Generalization
  and optimization of neural nets v.s. their induced kernel,'' in
  \emph{Advances in Neural Information Processing Systems}, 2019.

\bibitem{hu2019mean}
K.~Hu, Z.~Ren, D.~Siska, and L.~Szpruch, ``Mean-field langevin dynamics and
  energy landscape of neural networks,'' \emph{arXiv preprint
  arXiv:1905.07769}, 2019.

\bibitem{zhang2020can}
Y.~Zhang, Q.~Cai, Z.~Yang, Y.~Chen, and Z.~Wang, ``Can temporal-difference and
  q-learning learn representation? a mean-field theory,'' \emph{arXiv preprint
  arXiv:2006.04761}, 2020.

\bibitem{engel2001statistical}
A.~Engel and C.~Van~den Broeck, \emph{Statistical mechanics of learning}.\hskip
  1em plus 0.5em minus 0.4em\relax Cambridge University Press, 2001.

\bibitem{ambrosio2008gradient}
L.~Ambrosio, N.~Gigli, and G.~Savar{\'e}, \emph{Gradient flows: in metric
  spaces and in the space of probability measures}.\hskip 1em plus 0.5em minus
  0.4em\relax Springer Science \& Business Media, 2008.

\bibitem{chen2020mean}
Z.~Chen, Y.~Cao, Q.~Gu, and T.~Zhang, ``A generalized neural tangent kernel
  analysis for two-layer neural networks,'' in \emph{Thirty-fourth Conference
  on Neural Information Processing Systems}, 2020.

\bibitem{fang2019over}
C.~Fang, H.~Dong, and T.~Zhang, ``Over parameterized two-level neural networks
  can learn near optimal feature representations,'' \emph{arXiv preprint
  arXiv:1910.11508}, 2019.

\bibitem{atanov2018deep}
A.~Atanov, A.~Ashukha, K.~Struminsky, D.~Vetrov, and M.~Welling, ``The deep
  weight prior,'' in \emph{International Conference on Learning
  Representation}, 2019.

\bibitem{erhan2009visualizing}
D.~Erhan, Y.~Bengio, A.~Courville, and P.~Vincent, ``Visualizing higher-layer
  features of a deep network,'' \emph{University of Montreal}, vol. 1341,
  no.~3, p.~1, 2009.

\bibitem{kingma2013auto}
D.~P. Kingma and M.~Welling, ``Auto-encoding variational bayes,'' \emph{arXiv
  preprint arXiv:1312.6114}, 2013.

\bibitem{ghorbani2019limitations}
B.~Ghorbani, S.~Mei, T.~Misiakiewicz, and A.~Montanari, ``Limitations of lazy
  training of two-layers neural network,'' in \emph{Advances in Neural
  Information Processing Systems}, 2019.

\bibitem{yehudai2019power}
G.~Yehudai and O.~Shamir, ``On the power and limitations of random features for
  understanding neural networks,'' in \emph{Advances in Neural Information
  Processing Systems}, 2019.

\bibitem{ghorbani2019linearized}
B.~Ghorbani, S.~Mei, T.~Misiakiewicz, and A.~Montanari, ``Linearized two-layers
  neural networks in high dimension,'' \emph{arXiv preprint arXiv:1904.12191},
  2019.

\bibitem{ghorbani2020neural}
B.~Ghorbani, S.~Mei, T.~Misiakiewicz, and A.~Montanari, ``When do neural
  networks outperform kernel methods?'' in \emph{Advances in Neural Information
  Processing Systems}, 2020.

\bibitem{chizat2020implicit}
L.~Chizat and F.~Bach, ``Implicit bias of gradient descent for wide two-layer
  neural networks trained with the logistic loss,'' in \emph{Annual Conference
  on Learning Theory}, 2020.

\bibitem{sirignano2019mean2}
J.~Sirignano and K.~Spiliopoulos, ``Mean field analysis of deep neural
  networks,'' \emph{arXiv:1903.04440}, 2019.

\bibitem{fang2019convex}
C.~Fang, Y.~Gu, W.~Zhang, and T.~Zhang, ``Convex formulation of
  overparameterized deep neural networks,'' \emph{arXiv:1911.07626}, 2019.

\bibitem{araujo2019mean}
D.~Ara{\'u}jo, R.~I. Oliveira, and D.~Yukimura, ``A mean-field limit for
  certain deep neural networks,'' \emph{arXiv:1906.00193}, 2019.

\bibitem{nguyen2020rigorous}
P.-M. Nguyen and H.~T. Pham, ``A rigorous framework for the mean field limit of
  multilayer neural networks,'' \emph{arXiv:2001.11443}, 2020.

\bibitem{fang2020modeling}
C.~Fang, J.~D. Lee, P.~Yang, and T.~Zhang, ``Modeling from features: a
  mean-field framework for over-parameterized deep neural networks,''
  \emph{arXiv preprint arXiv:2007.01452}, 2020.

\bibitem{dawson1982wandering}
D.~A. Dawson, K.~J. Hochberg \emph{et~al.}, ``Wandering random measures in the
  fleming-viot model,'' \emph{The Annals of Probability}, vol.~10, no.~3, pp.
  554--580, 1982.

\bibitem{dawson2018multilevel}
D.~A. Dawson, ``Multilevel mutation-selection systems and set-valued duals,''
  \emph{Journal of mathematical biology}, vol.~76, no. 1-2, pp. 295--378, 2018.

\bibitem{glorot2010understanding}
X.~Glorot and Y.~Bengio, ``Understanding the difficulty of training deep
  feedforward neural networks,'' in \emph{International Conference on
  Artificial Intelligence and Statistics}, 2010, pp. 249--256.

\bibitem{ZGFLZ2020landscape}
Y.~Gu, W.~Zhang, C.~Fang, J.~Lee, and T.~Zhang, ``How to characterize the
  landscape of overparameterized convolutional neural networks,'' in
  \emph{Advances in Neural Information Processing Systems}, 2020.

\bibitem{hanin2019finite}
B.~Hanin and M.~Nica, ``Finite depth and width corrections to the neural
  tangent kernel,'' in \emph{International Conference on Learning
  Representation}, 2020.

\bibitem{bai2019beyond}
Y.~Bai and J.~D. Lee, ``Beyond linearization: On quadratic and higher-order
  approximation of wide neural networks,'' in \emph{International Conference on
  Learning Representation}, 2019.

\bibitem{huang2019dynamics}
J.~Huang and H.-T. Yau, ``Dynamics of deep neural networks and neural tangent
  hierarchy,'' in \emph{International Conference on Machine Learning}, 2020.

\bibitem{chen2020towards}
M.~Chen, Y.~Bai, J.~D. Lee, T.~Zhao, H.~Wang, C.~Xiong, and R.~Socher,
  ``Towards understanding hierarchical learning: Benefits of neural
  representations,'' in \emph{Advances in Neural Information Processing
  Systems}, 2020.

\bibitem{ge2015escaping}
R.~Ge, F.~Huang, C.~Jin, and Y.~Yuan, ``Escaping from saddle points -- online
  stochastic gradient for tensor decomposition,'' in \emph{Annual Conference on
  Learning Theory}, 2015, pp. 797--842.

\bibitem{jin2017escape}
C.~Jin, R.~Ge, P.~Netrapalli, S.~M. Kakade, and M.~I. Jordan, ``How to escape
  saddle points efficiently,'' in \emph{International Conference on Machine
  Learning}, 2017.

\bibitem{fang2018spider}
C.~Fang, C.~J. Li, Z.~Lin, and T.~Zhang, ``Spider: Near-optimal non-convex
  optimization via stochastic path-integrated differential estimator,'' in
  \emph{Advances in Neural Information Processing Systems}, 2018.

\bibitem{fang2019sharp}
C.~Fang, Z.~Lin, and T.~Zhang, ``Sharp analysis for nonconvex {SGD} escaping
  from saddle points,'' in \emph{Annual Conference on Learning Theory}, 2019.

\bibitem{leshno1993multilayer}
M.~Leshno, V.~Y. Lin, A.~Pinkus, and S.~Schocken, ``Multilayer feedforward
  networks with a nonpolynomial activation function can approximate any
  function,'' \emph{Neural networks}, vol.~6, no.~6, pp. 861--867, 1993.

\bibitem{barron1993universal}
A.~R. Barron, ``Universal approximation bounds for superpositions of a
  sigmoidal function,'' \emph{IEEE Transactions on Information theory},
  vol.~39, no.~3, pp. 930--945, 1993.

\bibitem{liang2017deep}
S.~Liang and R.~Srikant, ``Why deep neural networks for function
  approximation?'' in \emph{International Conference on Learning
  Representation}, 2017.

\bibitem{telgarsky2016benefits}
M.~Telgarsky, ``Benefits of depth in neural networks,'' in \emph{Annual
  Conference on Learning Theory}, 2016.

\bibitem{lu2017expressive}
Z.~Lu, H.~Pu, F.~Wang, Z.~Hu, and L.~Wang, ``The expressive power of neural
  networks: A view from the width,'' in \emph{Advances in Neural Information
  Processing Systems}, 2017.

\bibitem{hanin2019universal}
B.~Hanin, ``Universal function approximation by deep neural nets with bounded
  width and relu activations,'' \emph{Mathematics}, vol.~7, no.~10, p. 992,
  2019.

\bibitem{tan2019efficientnet}
M.~Tan and Q.~V. Le, ``Efficientnet: Rethinking model scaling for convolutional
  neural networks,'' in \emph{International Conference on Machine Learning},
  2019.

\bibitem{hochreiter1997flat}
S.~Hochreiter and J.~Schmidhuber, ``Flat minima,'' \emph{Neural Computation},
  vol.~9, no.~1, pp. 1--42, 1997.

\bibitem{keskar2016large}
N.~S. Keskar, D.~Mudigere, J.~Nocedal, M.~Smelyanskiy, and P.~T.~P. Tang, ``On
  large-batch training for deep learning: Generalization gap and sharp
  minima,'' in \emph{International Conference on Learning Representation},
  2017.

\bibitem{li2018visualizing}
H.~Li, Z.~Xu, G.~Taylor, C.~Studer, and T.~Goldstein, ``Visualizing the loss
  landscape of neural nets,'' in \emph{Advances in Neural Information
  Processing Systems}, 2018.

\bibitem{papyan2018full}
V.~Papyan, ``The full spectrum of deepnet hessians at scale: Dynamics with sgd
  training and sample size,'' \emph{arXiv preprint arXiv:1811.07062}, 2018.

\bibitem{ghorbani2019investigation}
B.~Ghorbani, S.~Krishnan, and Y.~Xiao, ``An investigation into neural net
  optimization via hessian eigenvalue density,'' in \emph{International
  Conference on Machine Learning}, 2019.

\bibitem{kawaguchi2016deep}
K.~Kawaguchi, ``Deep learning without poor local minima,'' in \emph{Advances in
  Neural Information Processing Systems}, 2016.

\bibitem{soudry2016no}
D.~Soudry and Y.~Carmon, ``No bad local minima: Data independent training error
  guarantees for multilayer neural networks,'' \emph{arXiv preprint
  arXiv:1605.08361}, 2016.

\bibitem{hardt2016identity}
M.~Hardt and T.~Ma, ``Identity matters in deep learning,'' in
  \emph{International Conference on Learning Representation}, 2016.

\bibitem{freeman2016topology}
C.~D. Freeman and J.~Bruna, ``Topology and geometry of half-rectified network
  optimization,'' in \emph{International Conference on Learning
  Representation}, 2017.

\bibitem{safran2017spurious}
I.~Safran and O.~Shamir, ``Spurious local minima are common in two-layer relu
  neural networks,'' in \emph{International Conference on Machine Learning},
  2018.

\bibitem{nguyen2017loss}
Q.~Nguyen and M.~Hein, ``The loss surface of deep and wide neural networks,''
  in \emph{International Conference on Machine Learning}, 2017.

\bibitem{zhou2017critical}
Y.~Zhou and Y.~Liang, ``Critical points of neural networks: Analytical forms
  and landscape properties,'' in \emph{International Conference on Learning
  Representation}, 2018.

\bibitem{du2018power}
S.~S. Du and J.~D. Lee, ``On the power of over-parametrization in neural
  networks with quadratic activation,'' in \emph{International Conference on
  Machine Learning}, 2018.

\bibitem{sun2020global}
R.~Sun, D.~Li, S.~Liang, T.~Ding, and R.~Srikant, ``The global landscape of
  neural networks: An overview,'' \emph{IEEE Signal Processing Magazine},
  vol.~37, no.~5, pp. 95--108, 2020.

\bibitem{arjevani2019lower}
Y.~Arjevani, Y.~Carmon, J.~C. Duchi, D.~J. Foster, N.~Srebro, and B.~Woodworth,
  ``Lower bounds for non-convex stochastic optimization,'' \emph{arXiv preprint
  arXiv:1912.02365}, 2019.

\bibitem{lee2017deep}
J.~Lee, Y.~Bahri, R.~Novak, S.~S. Schoenholz, J.~Pennington, and
  J.~Sohl-Dickstein, ``Deep neural networks as gaussian processes,'' in
  \emph{International Conference on Learning Representation}, 2018.

\bibitem{lee2019wide}
J.~Lee, L.~Xiao, S.~Schoenholz, Y.~Bahri, R.~Novak, J.~Sohl-Dickstein, and
  J.~Pennington, ``Wide neural networks of any depth evolve as linear models
  under gradient descent,'' in \emph{Advances in Neural Information Processing
  Systems}, 2019.

\bibitem{galgha16-icml}
Y.~Gal and Z.~Ghahramani, ``Dropout as a {B}ayesian approximation: Representing
  model uncertainty in deep learning,'' in \emph{International Conference on
  Machine Learning}, 2016.

\bibitem{allen2020backward}
Z.~Allen-Zhu and Y.~Li, ``Backward feature correction: How deep learning
  performs deep learning,'' \emph{arXiv preprint arXiv:2001.04413}, 2020.

\bibitem{huang2020deep}
K.~Huang, Y.~Wang, M.~Tao, and T.~Zhao, ``Why do deep residual networks
  generalize better than deep feedforward networks?--a neural tangent kernel
  perspective,'' in \emph{Advances in Neural Information Processing Systems},
  2020.

\bibitem{nakkiran2019deep}
P.~Nakkiran, G.~Kaplun, Y.~Bansal, T.~Yang, B.~Barak, and I.~Sutskever, ``Deep
  double descent: Where bigger models and more data hurt,'' in
  \emph{International Conference on Learning Representation}, 2020.

\bibitem{he2019local}
H.~He and W.~Su, ``The local elasticity of neural networks,'' in
  \emph{International Conference on Learning Representation}, 2019.

\bibitem{papyan2020prevalence}
V.~Papyan, X.~Han, and D.~L. Donoho, ``Prevalence of neural collapse during the
  terminal phase of deep learning training,'' \emph{Proceedings of the National
  Academy of Sciences}, vol. 117, no.~40, pp. 24\,652--24\,663, 2020.

\end{thebibliography}

\bibliographystyle{IEEEtran}

\begin{IEEEbiographynophoto}{Cong Fang} Cong Fang
received his Ph.D. degree from Peking University in 2019. He is currently a Postdoctoral Researcher at University of Pennsylvania. His research interests include machine learning and optimization.
\end{IEEEbiographynophoto}

\begin{IEEEbiographynophoto}{Hanze Dong}
Hanze Dong is a PhD student at the Hong Kong University of Science and Technology. He received BSc degree in mathematics from Fudan University in 2019. His research includes machine learning algorithms and theory.
\end{IEEEbiographynophoto}

\begin{IEEEbiographynophoto}{Tong Zhang}
Tong Zhang received the BA degree in mathematics and computer science from Cornell University, and the PhD degree in computer science from Stanford University. He is a chair professor of Computer Science and Mathematics at the Hong Kong University of Science and Technology. He is a fellow of IEEE, American Statistical Association, and Institute of Mathematical Statistics. His research interests are machine learning, big data and their applications.     
\end{IEEEbiographynophoto}

\end{document}